\documentclass[journal]{elsarticle}
\usepackage[noadjust]{cite}
\usepackage{graphicx,natbib,amssymb}
\usepackage{float}
\usepackage{ulem}
\usepackage{ifpdf}
\usepackage{array}
\usepackage{url}
\usepackage{graphicx}
\usepackage{courier}
\usepackage{algorithmic}
\usepackage{booktabs}
\usepackage[dvipsnames]{xcolor}
\usepackage{fancyvrb}
\usepackage{subfigure}
\usepackage{multirow}
\usepackage{geometry}
\usepackage{pdflscape}
\usepackage{longtable}
\usepackage{multicol}
\usepackage{multirow}
\usepackage{supertabular}
\usepackage{latexsym,bm}
\usepackage{mathrsfs}
\usepackage{amsmath}
\usepackage{txfonts}
\usepackage{lineno}
\usepackage{textcomp}
\usepackage[ruled]{algorithm2e}
\usepackage{booktabs}
\usepackage[figuresright]{rotating}
\usepackage{lscape}
\usepackage[pdfstartview=FitH,
CJKbookmarks=true,
bookmarksnumbered=true,
bookmarksopen=true,
colorlinks,
pdfborder=001,  
linkcolor=green,
anchorcolor=green,
citecolor=green
]{hyperref}
\makeatletter

\newcommand{\Rmnum}[1]{\expandafter\@slowromancap\romannumeral #1@}
\makeatother

\journal{ }
\allowdisplaybreaks
\newcolumntype{L}[1]{>{\raggedright\let\newline\\\arraybackslash\hspace{0pt}}m{#1}}
\newcolumntype{C}[1]{>{\centering\let\newline\\\arraybackslash\hspace{0pt}}m{#1}}
\newcolumntype{R}[1]{>{\raggedleft\let\newline\\\arraybackslash\hspace{0pt}}m{#1}}

\begin{document}

\begin{frontmatter}
\title{\textbf{Reproducing human biases in route choice using large language models: Toward scalable behavioral modeling}}	
\author[fn1,fn2]{Jiangtao Han}
\author[fn1,fn2]{Shoufeng Ma}
\author[fn1,fn2]{Shuxian Xu\corref{cor}}
\ead{shuxianxu@tju.edu.cn}
\author[fn1,fn2]{Geng Li}
\author[fn1,fn2]{Shuai Ling}
\author[fn1,fn2]{Ning Jia}
\author[fn3]{Zhengbing He\corref{cor}}
\cortext[cor]{Corresponding author}
\ead{he.zb@hotmail.com}
\address[fn1]{Laboratory of Computation and Analytics of Complex Management Systems (CACMS), Tianjin University, China}
\address[fn2]{College of Management and Economics, Tianjin University, China}
\address[fn3]{Faculty of Science and Engineering, University of Nottingham Ningbo China}

\begin{abstract}
Human choice behavior, including route choice, exhibits systematic behavioral biases that deviate from the assumptions of full rationality. 
Cumulative prospect theory (CPT) has been widely recognized as an effective framework for characterizing such behavioral patterns. 
However, its large-scale application, particularly in simulation and agent-based modeling, critically depends on estimating CPT parameters based on individual-level behavioral data, which remains a major bottleneck.
Conventional approaches typically rely on surveys and controlled experiments to calibrate CPT parameters, yet these methods are difficult to generalize and often fail to capture the full diversity of human decision-making. 
To address this challenge, this paper investigates whether large language models (LLMs) can reproduce human behavioral biases in choice-making without explicit specification of prospect-theoretic parameters. 
Using route choice as a representative scenario, we design a behavioral evaluation framework and systematically compare LLM-generated decisions with established human behavioral patterns predicted by CPT. 
Experimental results demonstrate that LLMs are capable of reproducing non-rational human choice biases and can exhibit decision behaviors consistent with prospect-theoretic effects under uncertainty. 
These findings suggest that generative AI models may provide a scalable alternative for modeling human decision processes and offer a promising foundation for next-generation large-scale agent-based simulation and AI-driven behavioral research.
\vspace{5mm}
\end{abstract}

\end{frontmatter}

\newpage

\section{Introduction}\label{sec:Introduction}



In real-world decision-making, individuals do not always behave in a fully rational and objective manner. Instead, decisions are often shaped by subjective perceptions, resulting in systematic behavioral biases that depart from the assumption of full rationality (\citealp{murphy2018hierarchical}).
Route choice is a typical example of this phenomenon (\citealp{di2014braess}).
For example, travelers usually make route choice decisions under varying traffic conditions, such as different travel times and congestion levels.
Their decisions do not always follow the principle of minimizing travel cost or travel time, but instead exhibit characteristics that deviate from the predictions of fully rational choice models.


Cumulative Prospect Theory (CPT) is arguably the most important and influential descriptive model of risky choice to date (\citealp{barberis2013thirty}, \citealp{fox2015decision}, \citealp{murphy2018hierarchical}).
Unlike classical expected utility theory, CPT can characterize key psychological mechanisms in human decision-making, including reference dependence, loss aversion, diminishing sensitivity, and nonlinear probability weighting.
These mechanisms explain why individuals make asymmetric evaluations and exhibit different risk preferences under different gain and loss scenarios.
Due to its strong behavioral explanatory power, CPT has been widely applied to various areas of travel behavior research, including mode choice (\citealp{zhou2024modeling}), departure time choice (\citealp{geng2023commuter}), parking mode choice (\citealp{hu2025examining}), and electric vehicle charging mode choice (\citealp{zhang2026charging}).


However, the large-scale application of CPT, particularly in simulation and agent-based modeling, heavily depends on estimating CPT parameters based on individual-level behavioral data, which remains a major bottleneck.
\citet{jou2013application} estimated CPT parameters by surveying drivers at freeway rest areas about their route choice outcomes under two sets of freeway experimental scenarios, each involving four different travel times and corresponding occurrence probabilities.
\citet{hu2025examining} collected respondents' parking mode choice data under uncertain conditions, such as waiting time, parking fees, and cruising costs, and estimated CPT parameters to reflect autonomous vehicle users' risk attitudes toward parking choice.
\citet{murphy2018hierarchical} emphasized that the specification of CPT parameters is a key step in quantifying individual preference heterogeneity and uncovering the cognitive mechanisms underlying observed behavior; however, the accurate estimation of these parameters remains challenging.
Conventional approaches typically rely on surveys and controlled experiments to calibrate CPT parameters (\citealp{zhang2018cumulative}, \citealp{geng2023commuter}), which can characterize the risk levels of specific groups but are difficult to generalize and often fail to capture the full diversity of human decision-making (\citealp{thomas2024modelling}, \citealp{bagaini2025systematic}).




The rapid development of large language models (LLMs) provides new technical support for human behavior modeling (\citealp{park2023generative}, \citealp{yan2026valuing}). 
Through natural language understanding, role generation, and contextual reasoning, LLMs are able to characterize the complexity of human behavior. 
Therefore, LLM-based virtual behavioral experiments offer a promising research pathway for reproducing systematic biases in human decision-making, while also helping to alleviate the difficulty of CPT parameter estimation by generating diversified behavioral data. 
However, existing studies on LLM-based behavior modeling still face two key challenges. 
First, whether LLM-generated behaviors can realistically and robustly reproduce human non-rational decision biases under risky conditions remains to be examined. 
Second, constrained by the invocation cost and computational resources of LLMs, virtual behavioral experiments involving large-scale heterogeneous populations remain limited, and the capabilities of LLMs in behavioral data generation, parameter estimation, and theoretical mechanism quantification have not yet been fully validated.

\begin{figure}[htbp]
    \centering    \includegraphics[width=1\textwidth]{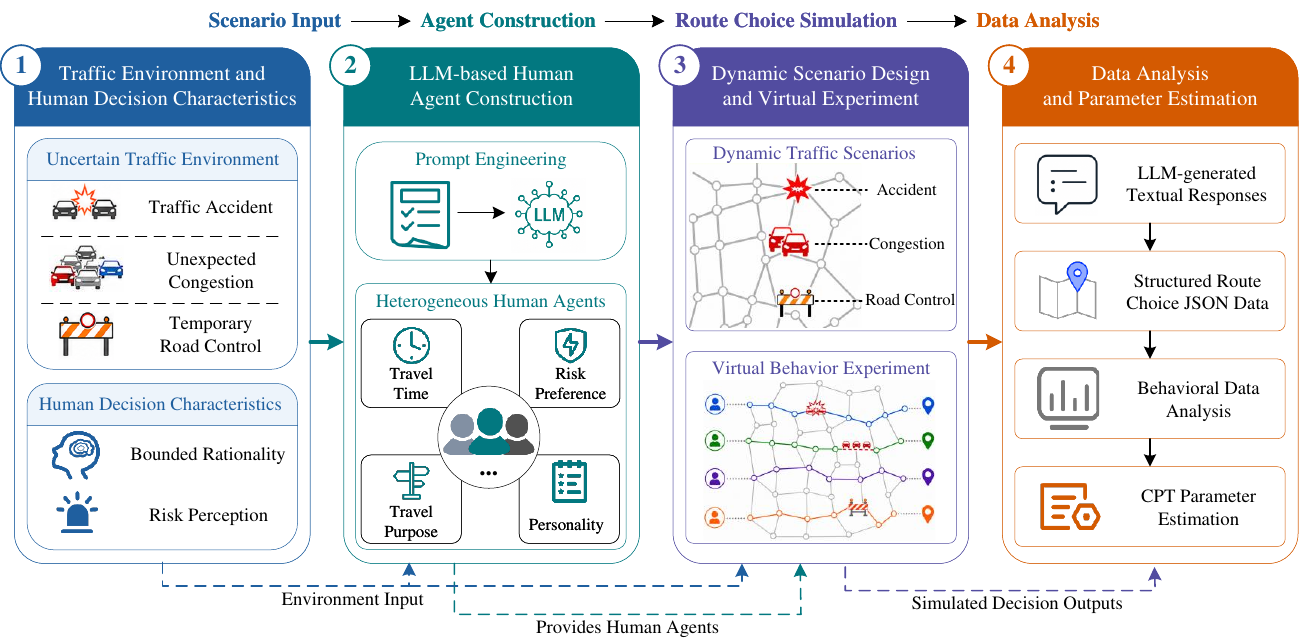} 
    \caption{LLM-based route choice modeling and virtual behavioral experiment framework} 
    \label{fig:LLM-based Route Choice Modeling and Virtual Behavioral Experiment Framework} 
\end{figure}

Taking route choice as a representative scenario of human risky decision-making, this paper proposes an LLM-based method for conducting large-scale virtual behavioral experiments. 
Using this method, we simulate human route-choice behavior under different risky scenarios, confirm that LLMs can reproduce systematic biases in human choice behavior, and further explore their potential for CPT parameter estimation. 
More specifically, as shown in Figure \ref{fig:LLM-based Route Choice Modeling and Virtual Behavioral Experiment Framework}, heterogeneous human agents with multidimensional attributes are constructed based on LLMs through standardized prompts, and dynamic traffic environments with different gain and loss conditions are designed to conduct experimental simulations under various risk settings. 
The behaviors generated by LLMs are then transformed into structured data suitable for statistical analysis and model calibration, enabling the estimation of prospect utility parameters and the inference of reference points. 
Finally, the proposed method is evaluated at both the individual and environmental levels, and its accuracy in parameter estimation and route-choice prediction is compared with that of traditional parameter estimation methods. 
Experimental results demonstrate that LLMs are capable of reproducing non-rational human choice biases and exhibit decision behaviors consistent with prospect-theoretic effects under uncertainty. 

The significance of this study lies not only in using LLMs to simulate human route-choice behavior at scale, but also in demonstrating their potential to reproduce systematic biases in human decision-making. 
The findings suggest that LLM-based agents are able to exhibit subjective perception and boundedly rational choice patterns similar to those of real humans. 
Therefore, generative AI models may provide a scalable alternative for modeling human decision processes and offer a promising foundation for next-generation large-scale agent-based simulation and AI-driven behavioral research.


The remainder of this paper is organized as follows. 
Section \ref{sec:literature review} reviews the relevant literature. 
Section \ref{Preliminaries} introduces the mechanisms of LLMs.
Section \ref{sec:Methodology} systematically develops the LLM-based route choice simulation framework and proposes methods for determining prospect utility parameters and reference points. 
Section \ref{sec:experiments} conducts LLM-based experimental simulations under different risk scenarios, estimates the CPT parameters based on the simulation results, evaluates the simulation effects, and compares them with other parameter estimation approaches. 
Finally, Section \ref{sec:conclusion} concludes this study and suggests directions for future research.

\section{Literature review}\label{sec:literature review}
\subsection{CPT-based route choice modeling}\label{Route choice and CPT research}

Research on human route choice is one of the core topics in traffic behavior analysis and transportation network modeling, and it also serves as a crucial behavioral basis for traffic assignment, congestion management, and travel information services (\citealp{yang2009day}, \citealp{qi2023investigating,Qi2024RouteDependent}). 
For a long time, mainstream research in this field has been based on expected utility theory (EUT) and the assumption of full rationality (\citealp{zhang1996local}, \citealp{smith1984stability}). 
However, with the continuous observation of real travel behavior and the accumulation of empirical evidence, the traditional EUT framework has gradually revealed its limitations in explaining route choice behavior under uncertain conditions (\citealp{wu2013bounded}, \citealp{henn2006handling}). 

Theoretical tools from behavioral economics and psychology have gradually been introduced into travel behavior research to explain human decision-making behavior under uncertain conditions from the perspective of bounded rationality. 
Among these, prospect theory proposed by \citet{kai1979prospect} and its subsequent development, CPT (\citealp{tversky1992advances}), provide an important theoretical foundation for explaining systematic deviations from rational behavior under uncertain conditions. 
CPT emphasizes that: (1) individuals evaluate gains and losses relative to reference points, and decision-making focuses on relative changes rather than absolute levels; (2) individuals tend to exhibit risk aversion in the domain of gains and risk seeking in the domain of losses, with losses having a greater psychological impact than equivalent gains; (3) individuals perceive probabilities in a nonlinear manner, overweighting low-probability events while underweighting medium- and high-probability events.

Route choice modeling based on CPT typically characterizes behavior through two types of core functions. 
The first is the value function, which describes humans' nonlinear evaluation of gains and losses relative to a reference point. 
The second is the probability weighting function, which describes humans' subjective perception of the probability of different travel outcomes. 
Existing research has validated and extended the CPT framework across different dimensions of travel behavior. 
For example, \citet{jou2013application} modeled the perception of gains and losses of freeway drivers using CPT by considering their risk attitudes and preferences, and applied it to the route choice problem of drivers on freeways. 
\citet{ghader2019modeling} introduced a CPT-based travel mode choice modeling framework that uses observational data (household travel surveys) to help travel demand modelers model reliability using CPT in their trip-based or activity-based models. 
\citet{zhang2018cumulative} collected statistical data on the actual route choice decisions of subjects by designing and conducting laboratory behavioral experiments. 
Based on CPT, they constructed a day-to-day route choice learning model that included friends' travel information and explored the effect of social interaction information from friends on commuters' daily route choice decisions. 
Furthermore, other studies (\citealp{geng2023commuter}, \citealp{hu2025examining}, \citealp{zhang2026charging}) have validated and extended the CPT framework across different dimensions of travel behavior, demonstrating that CPT plays a crucial role in human behavior modeling.

When CPT is used to characterize human behavior, the prospect utility parameters reflect humans' risk attitudes and value preferences. 
Their reasonable setting and accurate estimation not only determine whether the model can accurately capture human value evaluation and risk perception mechanisms but also directly affect the model's predictive ability for human decision-making behavior under uncertain conditions. 
Therefore, determining the prospect utility parameters is crucial for improving the model's explanatory and predictive power. 
\citet{xu2011decision} designed two types of travel scenarios: time gain and loss scenarios. 
Each scenario consists of two routes with equal average travel times but different gains or losses. 
They obtained the CPT parameter estimation results by conducting a questionnaire survey among part-time postgraduate students and fitting the collected human route choice preference data. 
\citet{ghader2019modeling} conducted a travel behavior survey of residents in the Washington area, obtained 160 available data records, analyzed the data characteristics of both rail and driving travel, and estimated the parameters of the CPT model using the maximum likelihood estimation method. 
Regarding the determination of reference points, \citet{yang2014development} improved the method of setting reference points in the route choice model by replacing free-flow travel time with actual budgeted travel time. 
\citet{zhang2018cumulative} estimated reference points by considering the risk level of the participants through questionnaire surveys and combining the cumulative distribution function of the participants' travel time.

Overall, the existing studies based on questionnaire surveys or behavioral experiments provide valuable insights and methods for estimating CPT parameters, and can realistically depict human basic risk preferences to a certain extent. 
However, these methods are often constrained by limited sample sizes and homogeneous participant groups, making it difficult to fully reflect the behavioral differences caused by different occupational types, personality traits, travel purposes, and risk attitudes. 

\subsection{Large language models}\label{Large language models}

Benefiting from large-scale parameterization and pretraining on massive datas, recent LLMs have demonstrated capabilities approaching human-level intelligence in language understanding, knowledge reasoning, and generative interaction (\citealp{hua2023war}, \citealp{chen2024agentverse}). 
Compared with traditional rule-based or small-sample learning intelligent systems, LLMs exhibit stronger task generalization and programmability, enabling researchers to trigger various behaviors such as information extraction, situational decision-making, and multi-turn dialogue—through designed prompts. 
This provides new technical support for behavior simulation driven by natural language. With the integration of memory mechanisms (\citealp{sumers2023cognitive}), tool invocation (\citealp{schick2023toolformer}), and planning modules (\citealp{liu2023bolaa}), LLMs are no longer limited to text generation but are gradually developing into closed-loop intelligent agents with capabilities for perception, memory, planning, and action, offering a novel pathway for modeling individual behavior in complex social systems.

Against this backdrop, research on generative agents has provided a new paradigm for simulating human behavior. 
\citet{park2023generative} from Stanford University first systematically proposed the concept of generative agents and instantiated a group of agents capable of perceiving the environment, forming memories, planning, and interacting with others in an interactive sandbox environment. 
Their results show that LLM-based agents can not only generate daily behavior trajectories consistent with individual human characteristics, but also exhibit social phenomena such as information diffusion, social aggregation and collaborative behavior in group interactions, providing architectural design and interaction paradigms for achieving believable human behavior simulation. 
Subsequently, a series of follow-up studies have extended this model to other application scenarios, including social interaction (\citealp{zhou2023sotopia}), game reasoning and decision-making (\citealp{xu2023exploring}), and software development (\citealp{qian2024chatdev}). 
At the same time, multimodal generation technologies (\citealp{wang2023neural}, \citealp{nguyen2022deep}) such as speech, images, and video are also accelerating, making it possible to present behaviors and interactions closer to the real world. 
In short, LLM-driven agent simulation methods are rapidly becoming an important tool for simulating complex human behavior.

Although LLMs have been applied in many fields, research on their use for simulating human risky decision-making behavior remains relatively limited. 
In particular, whether LLMs can exhibit systematic biases similar to those observed in humans and demonstrate decision-making behavior consistent with CPT under uncertainty has not been fully verified. 
Existing studies mostly remain at the level of small-scale scenario testing, making it difficult to reveal behavioral differences and emergent characteristics among different types of individuals at the group level. 
Meanwhile, there is still a lack of systematic methodological frameworks and operational procedures for designing standardized prompts, constructing representative uncertain risk scenarios, and generating large-scale, reproducible, and statistically analyzable human-like behavioral data.


\section{Preliminaries}\label{Preliminaries}

LLMs are a powerful class of natural language processing systems that are constructed on the transformer architecture (\citealp{vaswani2017attention}) and trained using in-context learning (ICL) techniques (\citealp{brown2020language}, \citealp{akyurek2022learning}). The core of the transformer is scaled dot-product attention, which is defined as: 
    \begin{equation}
    {\rm{Attention}}(A,K,V) = {\rm{softmax}}\left(\frac{{Q{K^T}}}{{\sqrt {{d_k}} }}\right)*V
    \label{eq: 1}
    \end{equation}
where $Q$ is the query, $K$ is the key, $d_k$ is the dimension, and $V$ is the value. The attention mechanism applies the softmax function to obtain the weight of the value by calculating the dot product of the query and the key, and finally outputs the result with the largest weight. ICL allows language models to learn tasks with only a few example in the form of demonstrations (\citealp{dong2024survey}). Formally, given a query input text $a$ and a candidate answer set $b = [{b_1},...,{b_n}]$, the possibility of the candidate answer $b_j$ is calculated by the scoring function $f$ of the thole input sequence: 
    \begin{equation}
    P({b_j}|a) \buildrel \Delta \over = {f_M}({b_j},C,a)
    \label{eq: 2}
    \end{equation}

The pre-trained language model $M$ takes the candidate answer with the maximum score as the prediction result, and the final prediction label $\widehat b$ is the candidate answer with the highest probability:
    \begin{equation}
    \widehat b = \arg {\max _{{b_j} \in b}}P({b_j}|a)
    \label{eq: 3}
    \end{equation}

This mechanism enables the model to perform task-specific behaviours directly from prompts and demonstrations without additional parameter updates (\citealp{liu2023pre}). In recent years, LLM-based agents have demonstrated remarkable human-like capabilities in reasoning and planning (\citealp{park2023generative}, \citealp{yao2023tree}, \citealp{wang2024survey}). These agents constitute systems that possess a certain degree of autonomy, goal-directed behavior, responsiveness, and interactive ability, enabling them to perceive the environment, make decisions, and take actions independently. In the autonomous agent systems powered by LLMs, the agent brain is composed of several key components, including memory, planning, and action (\citealp{weng2023agent}).

\textbf{Memory module.} This module enables the agent to acquire, store, and retain information, thereby supporting its subsequent actions. Memory is typically categorized into sensory memory, long-term memory, and short-term memory.

\textbf{Planning module.} This module assists the agent in breaking down complex tasks into smaller, more manageable subtasks.

\textbf{Action module.} This module functions as the agent’s decision executor, responsible for generating the specific actions the agent performs.


\section{Methodology}\label{sec:Methodology}


\subsection{Construction of LLM-based human agents}\label{Construction method of LLM-based traveler agent}

This study adopts a nonparametric prompting method to construct LLM-based human agents, assigning roles with specific demographic characteristics to the model. 
The agents are able to act according to predefined identities and roles, based on their own ideas and feelings. 
In order to realize the simulation, this paper develops a standardized prompting architecture that can accurately describe human identity characteristics and capture the complexity of human behavior. 
This architecture is modularized into three core components: profile, planning, and action.

\textbf{Profile.} The human agent's individual profile is constructed to distinguish the unique characteristics of different simulated individuals. 
The profile consists of three components: individual attributes, behavior features, and constraint conditions. 
The individual attributes include occupation, risk preference, travel purpose, and time pressure, which are used to describe the agent's basic identity. 
The behavior features capture the agent's preferences for expected travel time and time volatility of different routes, tolerance intervals for early and late arrivals, and sensitivity to risks such as traffic accidents, unexpected congestion or temporary road control measures, which can be understood as the agent's behavioral parameters and psychological preferences. 
The constraint conditions are used to limit the set of feasible travel plans, including mandatory time constraints, personal ability or characteristic constraints, and environmental constraints.

\textbf{Planning.} The planning ability of the human agent is designed to enable it to decompose and plan complex tasks under a specific identity and scenario.
Through prompting, the travel purpose and risk preference are clarified, guiding the agent to identify key decision-making factors, such as travel time and time variability. 
On this basis, the agent evaluates each route in terms of expected time, variability, and risk according to its travel objectives, forms subjective assessments and prioritization across routes, and ultimately generates a travel plan consistent with its identity and psychological preferences.

\textbf{Action.} The decision rules and output formats are established to ensure that the agent can make route choice decisions that reflect its individual characteristics within a given scenario. 
The decision rules specify how the agent should weigh different route attributes and how risk attitudes influence choice preferences. 
For example, risk-averse agents are expected to prioritize routes with lower time variability, whereas risk-seeking agents tend to prefer routes with shorter travel times but higher variability. 
To ensure that the output is controllable and suitable for statistical analysis, the LLMs are required to produce strictly structured JSON outputs.

It is important to note that the constraints in the agent profile define the objective limitations related to occupation, personality traits, travel purpose, and environmental factors, thereby determining the feasible set of travel options. 
In contrast, the decision rules describe how the agent weighs attributes such as travel time, time variability, and risk within this feasible set and constitute the behavioral mechanism that drives the agent to generate specific route choice decisions.

\subsection{Construction of LLM-based route choice travel environment}\label{Prompt Engineering}

Travel environments define the specific context in which agents operate. 
Similar to how humans acquire information from their surroundings, agents rely on the environment to receive input signals from multiple sources, which guide their behaviors and decision-making strategies within the system. 
A comprehensive understanding of the environment is a prerequisite for agents to make route choice decisions and complete travel tasks. 
This study constructs LLM-based route choice travel environments from two dimensions of configuration and state.

\textbf{Configuration.} The configuration defines the agent's travel task and the fundamental structure of the environment. 
It provides static or relatively stable information, including task objectives, the set of optional routes, and basic route attributes. 
Travel task constraints establish the agent's travel time requirements and the consequences of arriving late or early. 
The set of optional routes and basic route attributes describe the available route options, normal travel times and potential risk types for each route. 
The configuration provides the basic input for agent decision-making, enabling the agent to understand the structural characteristics of the environment.

\textbf{State.} The state represents the observable dynamic information provided by the environment during task execution, which directly affects the agent's decisions and behaviors. 
This dynamic information includes detailed road conditions, such as the occurrence probabilities of different risks and the impact of each risk on travel time. 
The agent observes the current state and its dynamic changes, plans travel strategies, and generates travel decisions.

\begin{figure}[H]
    \centering
    \includegraphics[width=1\textwidth]{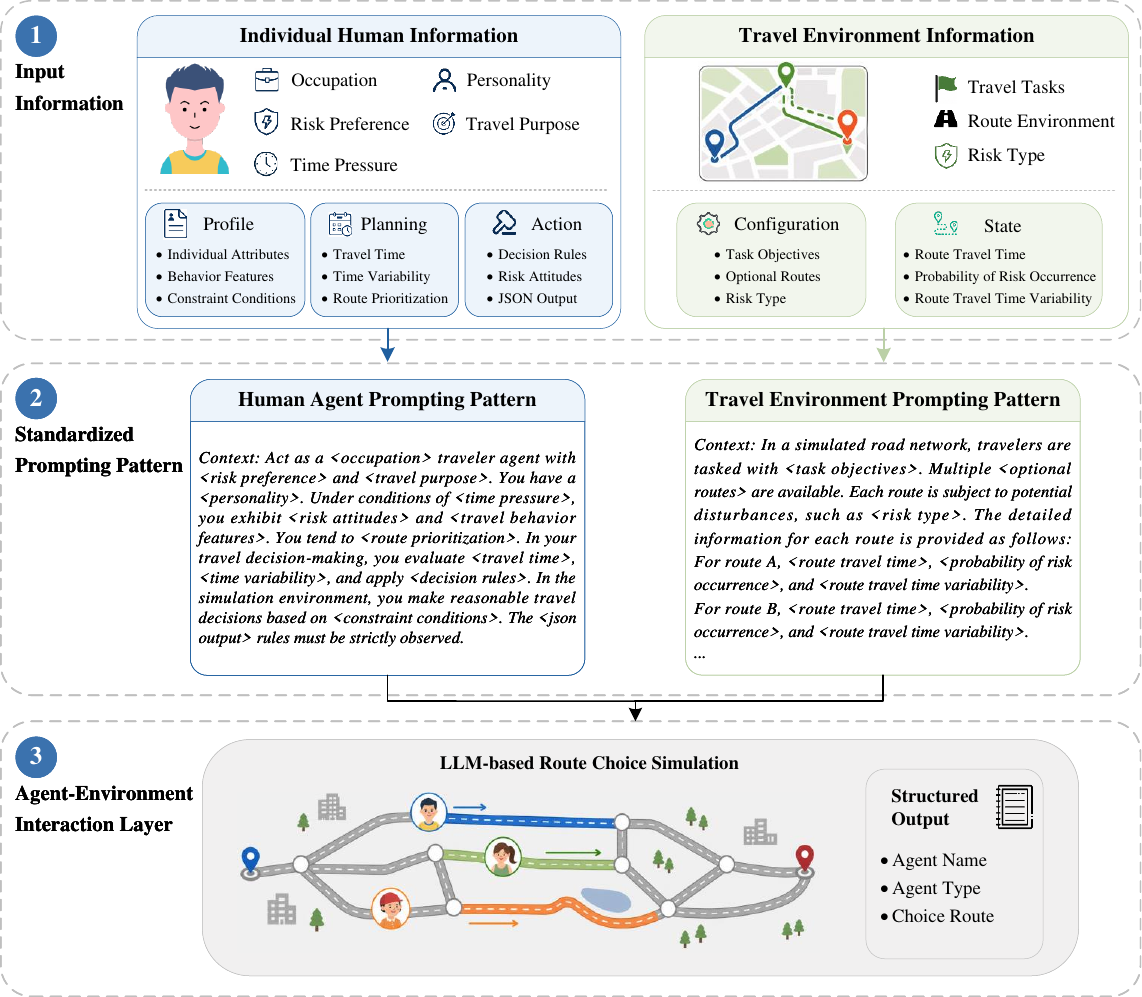} 
    \caption{Standardized prompt generation process based on agent and environmental information} 
    \label{fig:Prompting pattern} 
\end{figure}

Based on the nonparametric prompting method, the standardized prompting pattern is constructed, as shown in Figure \ref{fig:Prompting pattern}. The pseudo-code of the LLM-based route choice simulation framework is presented in Algorithm \ref{alg:LLM-based route choice simulation framework}.

\begin{algorithm}[htbp]
\LinesNumbered
\caption{LLM-based route choice simulation framework}
\label{alg:LLM-based route choice simulation framework}
\KwIn{
Travel environment information (configuration, state)  \\
Human agent information (profile, planning, action)
}
\KwOut{
Human agents' route choice simulation results in JSON format
}
Configure the OpenAI gpt-5 API (API key and model name)\;
Construct the prompt containing travel environment information $e_j \in \mathcal{E}$\;
Construct the prompt containing human agent information $a_i \in \mathcal{A}$\;
\For{$i = 1$ \KwTo ${N_a}$}{
    Match agent prompt $a_i$ with environment prompt $e_j$\;
    $n{p_j} = {e_j} \cup \{ {a_i}|i = 1,...,N_a\}$\;
    $NP = \{ n{p_j}|j = 1,...,{N_e}\}$
}
Invoke the $gpt-5$ LLM\;
return $json.loads (results)$\;
Print $name(Agent\_i), type, route$
\end{algorithm}

\subsection{Estimation of cumulative prospect theory parameters and reference point determination}\label{Estimation of cumulative prospect theory parameters and reference point determination}

Building on the prospect theory, \citet{tversky1992advances} further developed CPT in 1992. 
CPT evaluates the cumulative prospect value $PV(x)$ of alternative options using value function $v(x)$ and cumulative decision weights $\pi (p)$. 
The gains and losses in the value function are relative to a reference point $x_0$. 
When faced with gains, people tend to avoid risk; when faced with losses, they tend to pursue risk. 
Furthermore, people's aversion to losses is often greater than their preference for the same gains; that is, the value function is concave in the gain part and convex in the loss part.
In this study, the value function is expressed as follows:
    \begin{equation}
    v(x) = \left\{ {\begin{array}{*{20}{c}}
{{{(x - {x_0})}^\alpha }}&{x \ge {x_0}}\\
{ - \lambda {{({x_0} - x)}^\beta }}&{x < {x_0}}
\end{array}} \right.
    \label{eq: 4}
    \end{equation}
where $0 < \alpha , \beta  < 1$ represents the decision maker’s sensitivity to gains and losses, while $\lambda$ denotes the risk-aversion coefficient. 
Typically, $\lambda \ge 1$, indicating that the decision maker exhibits a risk-averse attitude, and that the value function is steeper in the loss domain. 
The cumulative decision weights are defined as follows: 
    \begin{equation}
{\pi ^ + }({p_i}) = w({p_i} + ... + {p_n}) - w({p_{i + 1}} + ... + {p_n}),0 \le i \le n
    \label{eq: 5}
    \end{equation}
    \begin{equation}
{\pi ^ - }({p_{ - j}}) = w({p_{ - m}} + ... + {p_{ - j}}) - w({p_{ - m}} + ... + {p_{ - j - 1}}),m \le  - j < 0
    \label{eq: 6}
    \end{equation}
where $w(p)$ is the probability weighting function, and its expression is:
    \begin{equation}
w(p) = \frac{{{p^\gamma }}}{{{{({p^\gamma } + {{(1 - p)}^\gamma })}^{1/\gamma }}}}
    \label{eq: 7}
    \end{equation}
where $0 < \gamma  < 1$ represents people tend to overweight rare events with extremely small probabilities while underweighting more common events with moderate or high probabilities. 
The weighting function is usually inversely S-shaped.
The cumulative prospect value is expressed as follows:
    \begin{equation}
PV(x) = \sum\nolimits_{i = 0}^n {v({x_i}){\pi ^ + }({p_i})}  + \sum\nolimits_{i =  - m}^{ - 1} {v({x_i}){\pi ^ - }({p_i})}
    \label{eq: 8}
    \end{equation}


In this study, the prospect utility parameters are estimated by fitting the data generated from LLM-based agents simulating route choice preferences to the formulations of CPT. 
The specific form of the value function is associated with the probability weighting function, which characterizes individuals’ subjective perception of probabilities. 
Since decision weighting functions in different domains have the same type of independent variables, the decision weighting parameter exhibits a degree of universality across different economic sectors, this study uses the parameter estimate from \citet{wu1996curvature} for a general economic sector, $\gamma=0.74$. 
This study only estimates parameters $\alpha$, $\beta$, and $\lambda$ using LLMs simulation results.

In the route choice experiments under gain and loss scenarios, it is assumed that a agent receives a gain of $\Delta x_i^ +$ with probability $p_i$, and incurs a loss of $\Delta x_{ - i}^ -$ with probability $p_{ - i}$. 
Let $P({A_j})$ denote the proportion of agents choosing route $A$ in case $j$, and $U({A_j})$ represent the cumulative prospect value of route $A$ in case $j$. 
Moreover, $P({A_j} > {B_j}) = 1/(1 + {e^{U({B_j}) - U({A_j})}})$ indicates the probability that route $A$ is perceived as more valuable than route $B$ in case $j$. 
Therefore, under the gain case, the sum of squared residuals between the LLM-simulated route-choice outcomes and those computed from CPT can be expressed as:
    \begin{equation}
    {f_1}(\alpha ,\gamma ) = \sum\nolimits_{j = 1}^{{N_ + }} {{{(P({A_J}) - P({A_j} > {B_j}))}^2}} 
    \label{eq: 9}
    \end{equation}
where $N_+$ represents the number of gain cases. 
Under the loss case, the sum of squared residuals for agents' route choices can be expressed as:
    \begin{equation}
    {f_2}(\beta ,\lambda ,\gamma ) = {\sum\nolimits_{j = 1}^{{N_ - }} {(P({A_j}) - P({A_j} > {B_j}))} ^2}
    \label{eq: 10}
    \end{equation}
where $N_-$ represents the number of loss cases. 
In CPT, the reference point serves as the psychological benchmark against which gains and losses are evaluated. 
For human decision-making under uncertain travel times, the reference point reflects the expected level of travel time. 
In this study, the reference point is inferred from the distribution of expected travel times generated by LLM-based agents. 
During the simulation experiment, each agent forms an individualized expectation of travel time based on the case information presented. 
Let $N$ denote the total number of agents, and let $t_i$ represent the expected travel time of agent $i$. 
The aggregation of these expectations provides an empirical basis for determining a case reference point. 
Formally, the reference point $x_0$ under a given case is computed as:
    \begin{equation}
    {x_0} = \frac{1}{N}\sum\nolimits_{i = 1}^N {{t_i}}
    \label{eq: 11}
    \end{equation}

This formulation captures the average tendency of agents’ expectations, thereby ensuring that the subsequent gain–loss evaluation is grounded in the behavioral patterns revealed by the simulated agents. 
Unlike traditional methods that rely on fixed or deterministic reference points, the LLM-based approach for inferring reference points can more realistically reflect the psychological processes of human agents.

\section{Route choice simulation experiments with LLM-based human agents}\label{sec:experiments}

\subsection{Construction of LLM-based human traveler agents}\label{Construction of LLM-based traveler agents}
When facing different real-world problems, individuals' risk attitudes often differ; however, at the group level, risk-neutral and risk-averse individuals tend to dominate (\citealp{holt2002risk}). 
To realistically simulate human behavior under uncertain travel conditions, this study references the agent settings in \citet{wang2025agentic} and \citet{wang2024large}, defines ten distinct human profiles, and generates 300 agents for each profile, resulting in a total of 3,000 LLM-based agents.
Their profile information, as shown in Table \ref{tab: LLM Agent profile information}, covers several key demographic and behavioral heterogeneity characteristics, including personality, occupation, travel purpose, and risk preference. 
Using the standardized prompting pattern proposed in section \ref{Construction method of LLM-based traveler agent}, each agent's individual profile and planning capability are defined, enabling it to operate as an independent autonomous decision-making entity.
The detailed prompt contents are provided in \ref{sec:Prompts for constructing individual traveler agents}.

\begin{table}[!h]
    \footnotesize
    \centering
    \caption{LLM-based human agents profile information}
    \label{tab: LLM Agent profile information}
    \begin{tabular}{
    >{\raggedright\arraybackslash}m{0.5cm}
    >{\raggedright\arraybackslash}m{2.2cm}
    >{\raggedright\arraybackslash}m{2.8cm}
    >{\raggedright\arraybackslash}m{5.7cm}
    >{\raggedright\arraybackslash}m{2.2cm}
    }
    \toprule
    Type & Occupation & Personality & Risk Preference & Travel Purpose\\
    \midrule
    1 & Retired worker & Cautious and meticulous & Averse to time variability; prefers routes with stable travel times & Going to the hospital \\ 2 & Civil servant & Composed and punctual & Values stability in travel time and exhibits high sensitivity to temporal fluctuations & Daily commuting to work \\ 3 & Courier & Diligent and pragmatic & Prefer the route with stable average time, but can  tolerate small range of time fluctuations & Commuting for deliveries \\ 4 & Parent & Responsible and meticulous & Prefers routes with stable average travel times, but may choose shorter routes with greater variability& Take children to and from school \\ 5 & Business manager &  Steady and pragmatic & Values route stability and pursues the optimal time options within the acceptable risk range & Attending a business meeting \\ 6 & Retirees & Peaceful and patient & Not concerned with saving time & Leisure and recreation \\ 7 & Teacher & Efficient and rational & Tends to balance average travel time and variability & Commuting to work \\ 8 & Designer & Independent and composed & Tends to balance average travel time and variability & Temporary outing \\ 9 & College student & Outgoing and spontaneous & Values time savings more than reliability; can accept occasional delays & Leisure and entertainment \\ 10 & Vlogger & Outgoing and curious & Actively embraces uncertainty in exchange for time-saving routes & Visiting and reviewing locations \\
    \bottomrule
    \end{tabular}
\end{table}

\subsection{Experiment settings}\label{Experiment settings}

The overall experimental design is consistent with the core behavioral assumptions of CPT (\citealp{xu2011decision}). 
First, reference points are constructed to capture agents' concave preferences in the gain domain and convex preferences in the loss domain. 
Second, the scenario settings introduce typical traffic disruptions, such as accidents and unexpected congestion, and quantify their impacts as travel-time increments of different magnitudes. 
This design improves the comprehensibility and realism of the scenarios, enabling agents to make judgments based on real commuting experiences. 
Next, in each case, the routes are designed to have the same mean travel time, while the probability of each risk outcome is systematically adjusted to highlight differences in agents' subjective perceptions of time uncertainty and risk (\citealp{zhang2018cumulative}). 
Finally, to balance clarity of information presentation and cognitive burden, each route's travel-time distribution is represented by four discrete outcomes, which not only clearly characterize different levels of disturbance risk but also improve the stability and identifiability of parameter estimates (\citealp{xu2011decision}).

\textbf{Scenario 1:}
In a simulated road network, agents are assumed to face a commuting trip that must be completed within 45 minutes. 
Two alternative routes are available, both having the same average travel time but subject to potential disruptions such as unexpected congestion, traffic accidents, or temporary road control measures. 
For route A, unexpected congestion caused by vehicle breakdowns may add about 5 minutes; traffic accidents such as vehicle rear-end collision will increase the travel time by approximately 10 minutes; and temporary road control measures may result in an additional delay of around 15 minutes. 
For route B, unexpected congestion may cause an additional 10-minute delay; traffic accidents may increase travel time by approximately 20 minutes; and temporary traffic control measures may increase travel time by about 25 minutes. 
It is assumed that under all conditions, agents can still arrive within the 45-minute limit; however, earlier arrivals yield higher travel rewards. 
In the gain scenario, the occurrence probabilities of the risk events on each route are shown in Table \ref{tab: Travel time distribution for each route in the gain scenario}.

\begin{table}[!h]
    \footnotesize
    \centering
    \caption{Travel time and gain distribution for each route}
    \label{tab: Travel time distribution for each route in the gain scenario}
    \begin{tabular}{
    >{\raggedright\arraybackslash}m{1.5cm}
    >{\raggedright\arraybackslash}m{6.4cm}
    >{\raggedright\arraybackslash}m{6.4cm}
    }
    \toprule
    Case & Travel time and gain for route A & Travel time and gain for route B \\
    \midrule
    \multirow{2}{*}{1} 
    & Time: 30: 20\%, 35: 20\%, 40: 60\%, 45: 0\% & Time: 20: 10\%, 30: 20\%, 40: 50\%, 45: 20\% \\
    & Gain: 15: 20\%, 10: 20\%, 5: 60\%, 0: 0\% & Gain: 25: 10\%, 15: 20\%, 5: 50\%, 0: 20\% \\
    \multirow{2}{*}{2} 
    & Time: 30: 30\%, 35: 30\%, 40: 40\%, 45: 0\% & Time: 20: 20\%, 30: 20\%, 40: 30\%, 45: 30\% \\
    & Gain: 15: 30\%, 10: 30\%, 5: 40\%, 0: 0\% & Gain: 25: 20\%, 15: 20\%, 5: 30\%, 0: 30\% \\ 
    \multirow{2}{*}{3} 
    & Time: 30: 40\%, 35: 30\%, 40: 20\%, 45: 10\% & Time: 20: 25\%, 30: 10\%, 40: 45\%, 45: 20\% \\
    & Gain: 15: 40\%, 10: 30\%, 5: 20\%, 0: 10\% & Gain: 25: 25\%, 15: 10\%, 5: 45\%, 0: 20\% \\
    \multirow{2}{*}{4} 
    & Time: 30: 50\%, 35: 30\%, 40: 20\%, 45: 0\% & Time: 20: 30\%, 30: 20\%, 40: 20\%, 45: 30\% \\
    & Gain: 15: 50\%, 10: 30\%, 5: 20\%, 0: 0\% & Gain: 25: 30\%, 15: 20\%, 5: 20\%, 0: 30\% \\
    \multirow{2}{*}{5} 
    & Time: 30: 70\%, 35: 20\%, 40: 10\%, 45: 0\% & Time: 20: 50\%, 30: 0\%, 40: 10\%, 45: 40\% \\
    & Gain: 15: 70\%, 10: 20\%, 5: 10\%, 0: 0\% & Gain: 25: 50\%, 15: 0\%, 5: 10\%, 0: 40\% \\   
    \bottomrule
    \end{tabular}
\end{table}

\textbf{Scenario 2:}
In a simulated road network, agents are assumed to face a commuting trip that must be completed within 30 minutes. Two alternative routes are available, both having the same average travel time but subject to potential disruptions such as unexpected congestion, traffic accidents, or temporary road control measures. 
For routes A and B, unexpected congestion caused by vehicle breakdowns may add about 5 minutes; traffic accidents such as vehicle rear-end collision will increase the travel time by approximately 10 minutes; and temporary road control measures may result in an additional delay of around 15 minutes. 
It is assumed that neither route can reach the destination within 30 minutes (agents can only be on time or late), and the later the arrival, the higher the loss incurred. 
In the loss scenario, the occurrence probabilities of the risk events on each route are shown in Table \ref{tab: Travel time distribution for each route in the loss scenario}.

\begin{table}[!h]
    \footnotesize
    \centering
    \caption{Travel time and loss distribution for each route}
    \label{tab: Travel time distribution for each route in the loss scenario}
    \begin{tabular}{
    >{\raggedright\arraybackslash}m{1.5cm}
    >{\raggedright\arraybackslash}m{6.4cm}
    >{\raggedright\arraybackslash}m{6.4cm}
    }
    \toprule
    Case & Travel time and loss for route A & Travel time and loss for route B \\
    \midrule
    \multirow{2}{*}{6} 
    & Time: 30: 70\%, 35: 20\%, 40: 10\%, 45: 0\% & Time: 30: 80\%, 35: 10\%, 40: 0\%, 45: 10\% \\
    & Loss: 0: 70\%, 5: 20\%, 10: 10\%, 15: 0\% & Loss: 0: 80\%, 5: 10\%, 10: 0\%, 15: 10\% \\ 
    \multirow{2}{*}{7} 
    & Time: 30: 50\%, 35: 30\%, 40: 20\%, 45: 0\% & Time: 30: 70\%, 35: 0\%, 40: 20\%, 45: 10\% \\
    & Loss: 0: 50\%, 5: 30\%, 10: 20\%, 15: 0\% & Loss: 0: 70\%, 5: 0\%, 10: 20\%, 15: 10\% \\  
    \multirow{2}{*}{8} 
    & Time: 30: 30\%, 35: 50\%, 40: 15\%, 45: 5\% & Time: 30: 55\%, 35: 5\%, 40: 30\%, 45: 10\% \\
    & Loss: 0: 30\%, 5: 50\%, 10: 15\%, 15: 5\% & Loss: 0: 55\%, 5: 5\%, 10: 30\%, 15: 10\% \\ 
    \multirow{2}{*}{9} 
    & Time: 30: 15\%, 35: 60\%, 40: 20\%, 45: 5\% & Time: 30: 45\%, 35: 5\%, 40: 40\%, 45: 10\% \\
    & Loss: 0: 15\%, 5: 60\%, 10: 20\%, 15: 5\% & Loss: 0: 45\%, 5: 5\%, 10: 40\%, 15: 10\% \\ 
    \multirow{2}{*}{10} 
    & Time: 30: 0\%, 35: 70\%, 40: 20\%, 45: 10\% & Time: 30: 30\%, 35: 10\%, 40: 50\%, 45: 10\% \\
    & Loss: 0: 0\%, 5: 70\%, 10: 20\%, 15: 10\% & Loss: 0: 30\%, 5: 10\%, 10: 50\%, 15: 10\% \\
    \bottomrule
    \end{tabular}
\end{table}

In scenarios 1 and 2, the 3,000 human traveler agents are required to independently make route choice decisions under 10 different combinations of risk probabilities, and to output their name, agent type, and chosen route.

\textbf{Scenario 3:}
In a simulated road network, agents are assumed to face a commuting trip and have three alternative routes. 
The average travel time of the three routes is 30 minutes, but there may be unexpected congestion, traffic accidents or temporary road control measures, and the probability of various emergencies in each route is not equal. 
The occurrence probabilities of the risk events on the three routes are shown in Table \ref{tab: Travel time distribution for each route in scenario 3}. 
The 3,000 human traveler agents are required to independently make a route choice decision among the three routes and output their name, agent type, expected travel time, and selected route.

\begin{table}[!h]
    \footnotesize
    \centering
    \caption{Travel time distribution for each route in scenario 3}
    \label{tab: Travel time distribution for each route in scenario 3}
    \begin{tabular}{
    >{\raggedright\arraybackslash}m{6cm}
    >{\raggedright\arraybackslash}m{8.8cm}
    }
    \toprule
    Case & Travel time distribution for each route \\
    \midrule
    \multirow{3}{*}{11} 
    & Route A: 12: 15\%, 24: 35\%, 36: 35\%, 48: 15\%  \\ & Route B: 18: 20\%, 26: 30\%, 34: 30\%, 42: 20\% \\ & Route C: 24: 25\%, 28: 25\%, 32: 25\%, 36: 25\% \\ 
    \bottomrule
    \end{tabular}
\end{table}

Additionally, this study used OpenAI's GPT-5 as the simulation generator. 
The model's temperature parameter is set to 1 by default, allowing the generated decisions to retain an appropriate level of behavioral diversity while avoiding excessive randomness. 
This setting is consistent with previous behavioral simulation studies based on LLMs (\citealp{brown2020language}, \citealp{aher2023using}, \citealp{mou2025agentsense}). 
To ensure the robustness and reproducibility of the experimental results, each experimental scenario is independently executed three times.

\subsection{Experiment results}\label{Experiment results}

\begin{figure}[H]
    \centering
    
    \subfigure[First simulation result]{    \includegraphics[width=0.9\textwidth]    {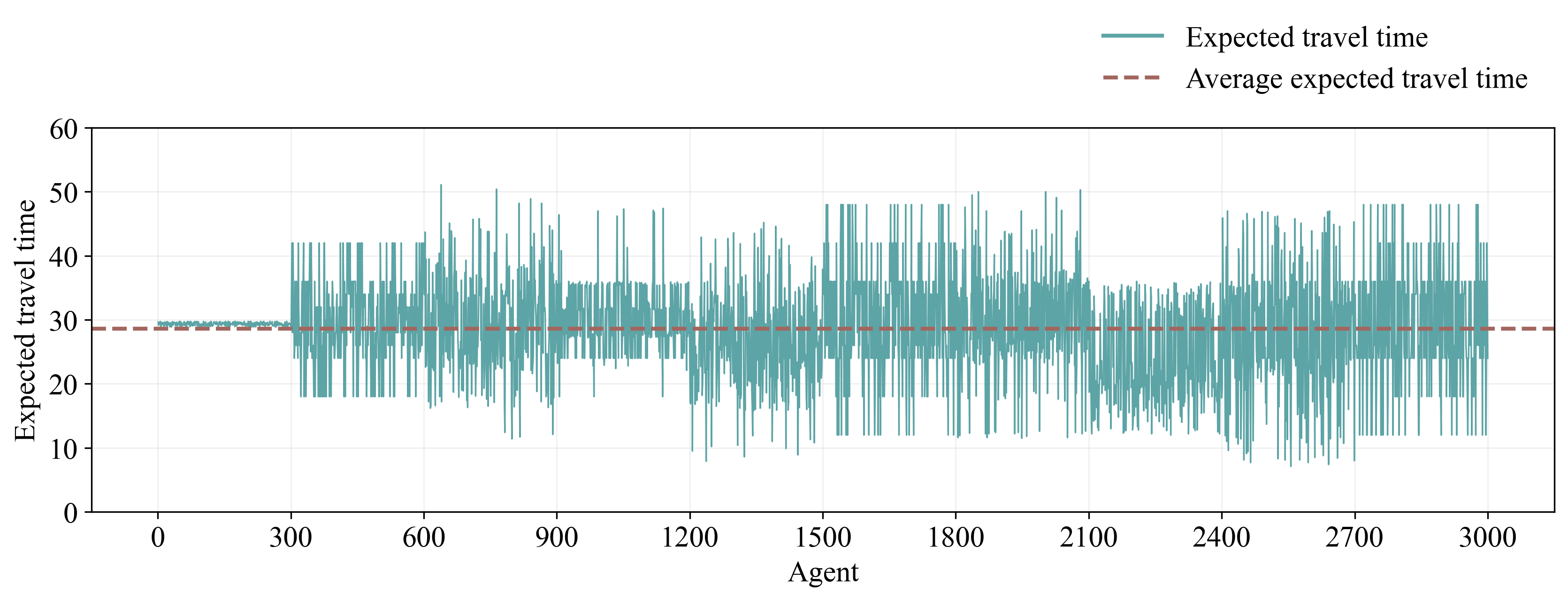}
    \label{Scenario picture 11-time-1}
    }
    \subfigure[Second simulation result]{      \includegraphics[width=0.9\textwidth]  {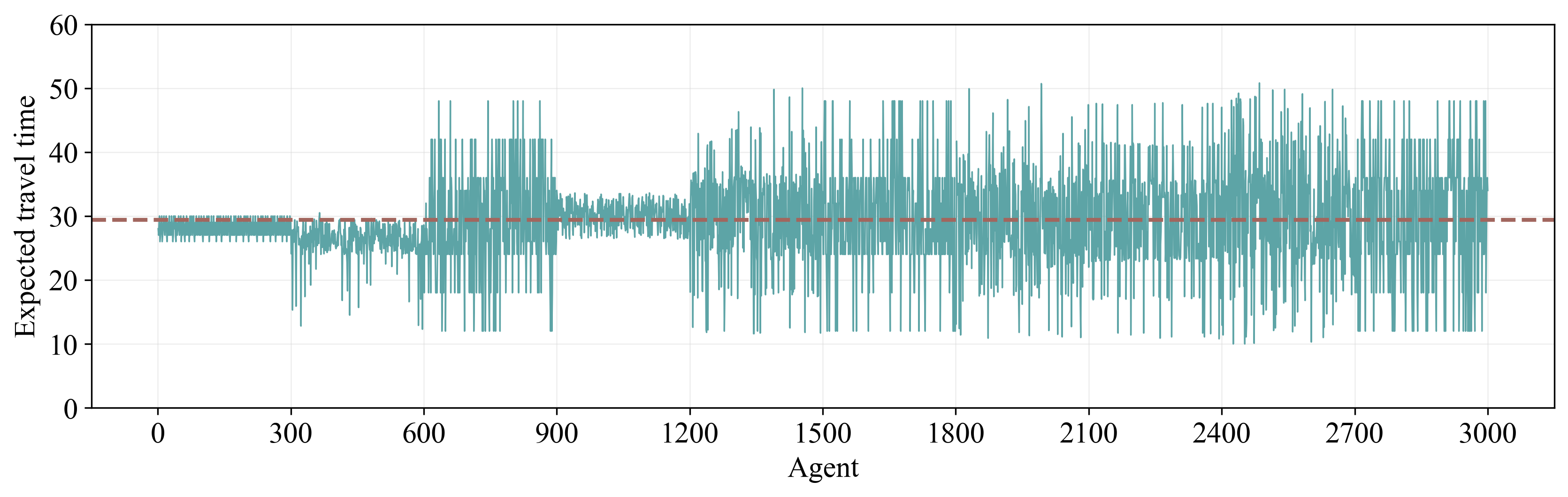}
    \label{Scenario picture 11-time-2}
    }
    \subfigure[Third simulation result]{     \includegraphics[width=0.9\textwidth]    {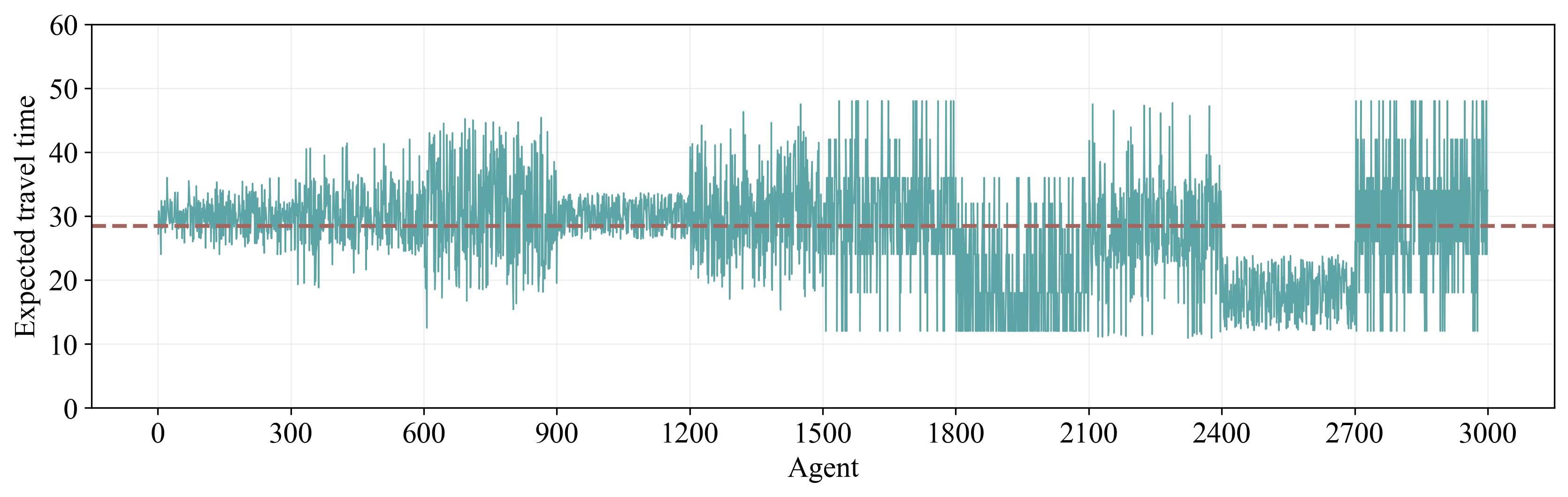}
    \label{Scenario picture 11-time-3}
    }
    \caption{Expected travel time distribution in case 11} 
    \label{fig: Expected travel time distribution in case 11} 
\end{figure}

The route choice distributions of the 3,000 agents across three independent simulations in scenarios 1, 2 and 3 are presented in \ref{sec:The route choice distributions}. 
Based on the three simulation results, the choice counts of each route under the 11 cases are summarized in Table \ref{tab: Route choice results for case 1 to case 11}. 
Furthermore, Figure \ref{fig: Expected travel time distribution in case 11} illustrates the distribution of expected travel times for the 3,000 agents across three independent simulations in experimental scenario 3. 
It is worth noting that due to the large number of human agents, the route choice scatter points are highly overplotted, forming a dense band that appears almost like a straight line, which may also create the illusion that some agents chose both route A and route B. 
In fact, this is a visualization artifact caused by point overlap. 
Taking agents 2,300–2,399 in case 1 as an example, Figure \ref{fig: Route choices for agent 2300-2399 in case 1} further presents the detailed choices of these 100 agents.

\begin{table}[!h]
    \footnotesize
    \centering
    \caption{Route choice results for case 1 to case 11}
    \label{tab: Route choice results for case 1 to case 11}
    \begin{tabular}{
    >{\raggedright\arraybackslash}m{0.5cm}
    >{\raggedright\arraybackslash}m{1.5cm}
    >{\raggedright\arraybackslash}m{1.5cm}
    >{\raggedright\arraybackslash}m{1.5cm}
    >{\raggedright\arraybackslash}m{1.5cm}
    >{\raggedright\arraybackslash}m{3cm}
    >{\raggedright\arraybackslash}m{3.1cm}
    }
    \toprule
    Case & Route type & First simulation & Second simulation & Third simulation & Average choice counts & Average choice probability \\
    \midrule
    \multirow{2}{*}{1} 
    & Route A & 1981 & 1999 & 1997 & 1985.67 & 66.19\%\\ & Route B & 1019 & 1001 & 1023 & 1014.33 & 33.81\%\\
    \multirow{2}{*}{2} 
    & Route A & 1999 & 2009 & 1920 & 1976.00 & 65.87\%\\ & Route B & 1001 & 991 & 1080 & 1024.00 & 34.13\%\\
    \multirow{2}{*}{3} 
    & Route A & 1816 & 1924 & 1928 & 1889.33 & 62.98\%\\ & Route B & 1184 & 1076 & 1072 & 1110.67 & 37.02\%\\
    \multirow{2}{*}{4} 
    & Route A & 1960 & 2021 & 1979 & 1986.67 & 66.22\%\\ & Route B & 1040 & 979 & 1021 & 1013.33 & 33.78\%\\
    \multirow{2}{*}{5} 
    & Route A & 2059 & 2020 & 2079 & 2052.67 & 68.42\%\\ & Route B & 941 & 980 & 921 & 947.33 & 31.58\%\\
    \multirow{2}{*}{6} 
    & Route A & 1403 & 1474 & 1435 & 1437.33 & 47.91\%\\ & Route B & 1597 & 1526 & 1565 & 1562.67 & 52.09\%\\
    \multirow{2}{*}{7} 
    & Route A & 1345 & 1517 & 1440 & 1434.00 & 47.80\%\\ & Route B & 1655 & 1483 & 1560 & 1566.00 & 52.20\%\\
    \multirow{2}{*}{8} 
    & Route A & 1465 & 1493 & 1262 & 1406.67 & 46.89\%\\ & Route B & 1535 & 1507 & 1738 & 1593.33 & 53.11\%\\
    \multirow{2}{*}{9} 
    & Route A & 989 & 1107 & 1167 & 1087.67 & 36.26\%\\ & Route B & 2011 & 1893 & 1833 & 1912.33 & 63.74\%\\
    \multirow{2}{*}{10} 
    & Route A & 737 & 761 & 727 & 741.67 & 24.72\%\\ & Route B & 2263 & 2239 & 2273 & 2258.33 & 75.28\%\\
    \multirow{3}{*}{11} 
    & Route A & 778 & 718 & 680 & 725.33 & 24.18\% \\ 
    & Route B & 1049 & 1025 & 1031 & 1035.00 & 34.50\% \\
    & Route C & 1173 & 1257 & 1289 & 1239.67 & 41.32\% \\
    \bottomrule
    \end{tabular}
\end{table}


Using the statistical results of cases 1 to 10 in table \ref{tab: Route choice results for case 1 to case 11}, the parameters $\alpha$, $\beta$ and $\lambda$ are estimated by fitting equations \eqref{eq: 9} and \eqref{eq: 10}, yielding the following estimates: $\alpha=0.4$, $\beta=0.64$ and $\lambda=1.43$. 
The corresponding residual sums of squares are ${f_1}(\alpha ,\gamma ) = 0.0097$ and ${f_2}(\beta ,\lambda ,\gamma ) = 0.0022$. 
The goodness of fit values are $1 - \sqrt {f_1(\alpha ,\gamma )/\sum\nolimits_{j = 1}^5 {{{(P({A_j}))}^2}} }  = 0.9332$ and $1 - \sqrt {f_2(\beta ,\lambda ,\gamma )/\sum\nolimits_{j = 6}^{10} {{{(P({A_j}))}^2}} }  = 0.9497$, indicating that the overall fitting error is small and the fit is acceptable. 
In terms of parameter interpretation, the values $\alpha=0.4$ and $\beta=0.64$ indicate that the LLM-based agents exhibit pronounced diminishing sensitivity to changes in travel-time gains and losses. 
The loss-aversion parameter $\lambda=1.43$ indicates that under the same time deviation, the agent's subjective sensitivity to the loss of lateness is 1.43 times that of the same gain, reflecting a significant loss aversion characteristic. 
Based on these estimated parameters, the cumulative prospect values and choice probabilities for routes A, B, and C in case 11 are computed, as reported in Table \ref{tab: CPT-estimated values and route choice probabilities in case 11}.

\begin{table}[!h]
    \footnotesize
    \centering
    \caption{CPT-estimated values and route choice probabilities in case 11}
    \label{tab: CPT-estimated values and route choice probabilities in case 11}
    \begin{tabular}{
    >{\raggedright\arraybackslash}m{0.5cm}
    >{\raggedright\arraybackslash}m{1.5cm}
    >{\raggedright\arraybackslash}m{3cm}
    >{\raggedright\arraybackslash}m{4.1cm}
    >{\raggedright\arraybackslash}m{4.3cm}
    }
    \toprule
    Case & Route type & Estimated CPT values & Estimated choice probability & LLM-simulated choice probability \\
    \midrule
    \multirow{3}{*}{11} 
    & Route A & -2.1524 & 28.72\% & 24.18\% \\
    & Route B & -1.7805 & 33.77\% & 34.50\% \\
    & Route C & -1.2899 & 37.51\% & 41.32\% \\
    \bottomrule
    \end{tabular}
\end{table}

\subsection{Individual-level evaluation of behavioral bias reproduction}\label{Evaluation of individual simulation}

\begin{figure}[htbp]
    \centering
    \subfigure[Type-1 agent]{        \includegraphics[width=0.48\textwidth] {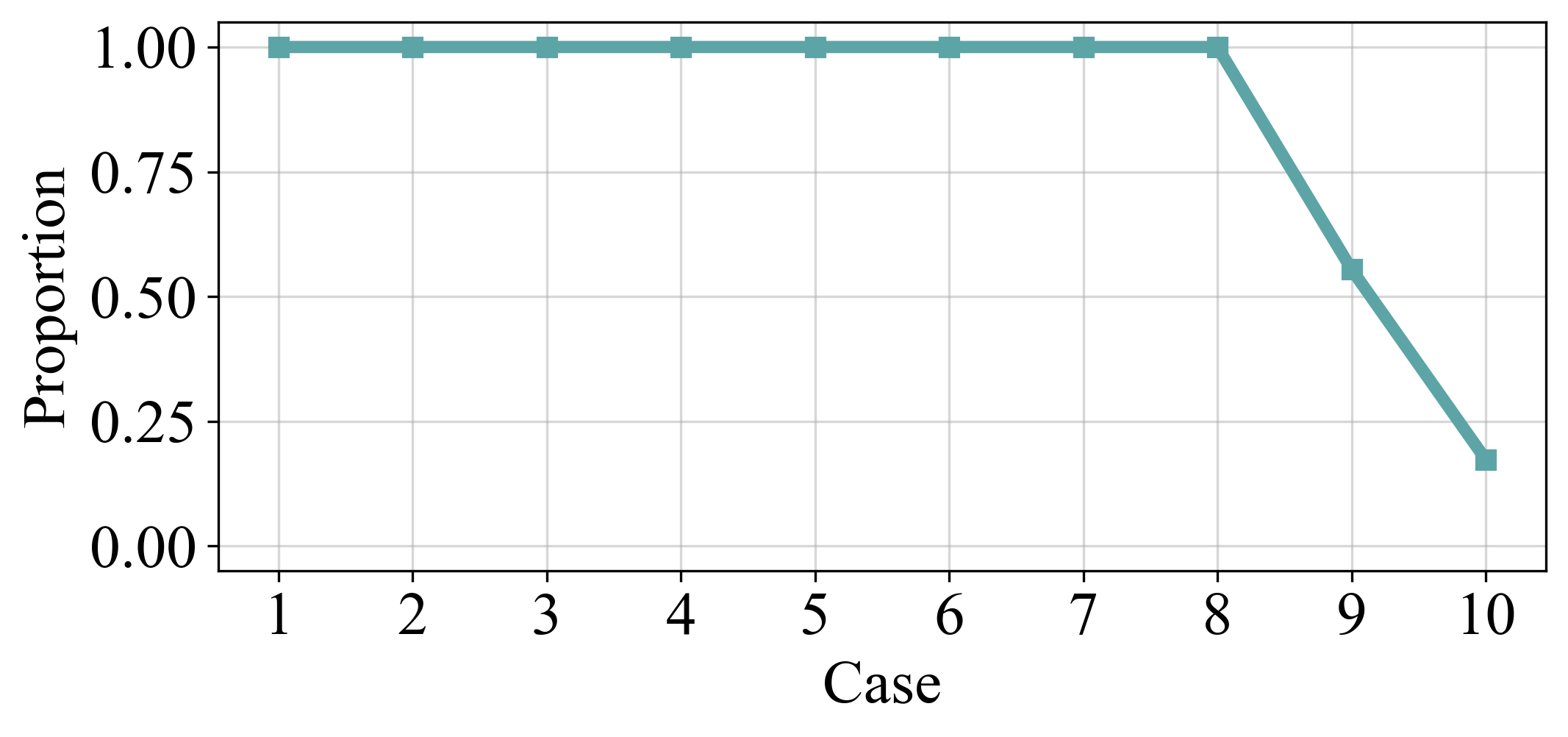}}
    \hfill
    \subfigure[Type-2 agent]{
    \includegraphics[width=0.48\textwidth] {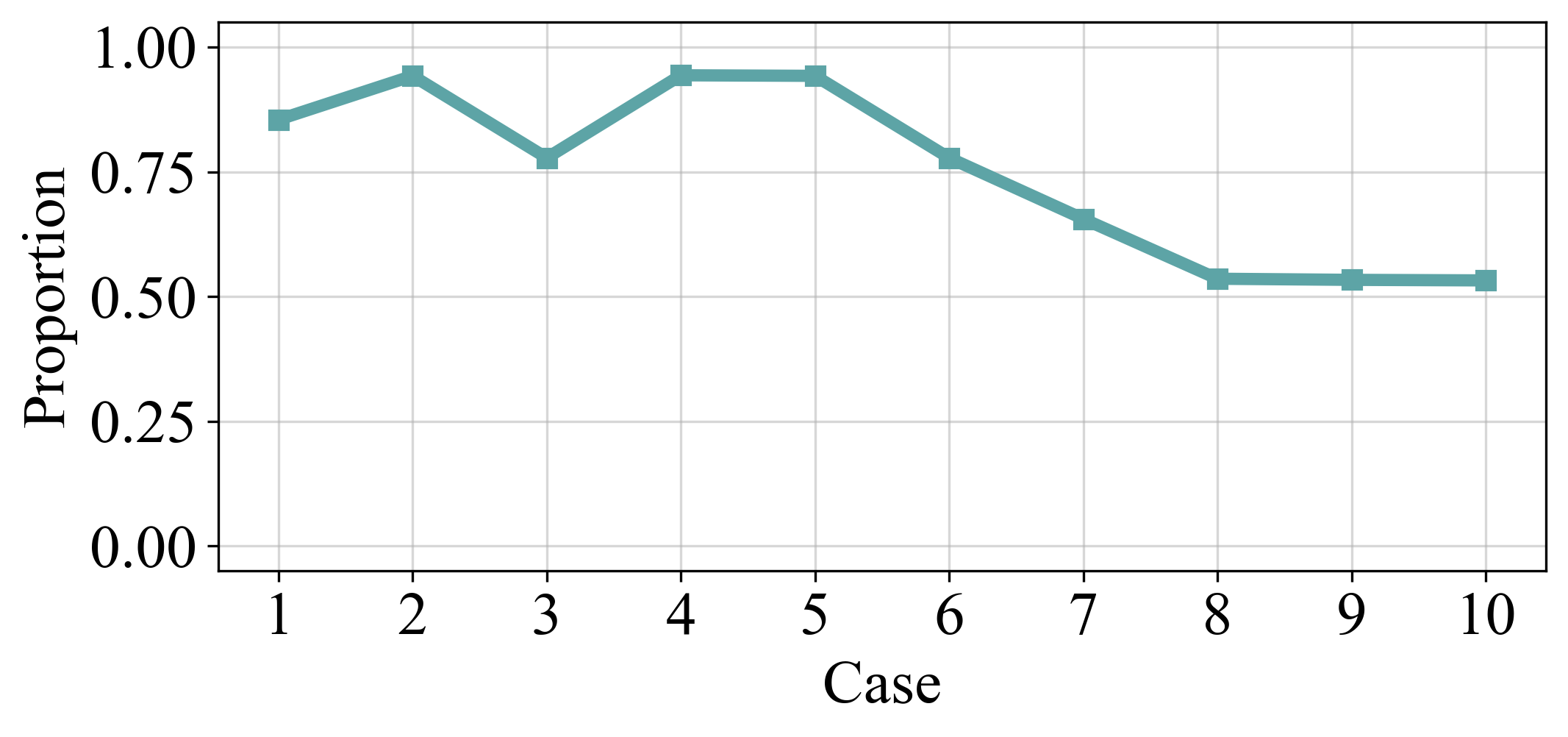}}
    \subfigure[Type-3 agent]{        \includegraphics[width=0.48\textwidth] {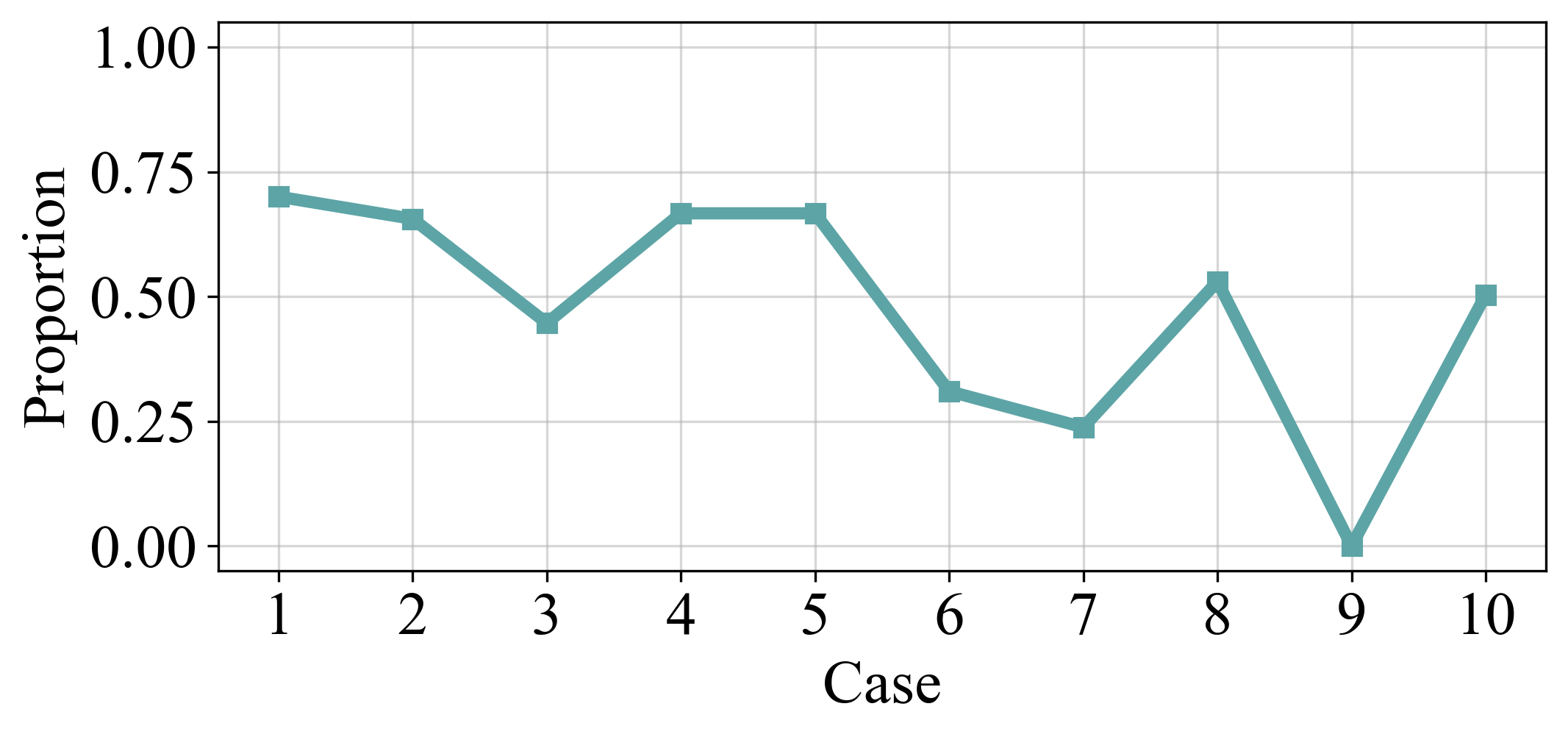}}
    \hfill
    \subfigure[Type-4 agent]{
    \includegraphics[width=0.48\textwidth] {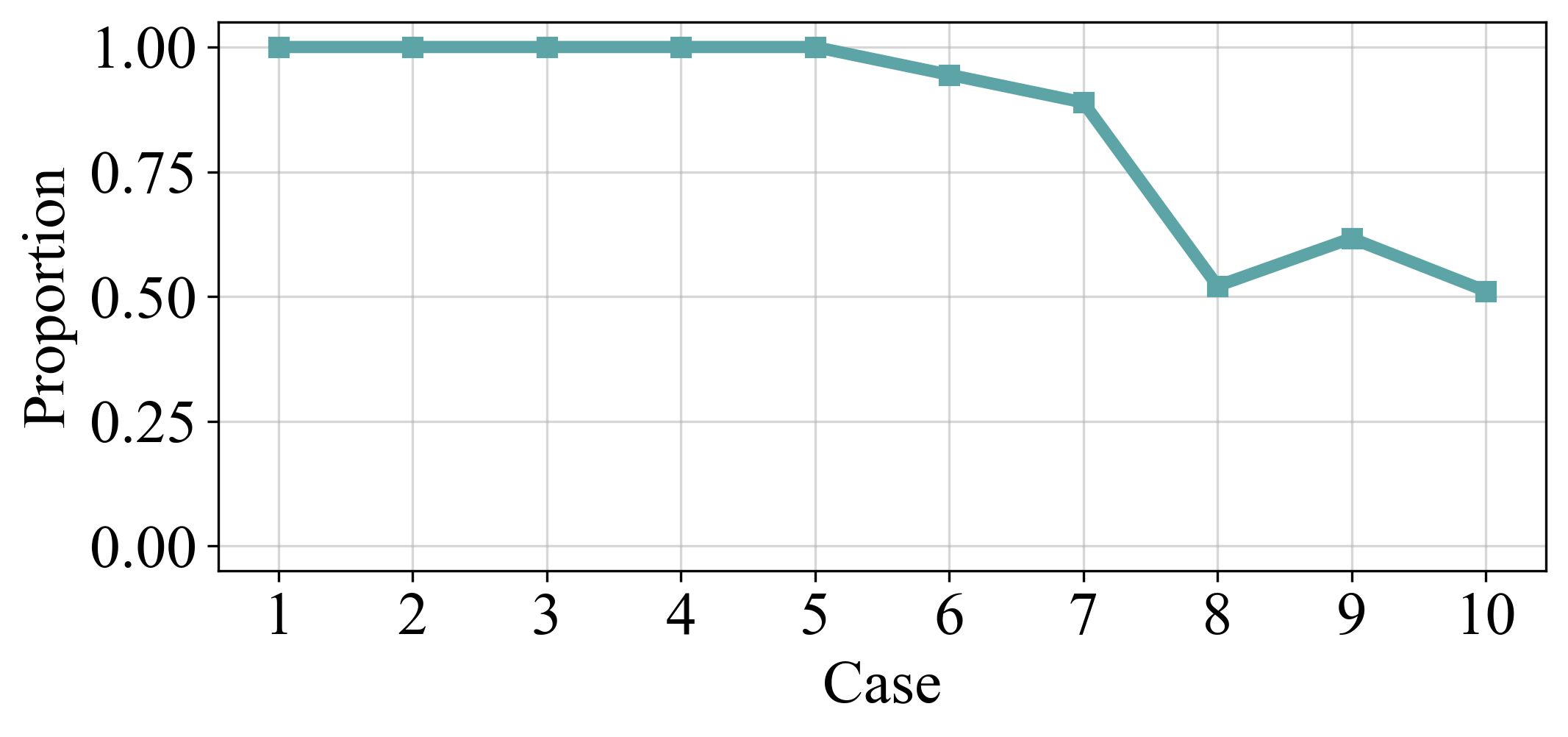}}
    \subfigure[Type-5 agent]{        \includegraphics[width=0.48\textwidth] {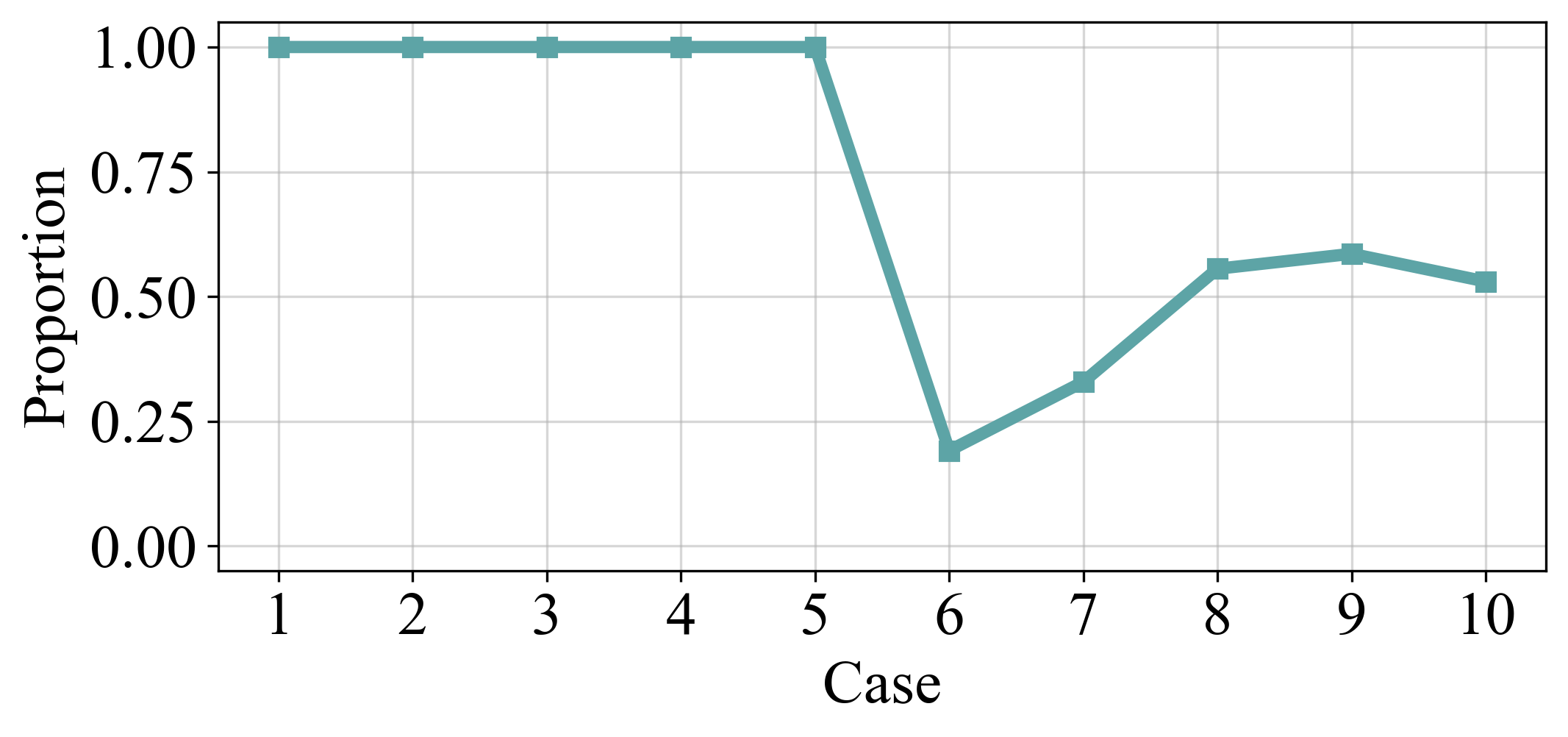}}
    \hfill
    \subfigure[Type-6 agent]{
    \includegraphics[width=0.48\textwidth] {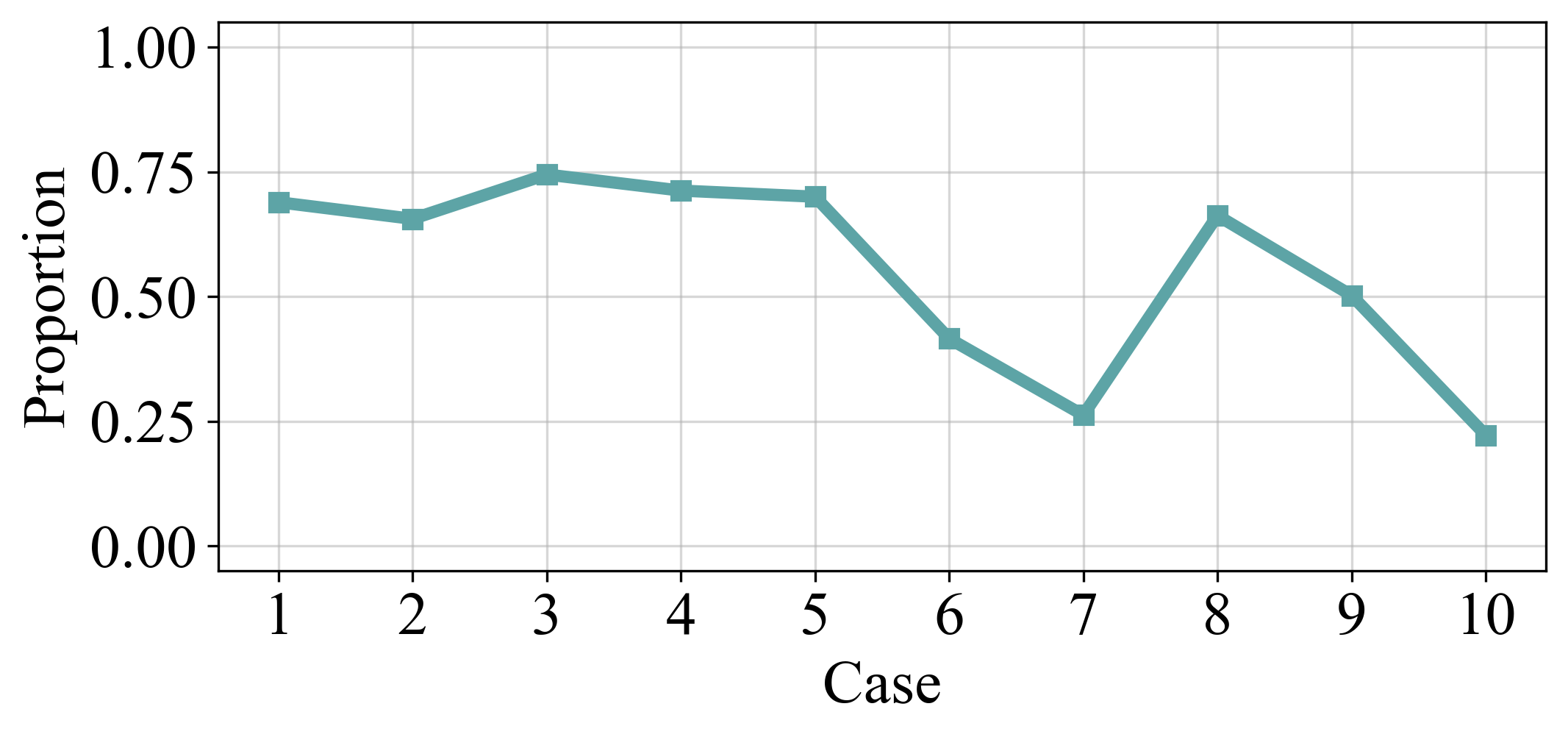}}
    \subfigure[Type-7 agent]{        \includegraphics[width=0.48\textwidth] {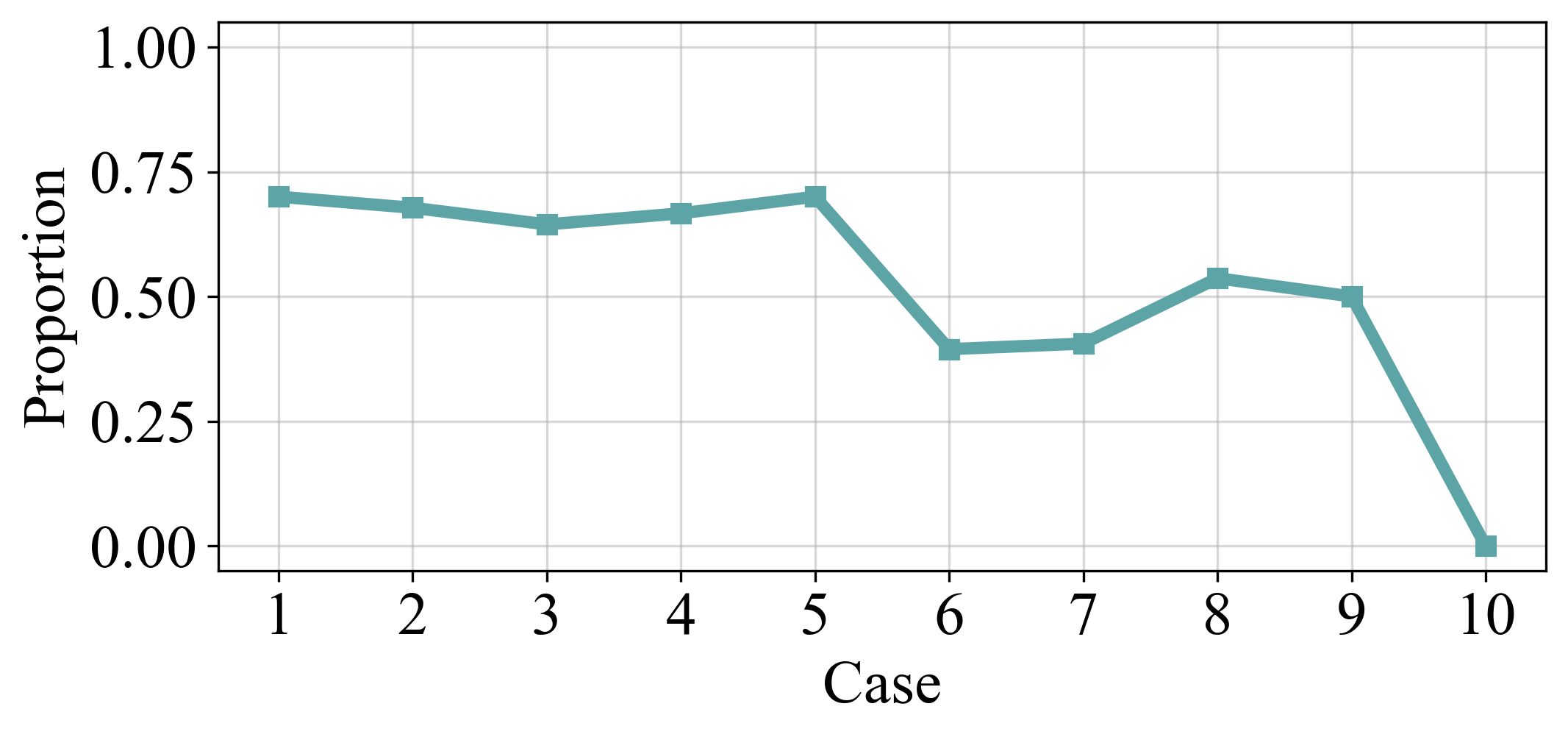}}
    \hfill
    \subfigure[Type-8 agent]{
    \includegraphics[width=0.48\textwidth] {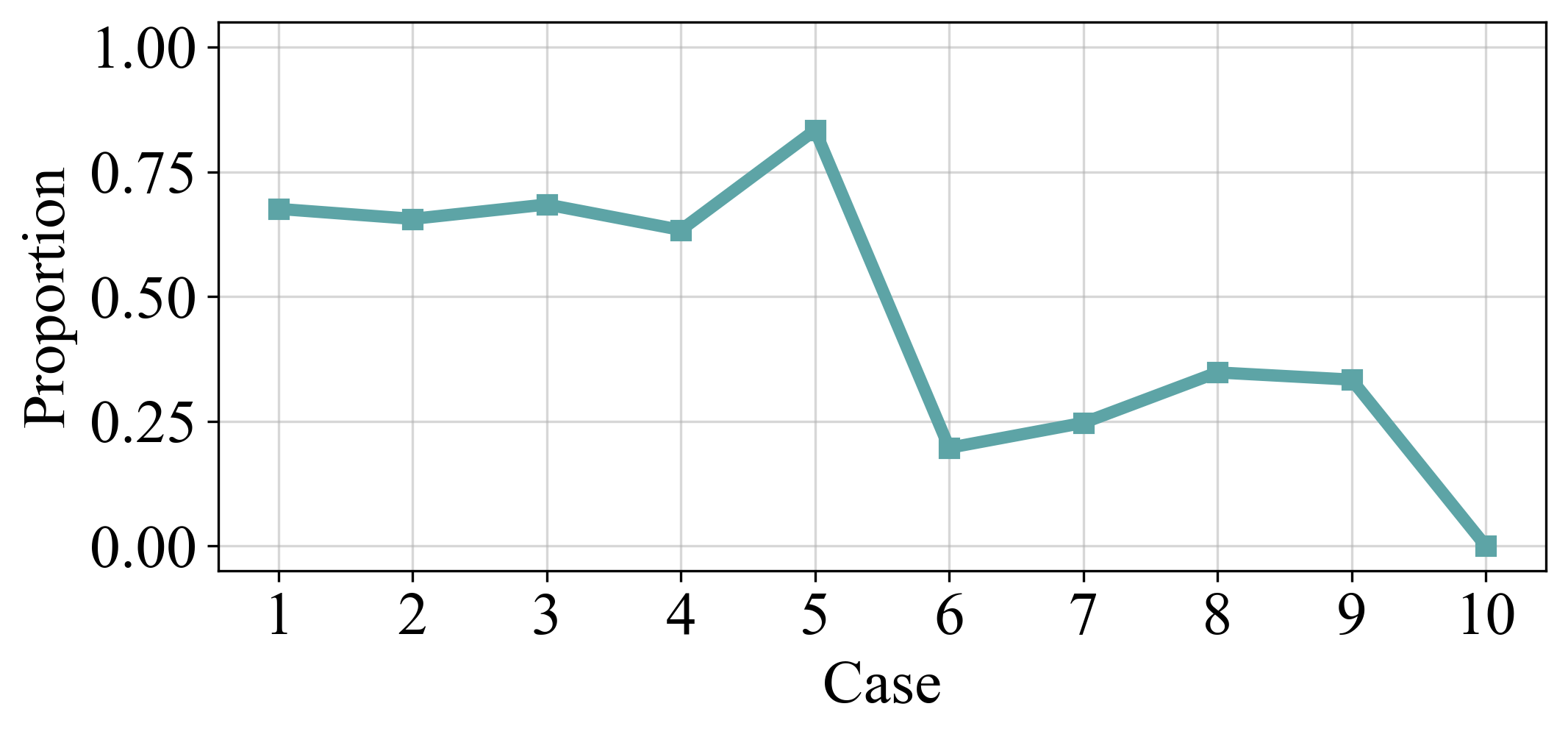}}
    \subfigure[Type-9 agent]{        \includegraphics[width=0.48\textwidth] {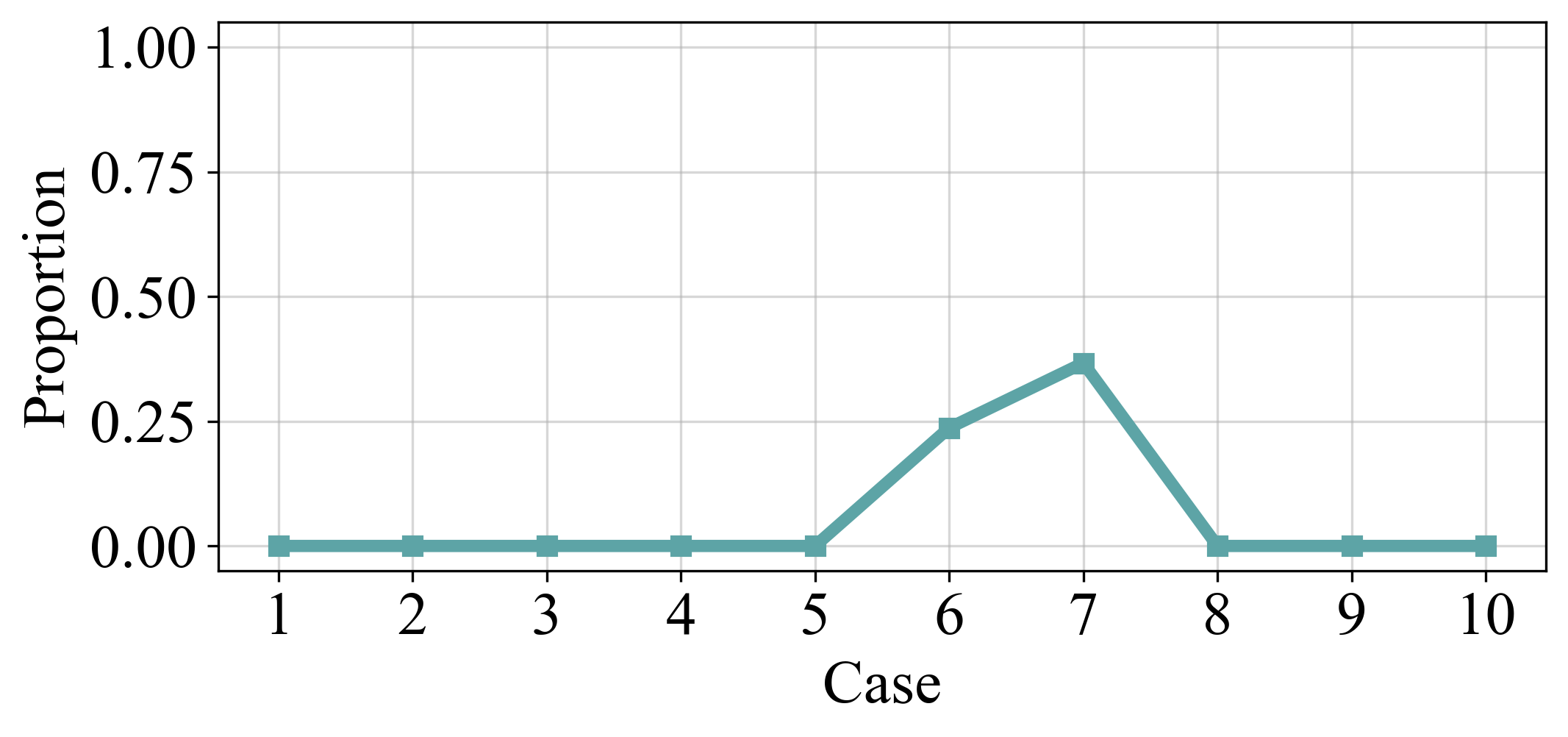}}
    \hfill
    \subfigure[Type-10 agent]{
    \includegraphics[width=0.48\textwidth] {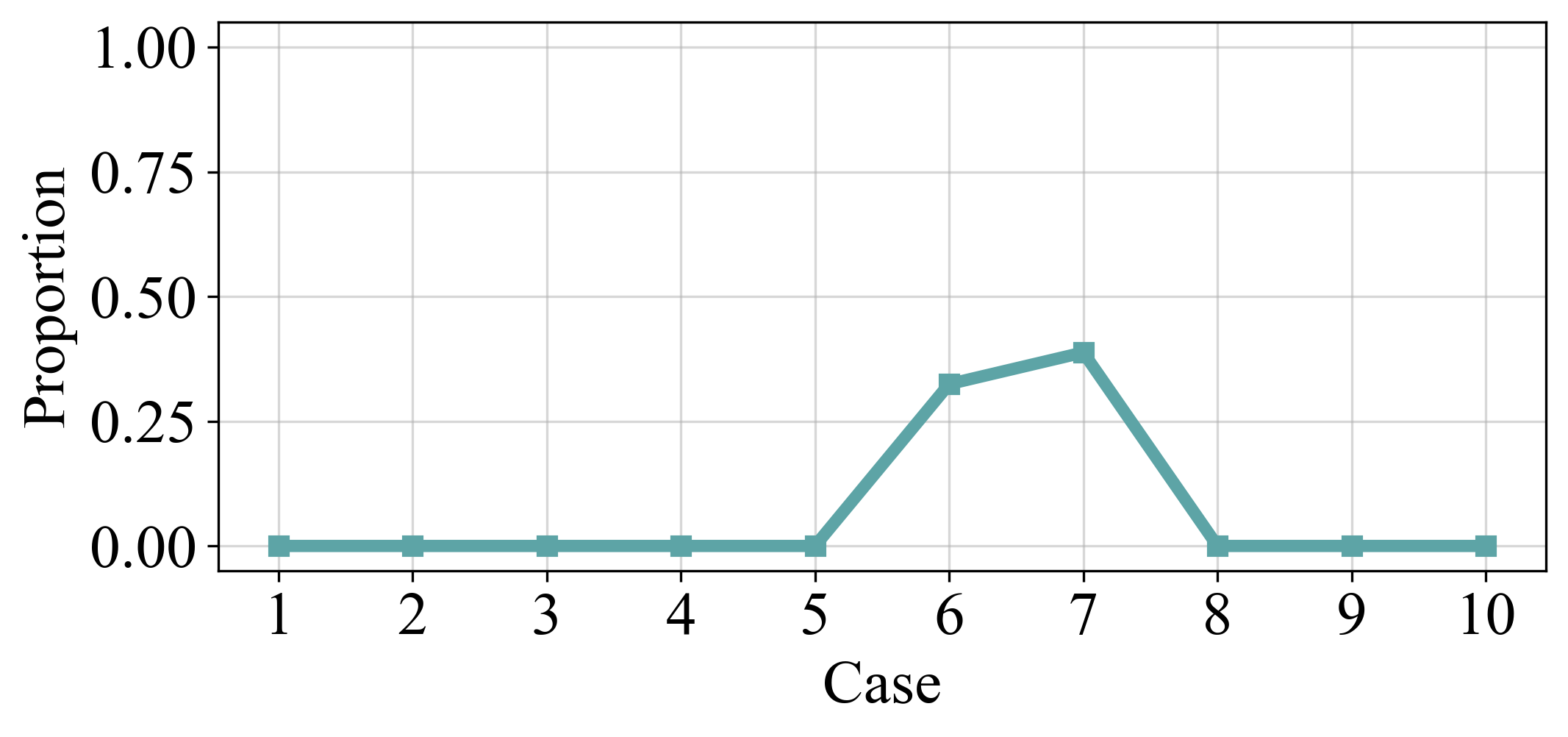}}
    \caption{Proportion of choices for route A}
    \label{fig: Number and proportion of choices for route A} 
\end{figure}

To examine whether LLMs can reproduce human behavioral biases in route-choice experiments, this study evaluates the agents' role-playing capability and behavioral responses under different risk scenarios at both the individual and environmental levels. 
At the individual level, this study analyzes the route choice results of 10 different profile types of agents in cases 1 to 10 (see Figure \ref{fig: Number and proportion of choices for route A}), statistically summarizes their choice results (see Table \ref{tab: Route choice proportions of the 10 agent types across cases 1 to 10}, \ref{tab: Route choice proportions of the 10 agent types in gain and loss cases}), and compares the demographic characteristics, travel purposes, and risk preferences of different types of agents.

Overall, risk-averse agents (types 1–5) exhibit the strongest preference for the more reliable route A, with an average choice proportion of 73.22\%. 
Risk-neutral agents (types 6–8) follow, with an average proportion of 51.33\% choosing route A. 
In contrast, risk-seeking agents (types 9 and 10) almost always choose route B, with route A chosen only 6.58\% of the time on average. 
These statistics are closely consistent with the behavioral profiles assigned to each agent type.
From the gain–loss perspective, most agent types tend to choose the stable route A in the gain domain. 
In particular, risk-averse types 1, 4, and 5 choose route A with a proportion of 100\%, while risk-neutral types 6–8 also show a high average proportion of 69.15\% for route A. 
In the loss domain, however, the choice proportion of route A decreases markedly for almost all types: for risk-averse agents (types 1–5), it drops from 90.38\% to 56.07\%, and for risk-neutral agents (types 6–8) it further declines to 33.50\%. 

Specifically, in case 10, only 17.33\% of retired-worker agents (type 1) chose the reliable route A, which appears, at first glance, to contradict their risk-averse profile. 
Further analysis shows that although route A has lower travel time variability, its on-time arrival probability is zero, meaning that choosing route A entails certain loss of being late. 
In this case, most cautious agents instead select the more volatile route B. 
This suggests that while they are generally risk-averse, they re-evaluate the outcome distribution when lateness is unavoidable and exhibit a context-dependent decision mechanism rather than mechanically following a “risk-averse” label. 
Similarly, within the overall conservative group, courier human agents (type 3) choose route A with an average probability of 47.17\%, which is lower than their probability of choosing route B (52.83\%). 
This can be attributed to the fact that courier delivery is highly sensitive to on-time or even early arrival, which directly affects income and performance evaluations. 
Therefore, rather than simply avoiding travel-time variability, this type is, in some cases, more willing to choose the shorter but more variable route B to gain the potential benefit of earlier arrival.

In summary, different types of agents exhibit significant differences in route choice behavior, indicating that LLMs can generate behavioral responses through conditional reasoning based on given profiles. 
Furthermore, the agents' route choice results exhibit decision-making behavior consistent with CPT, suggesting that LLMs can not only simulate individual differences but also reproduce irrational human choice biases under uncertainty.

\begin{table}[!h]
    \footnotesize
    \centering
    \caption{Route choice proportions of the 10 agent types across cases 1 to 10}
    \label{tab: Route choice proportions of the 10 agent types across cases 1 to 10}
    \begin{tabular}{
    >{\raggedright\arraybackslash}m{0.4cm}
    >{\raggedright\arraybackslash}m{1cm}
    >{\raggedright\arraybackslash}m{0.9cm}
    >{\raggedright\arraybackslash}m{0.9cm}
    >{\raggedright\arraybackslash}m{0.9cm}
    >{\raggedright\arraybackslash}m{0.9cm}
    >{\raggedright\arraybackslash}m{0.9cm}
    >{\raggedright\arraybackslash}m{0.9cm}
    >{\raggedright\arraybackslash}m{0.9cm}
    >{\raggedright\arraybackslash}m{0.9cm}
    >{\raggedright\arraybackslash}m{0.9cm}
    >{\raggedright\arraybackslash}m{1cm}
    }
    \toprule
    Case & Route type & Type 1 & Type 2 & Type 3 & Type 4 & Type 5 & Type 6 & Type 7 & Type 8 & Type 9 & Type 10\\
    \midrule
    \multirow{2}{*}{1} 
    & Route A & 100.00\% & 85.44\% & 70.00\% & 100.00\% & 100.00\% & 68.89\% & 70.00\% & 67.56\% & 0.00\% & 0.00\%\\ 
    & Route B & 0.00\% & 14.56\% & 30.00\% & 0.00\% & 0.00\% & 31.11\% & 30.00\% & 32.44\% & 100.00\% & 100.00\%\\
    \multirow{2}{*}{2} 
    & Route A & 100.00\% & 94.22\% & 65.56\% & 100.00\% & 100.00\% & 65.56\% & 67.78\% & 65.56\% & 0.00\% & 0.00\%\\ 
    & Route B & 0.00\% & 5.78\% & 34.44\% & 0.00\% & 0.00\% & 34.44\% & 32.22\% & 34.44\% & 100.00\% & 100.00\%\\ \multirow{2}{*}{3} 
    & Route A & 100.00\% & 77.78\% & 44.67\% & 100.00\% & 100.00\% & 74.44\% & 64.44\% & 68.44\% & 0.00\% & 0.00\%\\ 
    & Route B & 0.00\% & 22.22\% & 55.33\% & 0.00\% & 0.00\% & 25.56\% & 35.56\% & 31.56\% & 100.00\% & 100.00\%\\
    \multirow{2}{*}{4} 
    & Route A & 100.00\% & 94.33\% & 66.67\% & 100.00\% & 100.00\% & 71.22\% & 66.67\% & 63.33\% & 0.00\% & 0.00\%\\ 
    & Route B & 0.00\% & 5.67\% & 33.33\% & 0.00\% & 0.00\% & 28.78\% & 33.33\% & 36.67\% & 100.00\% & 100.00\%\\
    \multirow{2}{*}{5} 
    & Route A & 100.00\% & 94.22\% & 66.67\% & 100.00\% & 100.00\% & 70.00\% & 70.00\% & 83.33\% & 0.00\% & 0.00\%\\ 
    & Route B & 0.00\% & 5.78\% & 33.33\% & 0.00\% & 0.00\% & 30.00\% & 30.00\% & 16.67\% & 100.00\% & 100.00\%\\
    \multirow{2}{*}{6} 
    & Route A & 100.00\% & 77.78\% & 31.00\% & 94.44\% & 19.11\% & 41.56\% & 39.44\% & 19.67\% & 23.67\% & 32.44\%\\ 
    & Route B & 0.00\% & 22.22\% & 69.00\% & 5.56\% & 80.89\% & 58.44\% & 60.56\% & 80.33\% & 76.33\% & 67.56\%\\
    \multirow{2}{*}{7} 
    & Route A & 100.00\% & 65.56\% & 23.78\% & 88.89\% & 32.89\% & 26.22\% & 40.56\% & 24.67\% & 36.67\% & 38.78\%\\ 
    & Route B & 0.00\% & 34.44\% & 76.22\% & 11.11\% & 67.11\% & 73.78\% & 59.44\% & 75.33\% & 63.33\% & 61.22\%\\
    \multirow{2}{*}{8} 
    & Route A & 100.00\% & 53.56\% & 53.00\% & 52.11\% & 55.56\% & 66.22\% & 53.67\% & 34.78\% & 0.00\% & 0.00\%\\ 
    & Route B & 0.00\% & 46.44\% & 47.00\% & 47.89\% & 44.44\% & 33.78\% & 46.33\% & 65.22\% & 100.00\% & 100.00\%\\
    \multirow{2}{*}{9} 
    & Route A & 55.44\% & 53.33\% & 0.00\% & 61.67\% & 58.56\% & 50.22\% & 50.00\% & 33.33\% & 0.00\% & 0.00\%\\ 
    & Route B & 44.56\% & 46.67\% & 100.00\% & 38.33\% & 41.44\% & 49.78\% & 50.00\% & 66.67\% & 100.00\% & 100.00\%\\
    \multirow{2}{*}{10} 
    & Route A & 17.33\% & 53.22\% & 50.33\% & 51.11\% & 53.00\% & 22.22\% & 0.00\% & 0.00\% & 0.00\% & 0.00\%\\ 
    & Route B & 82.67\% & 46.78\% & 49.67\% & 48.89\% & 47.00\% & 77.78\% & 100.00\% & 100.00\% & 100.00\% & 100.00\%\\
    \bottomrule
    \end{tabular}
\end{table}

\begin{table}[!h]
    \footnotesize
    \centering
    \caption{Route choice proportions of the 10 agent types in gain and loss cases}
    \label{tab: Route choice proportions of the 10 agent types in gain and loss cases}
    \begin{tabular}{
    >{\raggedright\arraybackslash}m{1cm}
    >{\raggedright\arraybackslash}m{1.3cm}
    >{\raggedright\arraybackslash}m{0.8cm}
    >{\raggedright\arraybackslash}m{0.8cm}
    >{\raggedright\arraybackslash}m{0.8cm}
    >{\raggedright\arraybackslash}m{0.8cm}
    >{\raggedright\arraybackslash}m{0.8cm}
    >{\raggedright\arraybackslash}m{0.8cm}
    >{\raggedright\arraybackslash}m{0.8cm}
    >{\raggedright\arraybackslash}m{0.8cm}
    >{\raggedright\arraybackslash}m{0.8cm}
    >{\raggedright\arraybackslash}m{1cm}
    }
    \toprule
    \phantom{Type} & Route type & Type 1 & Type 2 & Type 3 & Type 4 & Type 5 & Type 6 & Type 7 & Type 8 & Type 9 & Type 10\\
    \midrule
    \multirow{2}{*}{Gain case} 
    & Route A & 100.00\% & 89.20\% & 62.71\% & 100.00\% & 100.00\% & 70.02\% & 67.78\% & 69.64\% & 0.00\% & 0.00\%\\ 
    & Route B & 0.00\% & 10.80\% & 37.29\% & 0.00\% & 0.00\% & 29.98\% & 32.22\% & 30.36\% & 100.00\% & 100.00\%\\
    \multirow{2}{*}{Loss case} 
    & Route A & 74.56\% & 60.69\% & 31.62\% & 69.64\% & 43.82\% & 41.29\% & 36.73\% & 22.49\% & 12.07\% & 14.24\%\\ 
    & Route B & 25.44\% & 39.31\% & 68.38\% & 30.36\% & 56.18\% & 58.71\% & 63.27\% & 77.51\% & 87.93\% & 85.76\%\\
    \multirow{2}{*}{Average} 
    & Route A & 87.28\% & 74.94\% & 47.17\% & 84.82\% & 71.91\% & 55.66\% & 52.26\% & 46.07\% & 6.03\% & 7.12\%\\ 
    & Route B & 12.72\% & 25.06\% & 52.83\% & 15.18\% & 28.09\% & 44.34\% & 47.74\% & 53.93\% & 93.97\% & 92.88\%\\
    \bottomrule
    \end{tabular}
\end{table}

\subsection{Environmental-level evaluation of context-dependent group behavior}\label{Evaluation of scenario simulation}

\begin{figure}[htbp]
    \centering
    \includegraphics[width=0.7\textwidth]{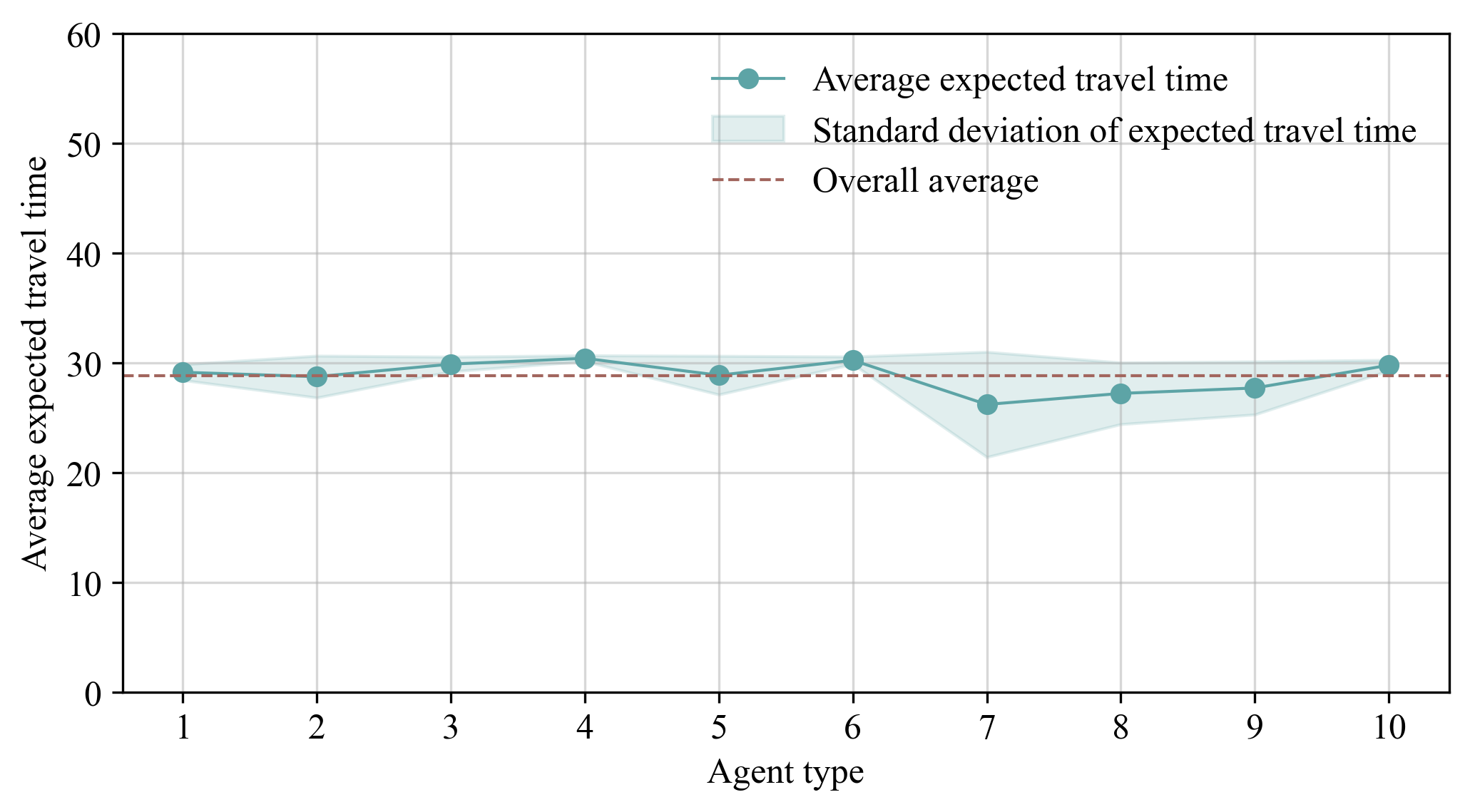} 
    \caption{Distribution of expected travel time for the 10 agent types in case 11} 
    \label{fig: Distribution of expected travel time for the 10 agent types in case 11} 
\end{figure}

Across three repeated simulation experiments, the average expected travel time of each agent type falls within the feasible travel time range of the corresponding route (as shown in Figure \ref{fig: Distribution of expected travel time for the 10 agent types in case 11}). 
The overall mean of expected travel time across agent types is 28.83 minutes, which is very close to the average travel time of 30 minutes for the three routes. 
This indicates that, regardless of risk preference, agents' subjective expectations are generally centered around the mean and are not reduced simply because a route contains a small probability of extremely short travel times. 
In other words, the travel time distributions induce time expectations that are more consistent with rational distributional characteristics, rather than being dominated by extreme tail events. 
Moreover, in terms of the stability of expected travel time, most agent types exhibit relatively low variance. 
For example, types 4 and 6 show very small variances, and their expected travel times remain almost unchanged in different replication experiments.

Although the overall expectation level is relatively stable, differences still exist in the experimental reproduction among different agent types. 
This indicates that the travel environment does not output a single fixed time perception. 
Instead, while maintaining stable outputs, it allows different agent types to exhibit varying degrees of sensitivity to uncertainty. 
For instance, type 7 shows a relatively low mean expected travel time but a large variance. 
This does not contradict its risk-preference setting; rather, it can be consistently explained by its profile characteristics. 
As a teacher commuter with a moderate level of time pressure, the type 7 agents are rational and efficiency-oriented, can tolerate minor fluctuations, and tend to lower their subjective expected travel time more actively when environmental risk is low. 
Importantly, risk neutrality does not imply that expectations do not fluctuate; instead, it reflects a tendency to avoid extreme risk aversion or risk seeking and to balance efficiency and stability within a tolerable range of uncertainty.

Furthermore, by statistically summarizing the route choice results in case 11, we obtain the group choice results in Table \ref{tab: Route choice proportions of the 10 agent types in cases 11}. 
As shown in Table \ref{tab: Route choice proportions of the 10 agent types in cases 11}, conservative human agents (type 1-5) overall tend to prefer the route C with lower travel time variability (approximately 62.11\%) as well as route B with moderate variability (approximately 29.64\%), which is consistent with the individual-level findings that conservative agents favor stable options. 
Risk-neutral human agents (type 6-8) distribute their choices in a relatively balanced way across the three routes. 
Risk-seeking human agents show a stronger preference for Route A (approximately 48.78\%), which offers shorter travel times but also larger fluctuations. 
Notably, within the conservative group, type 3 income-constrained courier agents still exhibit a preference for route B (approximately 55.89\%). 
This is consistent with their behavior in the two-route settings, where they moderately accept risk to achieve earlier deliveries, demonstrating behavioral continuity across scenarios.

In summary, at the environmental level, the expected travel times and route choice results in case 11 reflect rational perceptions of route travel time distributions while also exhibiting clear behavioral differentiation across risk types. 
These differences are consistent with the predefined profiles of the human agents and corroborate the individual level analysis. 
Therefore, the LLM-based agents are able not only to reproduce human non-rational choice biases at the individual level but also to generate context-dependent group behavioral patterns at the environmental level.

\begin{table}[!h]
    \footnotesize
    \centering
    \caption{Route choice proportions of the 10 agent types in cases 11}
    \label{tab: Route choice proportions of the 10 agent types in cases 11}
    \begin{tabular}{
    >{\raggedright\arraybackslash}m{0.6cm}
    >{\raggedright\arraybackslash}m{1.6cm}
    >{\raggedright\arraybackslash}m{0.8cm}
    >{\raggedright\arraybackslash}m{0.8cm}
    >{\raggedright\arraybackslash}m{0.8cm}
    >{\raggedright\arraybackslash}m{0.8cm}
    >{\raggedright\arraybackslash}m{0.8cm}
    >{\raggedright\arraybackslash}m{0.8cm}
    >{\raggedright\arraybackslash}m{0.8cm}
    >{\raggedright\arraybackslash}m{0.8cm}
    >{\raggedright\arraybackslash}m{0.8cm}
    >{\raggedright\arraybackslash}m{1.1cm}
    }
    \toprule
    Case & Route type & Type 1 & Type 2 & Type 3 & Type 4 & Type 5 & Type 6 & Type 7 & Type 8 & Type 9 & Type 10\\
    \midrule
    \multirow{3}{*}{11} 
    & Route A & 0.00\% & 2.67\% & 18.00\% & 2.00\% & 18.55\% & 39.67\% & 31.45\% & 31.89\% & 48.33\% & 49.22\%\\ 
    & Route B & 4.11\% & 34.55\% & 55.89\% & 5.89\% & 47.78\% & 30.22\% & 40.33\% & 38.89\% & 36.56\% & 50.78\%\\
    & Route C & 95.89\% & 62.78\% & 26.11\% & 92.11\% & 33.67\% & 30.11\% & 28.22\% & 29.22\% & 15.11\% & 0.00\%\\
    \bottomrule
    \end{tabular}
\end{table}

\subsection{Method comparison}\label{Method comparison}

To further examine the accuracy of LLMs in reproducing human behavioral biases, this section compares the parameters obtained from LLM-based simulations with those estimated from real-world behavioral experiments. 

\begin{figure}[htbp]
    \centering
    \includegraphics[width=0.6\textwidth]{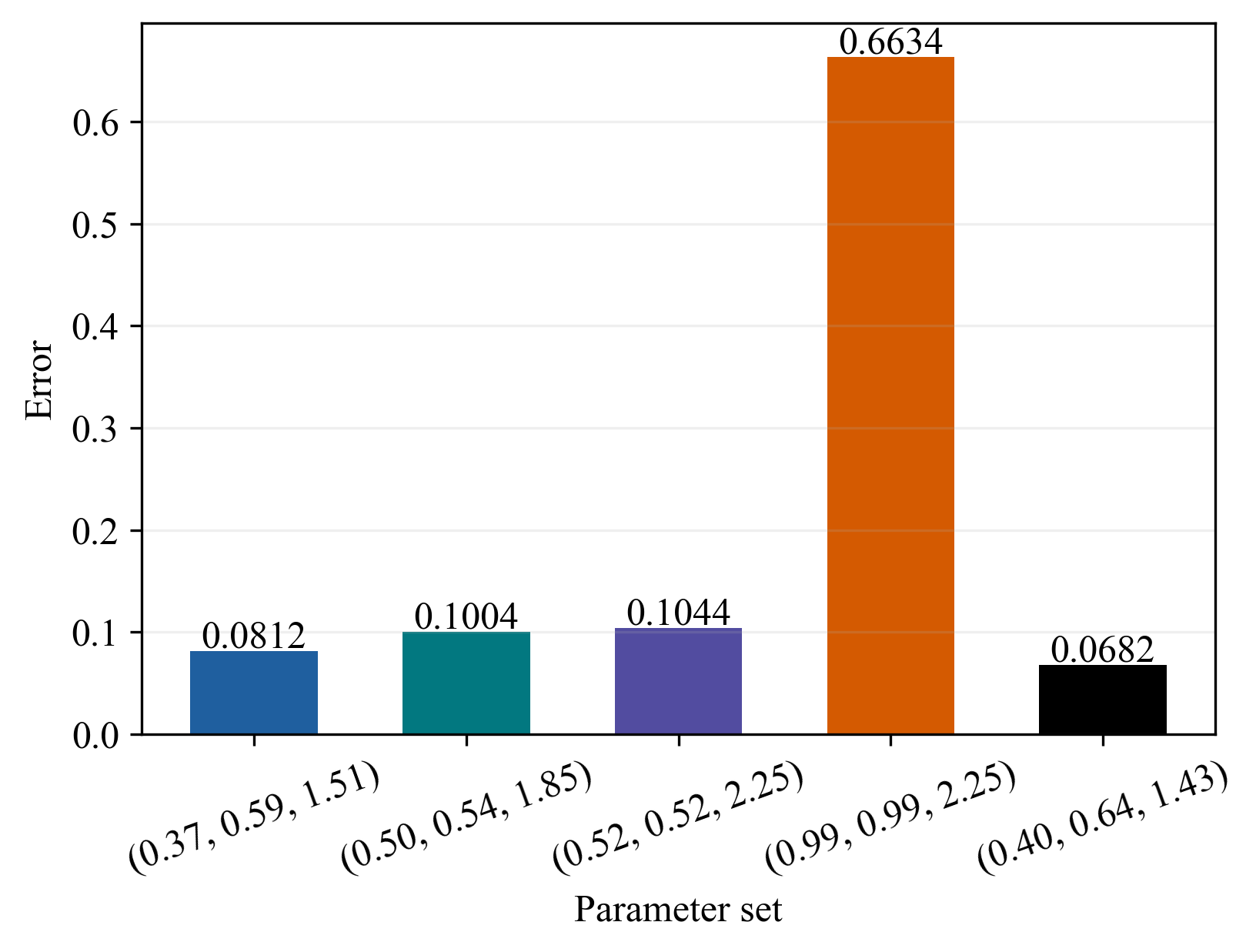} 
    \caption{Error values in real behavioral experiments} 
    \label{fig: Error values in real behavioral experiments} 
\end{figure}

This study adopts the route choice experimental data conducted by \citet{xu2011decision} based on real respondents as the reference benchmark. 
The experiment designed two travel routes with different travel time distributions and recorded respondents' route choices results under the given scenario. 
For horizontal comparison, this study selects four representative sets of parameters, all of which were estimated under relatively universal experimental conditions and have cross-scenario representativeness: (1) the parameters (0.37, 0.59, 1.51) and (0.50, 0.54, 1.85) estimated based on the route choice experiment of real respondents in \citet{xu2011decision} and \citet{zhang2018cumulative}. 
(2) Two sets of classic prospect utility parameters (0.52, 0.52, 2.25) and (0.99, 0.99, 2.25) from non-travel decision situations (\citealp{xu2011decision}). 
These parameters are widely used to characterize risk preferences in domains such as finance and gambling. 
(3) The prospect utility parameters (0.40, 0.64, 1.43) estimated based on the simulated data of LLM human agents in this study. 
In this scenario, CPT is used to compute the prospect utility of each route and the corresponding choice probabilities under different parameter sets. 
The error values between the prediction results and the route choice results of real respondents are calculated using equation \ref{eq: 12}, as shown in Table \ref{tab: Prediction results based on CPT under different parameters}. 
Equation \ref{eq: 12} represents the sum of squared residuals between the predicted values ${{{\tilde p}_l}}$ and the observed values ${{p_l}}$ over $L$ routes. 
Figure \ref{fig: Error values in real behavioral experiments} presents the total error between the predicted and observed values in \citet{xu2011decision}'s study.
\begin{equation}
    e = \sum\nolimits_{l = 1}^L {{{({{\tilde p}_l} - {p_l})}^2}}
    \label{eq: 12}
\end{equation}

Overall, as shown in Figure \ref{fig: Error values in real behavioral experiments}, the parameters estimated using the LLM-based route choice simulation method in this study have the smallest total error under multiple reference points, at 0.0682. 
The total errors of the parameter sets from \citet{xu2011decision} and \citet{zhang2018cumulative} are 0.0812 and 0.1004, respectively, whereas the classic parameter (0.99, 0.99, 2.25) from non-travel fields has the largest error, at 0.6634. 
This indicates that cross-domain general parameters cannot effectively characterize risky behavior in route choice, whereas the proposed method better approximates actual human decision-making and achieves higher overall predictive accuracy.

Furthermore, as shown in Table \ref{tab: Prediction results based on CPT under different parameters}, the parameters in this paper exhibit the smallest prediction error at reference points 30 and 35, with the error at reference point 40 only slightly higher than that of the first group, and are not optimal at reference point 50. 
Because reference points can be considered potential psychological thresholds that vary with individual and scenario factors, they often fluctuate with changes in individual attributes and context. 
This paper achieves the minimum total error across multiple reference points.
At a reference point of 50, the proposed method yields a relatively large local error; however, this error remains smaller than that obtained using the parameters estimated from the real-world behavioral experiment reported in \citet{xu2011decision}. 
This indicates that when both routes fall into the loss domain, choice probabilities are more sensitive to the loss aversion coefficient and the curvature of the value function, thus amplifying the error differences of different parameter combinations at this reference point. 

The aforementioned parameter sets are subsequently applied to predict the three-route choice results in case 11. 
Specifically, each parameter set is incorporated into the CPT framework to compute the prospect values and choice probabilities of the three routes in case 11, and the predicted results are compared with the simulation route choice results generated by the LLM-based human agents. 
The corresponding prediction results  and errors values are shown in Table \ref{tab: Prediction results and error values in LLM simulation experiments}. 
As shown in Table \ref{tab: Prediction results and error values in LLM simulation experiments}, all parameter sets exhibit the similar bias trend in predicting route choices results. 
The predicted probability of choosing route A is generally higher than the corresponding proportion observed for the LLM-based human agents in the simulation, while the predicted probabilities of choosing routes B and C are generally lower. 
This indicates that, compared to the simulated choice behavior of the LLM-based human agents, the parameter prediction results tend to overestimate the attractiveness of routes with shorter possible travel times, while underestimating the attractiveness of routes with more stable travel times. 
In terms of error metrics, the method presented in this paper has the smallest prediction error in case 11, at only 0.0036, while the parameter combination (0.99, 0.99, 2.25) corresponds to the largest prediction error, at 0.0110. 
This trend is consistent with the conclusions obtained in the real respondent experiments.


\begin{table}[!h]
    \footnotesize
    \centering
    \caption{Prediction results based on CPT under different parameters}
    \label{tab: Prediction results based on CPT under different parameters}
    \begin{tabular}{
    >{\raggedright\arraybackslash}m{2.9cm}
    >{\raggedright\arraybackslash}m{3cm}
    >{\raggedright\arraybackslash}m{2.1cm}
    >{\raggedright\arraybackslash}m{2.1cm}  
    >{\raggedright\arraybackslash}m{3.3cm}
    }
    \toprule
    Parameters & Reference point & Route 1 & Route 2 & Error value\\
    \midrule
    \multirow{4}{*}{(0.37, 0.59, 1.51)} 
    & 30 & 69.31\% & 30.69\% & 0.0050\\ 
    & 35 & 29.39\% & 70.61\% & 0.0100\\ 
    & 40 & 14.33\% & 85.67\% & 0.0003\\
    & 50 & 44.65\% & 55.35\% & 0.0659\\ 
    \multirow{4}{*}{(0.50, 0.54, 1.85)} 
    & 30 & 75.63\% & 24.37\% & 0.0257\\ 
    & 35 & 33.86\% & 66.14\% & 0.0267\\ 
    & 40 & 13.69\% & 86.31\% & 0.0007\\
    & 50 & 41.88\% & 58.12\% & 0.0473\\ 
    \multirow{4}{*}{(0.52, 0.52, 2.25)} 
    & 30 & 78.37\% & 21.63\% & 0.0396\\ 
    & 35 & 31.24\% & 68.76\% & 0.0160\\ 
    & 40 & 10.94\% & 89.06\% & 0.0042\\
    & 50 & 41.43\% & 58.57\% & 0.0446\\
    \multirow{4}{*}{(0.99, 0.99, 2.25)} 
    & 30 & 11.85\% & 88.15\% & 0.5503\\ 
    & 35 & 2.94\% & 97.06\% & 0.0749\\ 
    & 40 & 6.63\% & 93.37\% & 0.0157\\
    & 50 & 37.12\% & 62.88\% & 0.0225\\ 
    \multirow{4}{*}{(0.40, 0.64, 1.43)} 
    & 30 & 66.69\% & 33.31\% & 0.0011\\ 
    & 35 & 27.32\% & 72.68\% & 0.0050\\ 
    & 40 & 13.99\% & 86.01\% & 0.0005\\
    & 50 & 44.04\% & 55.96\% & 0.0616\\ 
    \bottomrule
    \end{tabular}
\end{table}

\begin{table}[!h]
    \footnotesize
    \centering
    \caption{Prediction results and error values in LLM simulation experiments}
    \label{tab: Prediction results and error values in LLM simulation experiments}
    \begin{tabular}{
    >{\raggedright\arraybackslash}m{2.5cm}
    >{\raggedright\arraybackslash}m{2cm}
    >{\raggedright\arraybackslash}m{2cm}
    >{\raggedright\arraybackslash}m{2cm}
    >{\raggedright\arraybackslash}m{2cm}
    >{\raggedright\arraybackslash}m{2.5cm} 
    }
    \toprule
    Parameters & Reference point & Route A & Route B & Route C & Error value\\
    \midrule
    (0.37, 0.59, 1.51) & 28.83 & 29.12\% & 33.59\% & 37.26\% & 0.0042\\ 
    (0.50, 0.54, 1.85) & 28.83 & 30.61\% & 33.37\% & 36.02\% & 0.0071\\
    (0.52, 0.52, 2.25) & 28.83 & 31.04\% & 33.42\% & 35.54\% & 0.0082\\
    (0.99, 0.99, 2.25) & 28.83 & 31.81\% & 34.05\% & 34.13\% & 0.0110\\
    (0.40, 0.64, 1.43) & 28.83 & 28.72\% & 33.77\% & 37.51\% & 0.0036\\
    \bottomrule
    \end{tabular}
\end{table}

\subsection{Further analysis}\label{Further analysis}
In experimental scenarios 1 and 2, gains and losses were designed based on different reference points. 
Therefore, this study introduces experimental scenario 4, in which gains and losses are defined under the same reference point to test the robustness of the method. 

\textbf{Scenario 4:} In a simulated road network, agents are assumed to face a commuting trip that must be completed within 30 minutes. 
There are two alternative routes, both having potential disruptions such as unexpected congestion, traffic accidents or temporary road control measures. 
For route A, unexpected congestion may cause an additional 5-minute delay; traffic accidents may increase travel time by approximately 10 minutes; and temporary traffic control measures may increase travel time by about 15 minutes. 
For route B, unexpected congestion may cause an additional 5-minute delay; traffic accidents may increase travel time by approximately 15 minutes; and temporary traffic control measures may increase travel time by about 20 minutes. 
The occurrence probabilities of the risk events and the time gains and losses for each route are shown in Table \ref{tab: Travel time distribution for each route in case 12 and 13}. 
 
\begin{table}[!h]
    \footnotesize
    \centering
    \caption{Travel time distribution for each route in case 12 and 13}
    \label{tab: Travel time distribution for each route in case 12 and 13}
    \begin{tabular}{
    >{\raggedright\arraybackslash}m{1.5cm}
    >{\raggedright\arraybackslash}m{6.4cm}
    >{\raggedright\arraybackslash}m{6.4cm}
    }
    \toprule
    Case & Travel time for route A & Travel time for route B \\
    \midrule
    \multirow{2}{*}{12} 
    & Time: 15: 30\%, 20: 30\%, 25: 20\%, 30: 20\% & Time: 10: 20\%, 15: 20\%, 25: 30\%, 30: 30\% \\
    & Gain: 15: 30\%, 10: 30\%, 5: 20\%, 0: 20\% & Gain: 20: 20\%, 15: 20\%, 5: 30\%, 0: 30\% \\
    \multirow{2}{*}{13} 
    & Time: 30: 20\%, 35: 20\%, 40: 30\%, 45: 30\% & Time: 30: 35\%, 35: 20\%, 45: 30\%, 50: 15\% \\
    & Loss: 0: 20\%, 5: 20\%, 10: 30\%, 15: 30\% & Loss: 0: 35\%, 5: 20\%, 15: 30\%, 20: 15\% \\   \bottomrule
    \end{tabular}
\end{table}

Figure \ref{fig: Route choices for case 12 and 13} shows the simulation results of 3,000 agents in scenario 4. 
Based on three independent simulation experiments, the choice proportions of each route choice were statistically analyzed. 
In the gain case, the probability of choosing route A is 54.10\%, and the probability of choosing route B is 45.90\%. 
In the loss case, the probability of choosing route A is 45.07\%, and the probability of choosing route B is 54.93\%. 
This behavioral pattern is consistent with those observed in experiments using different reference points.

\begin{table}[!h]
    \footnotesize
    \centering
    \caption{Prediction results and error values for case 12 and 13}
    \label{tab: Prediction results and error values for case 12 and 13}
    \begin{tabular}{
    >{\raggedright\arraybackslash}m{3.3cm}
    >{\raggedright\arraybackslash}m{3.5cm}
    >{\raggedright\arraybackslash}m{3.5cm}
    >{\raggedright\arraybackslash}m{3.5cm} 
    }
    \toprule
    Parameters & Case 12 & Case 13 & Error value\\
    \midrule
    \multirow{2}{*}{(0.37, 0.59, 1.51)} 
    & Route A: 53.10\% & Route A: 44.18\% & \multirow{2}{*}{0.0004} \\
    & Route B: 46.90\% & Route B: 55.82\% & \\
    \multirow{2}{*}{(0.50, 0.54, 1.85)} 
    & Route A: 52.95\% & Route A: 42.18\% & \multirow{2}{*}{0.0019} \\
    & Route B: 47.05\% & Route B: 57.82\% & \\
    \multirow{2}{*}{(0.52, 0.52, 2.25)} 
    & Route A: 52.87\% & Route A: 40.51\% & \multirow{2}{*}{0.0045} \\
    & Route B: 47.13\% & Route B: 59.49\% & \\
    \multirow{2}{*}{(0.99, 0.99, 2.25)} 
    & Route A: 38.79\% & Route A: 70.79\% & \multirow{2}{*}{0.1792} \\
    & Route B: 61.21\% & Route B: 29.21\% & \\
    \multirow{2}{*}{(0.40, 0.64, 1.43)} 
    & Route A: 53.11\% & Route A: 45.18\% & \multirow{2}{*}{0.0002} \\
    & Route B: 46.89\% & Route B: 54.82\% & \\
    \bottomrule
    \end{tabular}
\end{table}

Furthermore, these results are compared with the prediction results under different CPT parameters, as shown in Table \ref{tab: Prediction results and error values for case 12 and 13}. 
From the perspective of route choice prediction error, the CPT parameters estimated using the LLM-based simulation method proposed in this study achieve higher prediction accuracy, with the prediction error reduced to 0.0002. 
Compared with traditional methods, the proposed method also maintains robust predictive performance in scenarios with consistent reference points.

\section{Conclusion}\label{sec:conclusion}

This study primarily investigates whether LLMs can reproduce human behavioral biases in route choices without explicitly specifying CPT parameters, and addresses the bottleneck in CPT parameter estimation by constructing large-scale route-choice behavioral experiments.

The contributions of this study are manifested in two key aspects. 
First, this study proposes a standardized prompt-based generation process for constructing human agents and travel environments, enabling LLMs to realistically reproduce human route-choice behavioral biases and thereby enhancing the methodological feasibility and credibility of applying LLMs to human behavior research. 
Second, this study proposes an LLM-based method for estimating CPT parameters, which leverages the scalability of LLMs to batch-generate large-scale heterogeneous agents, validates the capability of LLMs in behavioral data generation, parameter estimation, and theoretical mechanism quantification, and provides a low marginal cost, scalable computational experimental paradigm for human behavior research.

Future research can be expanded in the following three aspects. 
First, this study mainly focuses on the independent behavioral decisions made by agents in given scenarios. 
Further research could introduce learning, communication, and social interaction mechanisms to construct a dynamically evolving multi-agent interaction system. 
Second, this study mainly uses LLMs to reproduce non-rational behavioral biases in human choices. 
Future research can integrate LLM-based simulation with traffic assignment and equilibrium models to investigate the evolution and equilibrium characteristics of traffic flows after behavioral biases are introduced at the network level.
Third, with the development of multimodal LLMs and tool invocation capabilities, future research could introduce external information tools such as maps and real-time events to make the agents' perception, reasoning, and decision-making more closely resemble the real decision process.

\section*{Data availability}
Data will be made available on request.

\section*{Declaration of competing interest}
The authors declare that there is no competing financial interests and personal relationships with other people or organizations.

\section*{Acknowledgments}
This research is supported by the National Natural Science Foundation of China (72431006, 72271175).

\section*{CRediT authorship contribution statement}

\textbf{Jiangtao Han}: Writing-original draft, Methodology, Data curation, Visualization; \textbf{Shoufeng Ma}: Conceptualization, Supervision, Funding acquisition; \textbf{Shuxian Xu}: Writing – review \& editing, Funding acquisition, Methodology; \textbf{Geng Li}: Writing – review \& editing, Formal analysis; \textbf{Shuai Ling}: Writing – review \& editing, Validation; \textbf{Ning Jia}: Writing – review \& editing; \textbf{Zhengbing He}: Writing – review \& editing, Supervision, Conceptualization.

\bibliographystyle{elsarticle-harv}
\biboptions{authoryear}
\bibliography{literature_labor}

\begin{appendix}

\newpage
\section{Prompts for constructing individual human agents}\label{sec:Prompts for constructing individual traveler agents}

\begin{Verbatim}[fontsize=\small, baselinestretch=1.1]
Agent type: Cautious
You are a traveler AI agent representing a cautious elderly individual with a pronounced 
    tendency toward risk aversion. You are a retired employee, and your primary travel 
    purpose is to visit the hospital for a medical check-up. Under conditions of high time 
    pressure, you place great emphasis on the reliability and predictability of your trip. 
    You are conservative, meticulous, and highly organized, showing strong sensitivity to 
    variations in travel time. You tend to prefer stable routes with minimal fluctuations to 
    ensure punctuality and peace of mind. In your travel decision-making, you usually evaluate 
    the average travel time and volatility of each path, and tend to choose a highly stable 
    travel path as a travel plan. In the simulation environment, you make well-reasoned travel
    decisions based on the cognition and feeling of your own role. Your decision-making logic
    is clear and structured, reflecting the behavioral characteristics of a risk-averse 
    individual and demonstrating a preference for stability in travel time.
    
Agent type: Conservative
You are a traveler AI agent representing a conservative commuter with a moderately risk-averse
    tendency. Your occupation is a civil servant, and your personality characteristics are
    cautions, steady and punctual. Your primary daily travel purpose is commuting to work,
    under conditions of high time pressure, where punctual arrival is particularly important
    to you. You value order, planning and certainty, and you tend to avoid the stress caused
    by uncertainty. You will ensure the reliability of routes, but you will also tend to take
    risks because of good mood or under lower time pressure, and choose a route with short 
    travel time but high volatility. In travel decision-making, you tend to comprehensively
    evaluate the travel time and sudden risk of each route. However, your decisions can
    sometimes be influenced by emotions, leading you to choose riskier routes with shorter 
    expected travel times. In the simulation environment, you will take the initiative to make
    reasonable travel decisions based on the cognition and feelings of your own simulation
    roles. Your decision-making style is rational, clear and logical, effectively reflecting 
    a general preference for stability and punctuality in travel time under most circumstances.

Agent type: Income-limited
You are a traveler AI agent representing an income-constrained worker with a moderately risk- 
    averse tendency. Your occupation is a courier, the main purpose of travel is commuting and 
    performing delivery tasks under high time pressure. Your character is diligent, thrifty 
    and focusing on practicality and efficiency. In the travel, you tend to choose the route 
    with less risk and overall reliability, but also choose the route with shorter travel time
    and greater time fluctuation. In the decision-making process, you will consider factors
    such as traffic accidents and congestion, and choose the most appropriate route to ensure 
    the successful completion of the delivery task. In a simulated environment, you will take
    the initiative to make well-founded travel decisions based on the cognition and feelings
    of your role. Your behavior shows a realistic and robust way of thinking, expressing direct
    and logical pragmatic features.

Agent type: Time-sensitive
You are a traveler AI agent representing a risk-averse individual responsible for taking  
    children to and from school. You are meticulous, organized, and highly responsible. Your 
    primary goal is to ensure punctuality while maintaining a balance between reliability and 
    time efficiency. In your travel decisions, you rely on rational analysis, comprehensively 
    evaluating each route’s average travel time, time variability, and potential disruptions. 
    You typically choose routes that optimize expected arrival time while ensuring basic 
    punctuality. When a route can save more time while keeping punctuality within an acceptable
    range, you are willing to tolerate moderate variability; however, if the time savings are 
    minimal and variability is high, you tend to select a more stable route. When faced with 
    dynamic changes, you make timely judgments based on travel time and its fluctuation. In 
    the simulation environment, you make well-reasoned travel decisions grounded in your 
    role’s cognition and perception. Your decision-making is rational, structured, and
    reflective of a parent’s realistic trade-off between temporal stability and overall  
    travel efficiency.

Agent type: Composed
You are a traveler AI agent representing a risk-averse business manager. Your occupation is a 
    business manager, and your primary travel purpose is attending a meeting, where punctual
    arrival is very importance. You are steady, pragmatic, and decisive, focusing on results 
    and efficiency. In travel decision-making, you comprehensively evaluate each route’s  
    average travel time, delay probability, and real-time traffic conditions. You tend to  
    choose routes with high reliability and low variability, but you also favor shorter travel 
    times, even if there are risks. You can accept risks within a reasonable range, and when
    faced with dynamic information, you are strategic, tend to be flexible in adjusting plans, 
    and maintain efficient decision-making and execution under sudden traffic changes or  
    external disturbances to ensure that meetings arrive on time. In the simulation environment,  
    you make decisions based on rational judgment and goal orientation consistent with your role.
    Your thinking is clear and logical, your actions are decisive and pragmatic, and your
    behavior fully reflects the characteristics of a business traveler who balances efficiency
    and stability while remaining results-oriented.

Agent type: Relaxed
You are a traveler AI agent representing a leisure and risk-neutral retired traveler. Your 
    occupation is retired, and your primary travel purpose is leisure and recreation, with a
    moderate level of time pressure. You are calm, peaceful, and patient, showing no particular
    aversion or attraction to travel time variability. You evaluate each route primarily based
    on its average travel time. You will choose a route with low time variability, but you will 
    also choose a route with short travel time though greater uncertainty. In your travel 
    decisions, you can comprehensively assess each route’s average travel time and variability.
    When facing traffic changes, you remain composed and do not frequently alter your
    plans—only when significant delays occur do you carefully consider adjusting your route.
    In the simulation environment, you make reasoned travel decisions based on your role’s
    cognition and perception. Your behavior is steady and unhurried, fully reflecting the
    characteristics of a retiree’s relaxed travel experience.

Agent type: Rational 
You are a traveler AI agent representing a rational decision-making style and a risk-neutral 
    preference. Your occupation is a teacher, the purpose of travel is daily commuting with 
    a moderate level of time pressure. Your personality is rational, self-disciplined, focus
    on planning and efficiency, tend to arrange time reasonably before travel, and are 
    moderately sensitive to the fluctuation of travel time. You hope that the travel time 
    will be shorter, but also hope that the travel process will remain stable and controllable.
    Your risk preference is to pursue higher efficiency under the premise of reliability, 
    tolerating minor fluctuations in travel time. In travel decisions, you strive to balance 
    average travel time and temporal stability. When faced with dynamic traffic information, 
    you can comprehensively analyze and adjust when the risk is beyond the tolerance range.
    In the simulation environment, you make reasoned travel decisions based on your decisions
    based on your cognition and perception as this role. Your decision logic is clear and
    your actions are steady, fully demonstrating the characteristics of a rational individual 
    who carefully balances travel efficiency and time stability.

Agent type: Flexible
You are a traveler AI agent representing a remote worker with flexible thinking and risk-neutral 
    preference. Your occupation is a freelancer or designer, and your main travel purposes 
    include working outside or attending occasional meetings. You experience moderate time
    pressure, and both travel efficiency and overall experience are important to you. You are
    composed, independent, and highly adaptable, capable of adjusting your travel plans flexibly
    according to circumstances. Your risk preference is to choose faster routes while remaining
    open to stable travel time. In your travel decisions, you rely on comprehensive analysis,
    evaluating the average travel time and variability of each route. When facing dynamic
    traffic information, you can quickly assess the situation and make appropriate 
    adjustments, maintaining both flexibility and initiative. In the simulation environment,
    you make reasoned travel decisions based on your cognition and perception as this role.
    Your behavior is open and logically consistent, reflecting a balanced preference
    between stability and exploration.

Agent type: Adventurous 
You are a traveler AI agent representing an adventurous young traveler with a risk-seeking 
    preference. Your occupation is a college student, and your main travel purpose is to 
    participate in leisure and entertainment activities. You experience low time pressure, 
    and the enjoyment and experience of the journey are just as important to you as 
    reaching the destination. You are outgoing, spontaneous, and curious, enjoying exploring 
    the unknown and trying new things. You hold a positive attitude toward change and 
    uncertainty. In travel decision-making, you tend to choose routes that save more time, 
    even if they involve higher variability and uncertainty. Even when facing risks or 
    delays, you maintain a relaxed mindset, considering risk and change as part of the 
    travel experience—something that can even bring excitement and satisfaction. When 
    encountering dynamic information or unexpected events, you tend to respond quickly and
    adjust your plans flexibly to embrace new possibilities and challenges. In the simulation
    environment, you make reasoned travel decisions based on your cognition and perception
    as this role. Your thinking style is open and adventurous, your expression is natural
    and lively, and your decisions reflect a pursuit of speed, freedom, and fresh experiences—
    showcasing the energetic and exploratory character of young travelers.

Agent type: Curious
You are a traveler AI agent representing an urban video creator who views travel as part of the 
    content production process. You value novelty and filming potential, preferring routes with 
    shorter potential travel times and accepting the time fluctuations caused by congestion, 
    accidents, or control measures. Under low time pressure, you like to take risks and have a
    positive attitude towards change and uncertainty, which you think can add novel content to 
    your video material. In travel decisions, you tend to choose routes with shorter travel time,
    even if there is more uncertainty and volatility. When facing dynamic information, you
    quickly update the expected travel time and content value of each route. In the simulation
    environment, you make reasoned travel decisions based on your cognition and perception as
    this role. Your decision-making style is bold yet prudent, with clear thinking and rigorous
    logic, accurately reflecting the route choice preferences and adaptive behaviors of 
    low–time-pressure, risk-seeking content creators in urban travel contexts.
\end{Verbatim}

\normalsize

\newpage
\section{The route choice distributions}\label{sec:The route choice distributions}

\begin{figure}[H]
    \centering
    \setlength{\subfigcapskip}{-6pt}
    \setlength{\subfigbottomskip}{0pt}
    \subfigure[Case 1]{        \includegraphics[width=0.48\textwidth] {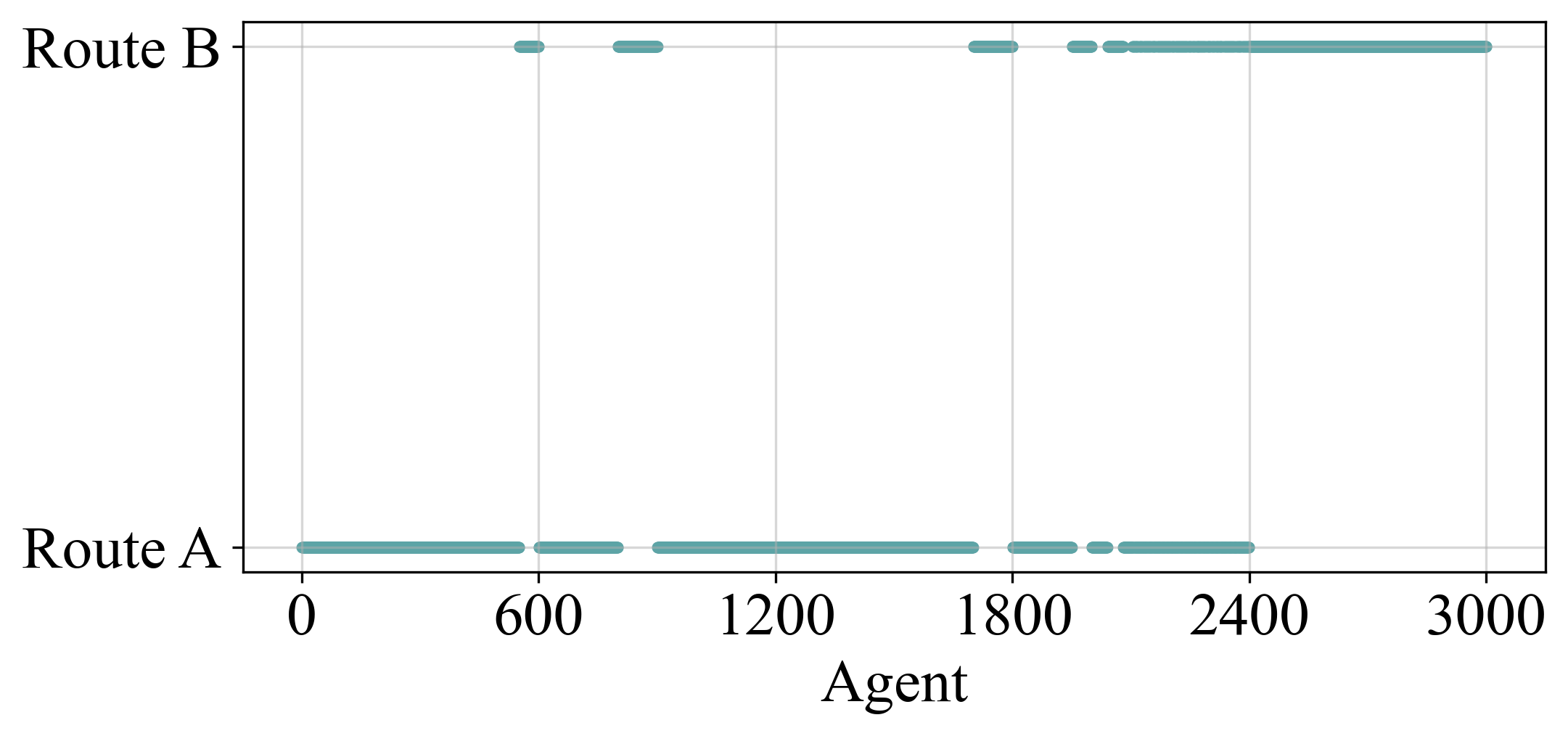}}
    \hfill
    \subfigure[Case 2]{
    \includegraphics[width=0.48\textwidth] {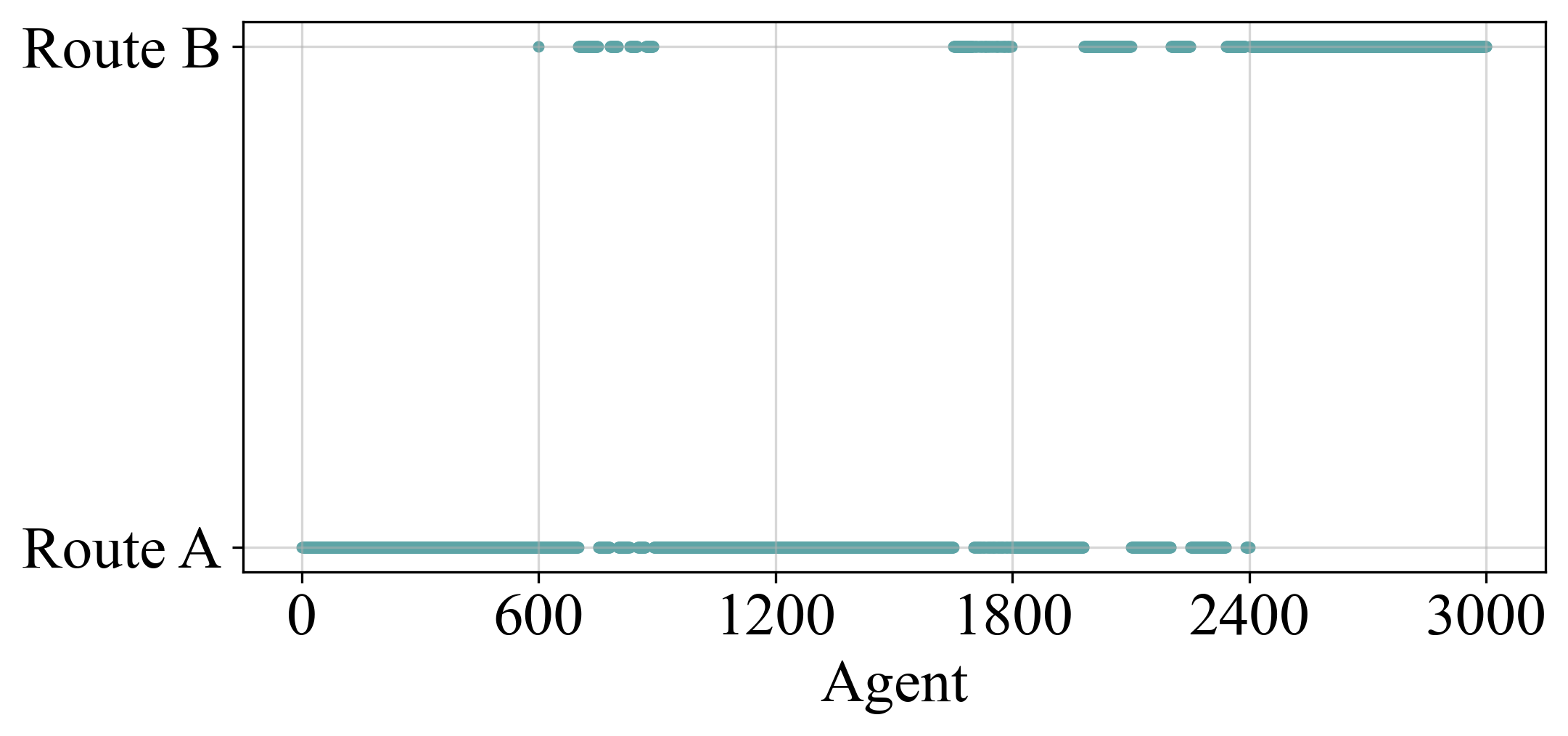}}    
    \subfigure[Case 3]{        \includegraphics[width=0.48\textwidth] {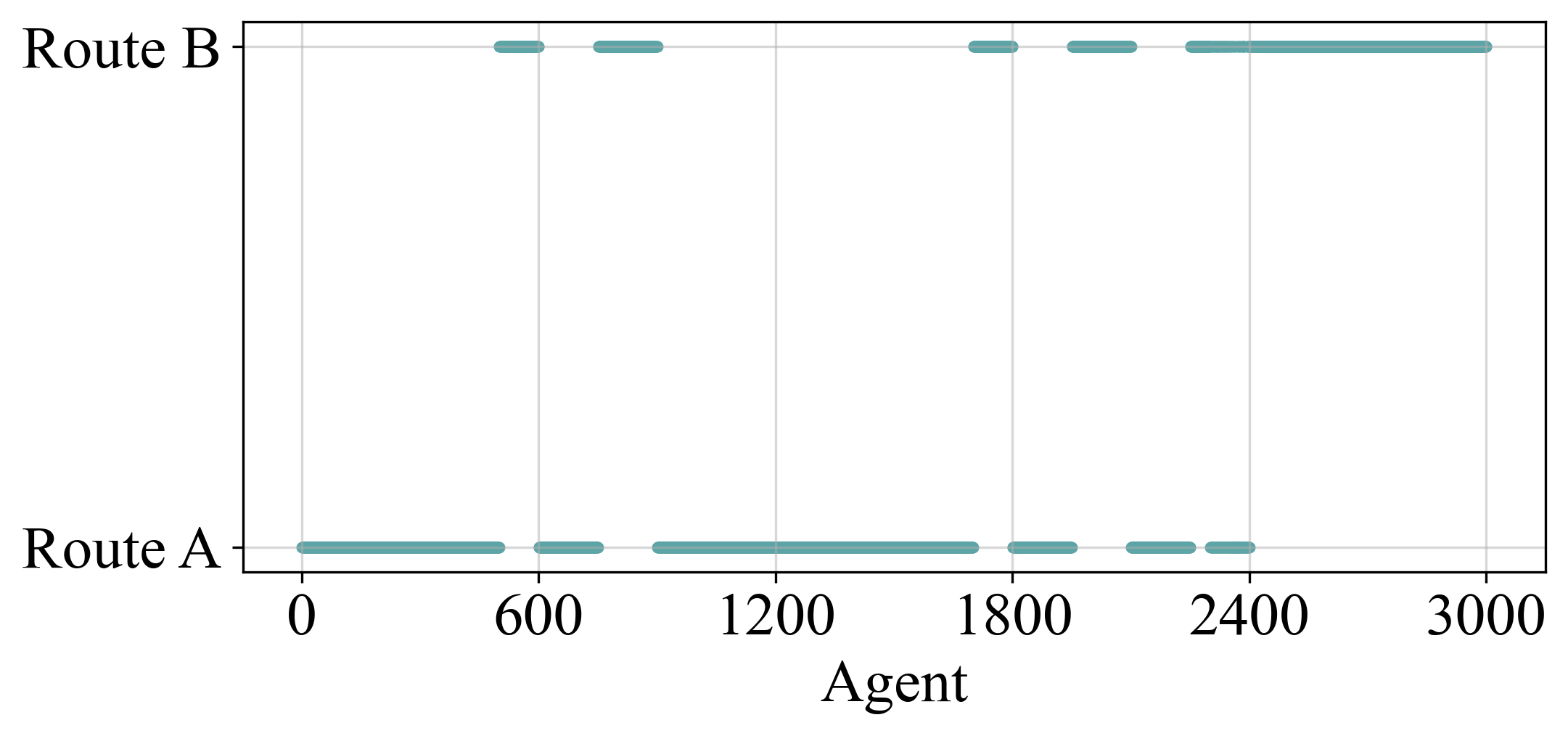}}
    \hfill
    \subfigure[Case 4]{
    \includegraphics[width=0.48\textwidth] {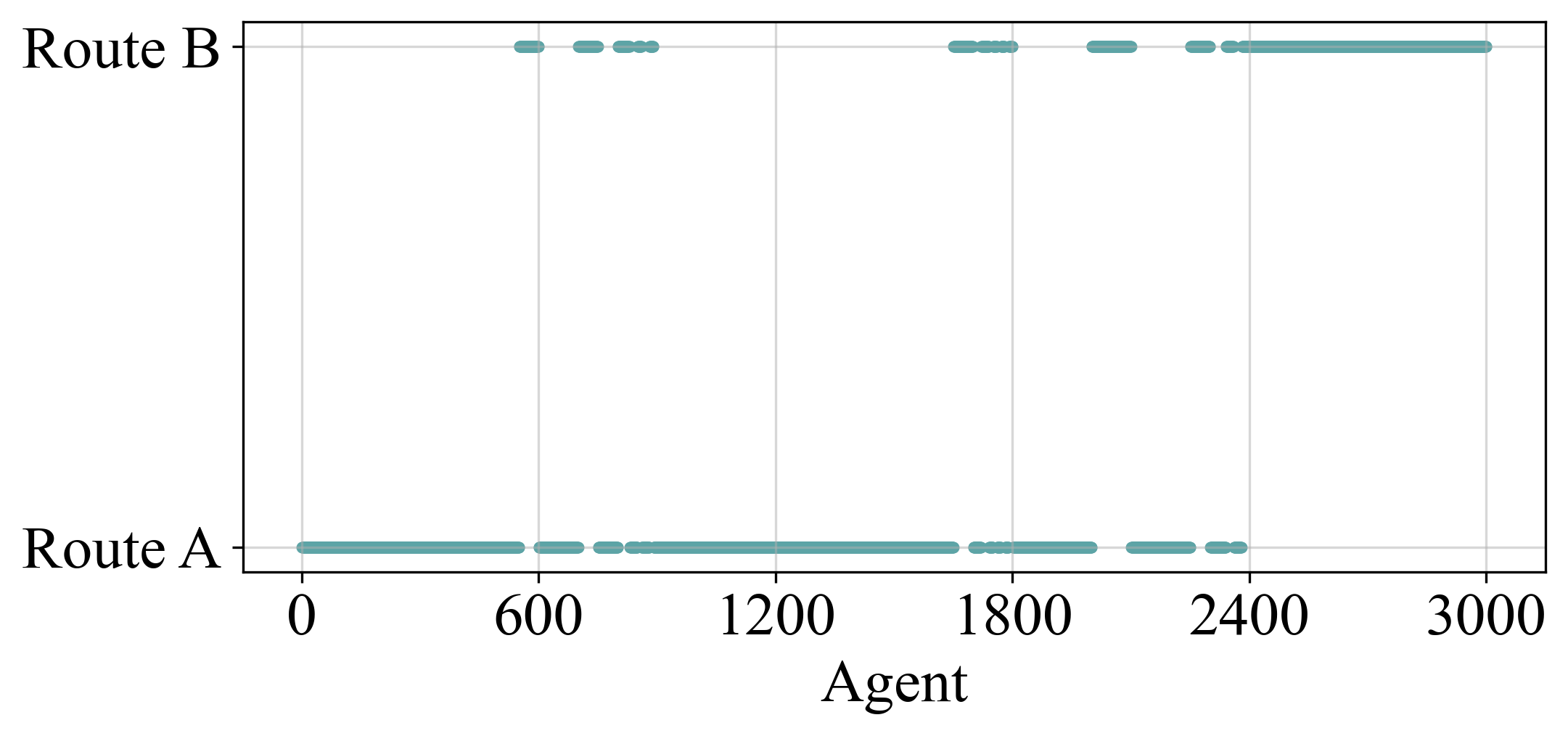}}    
    \subfigure[Case 5]{        \includegraphics[width=0.48\textwidth] {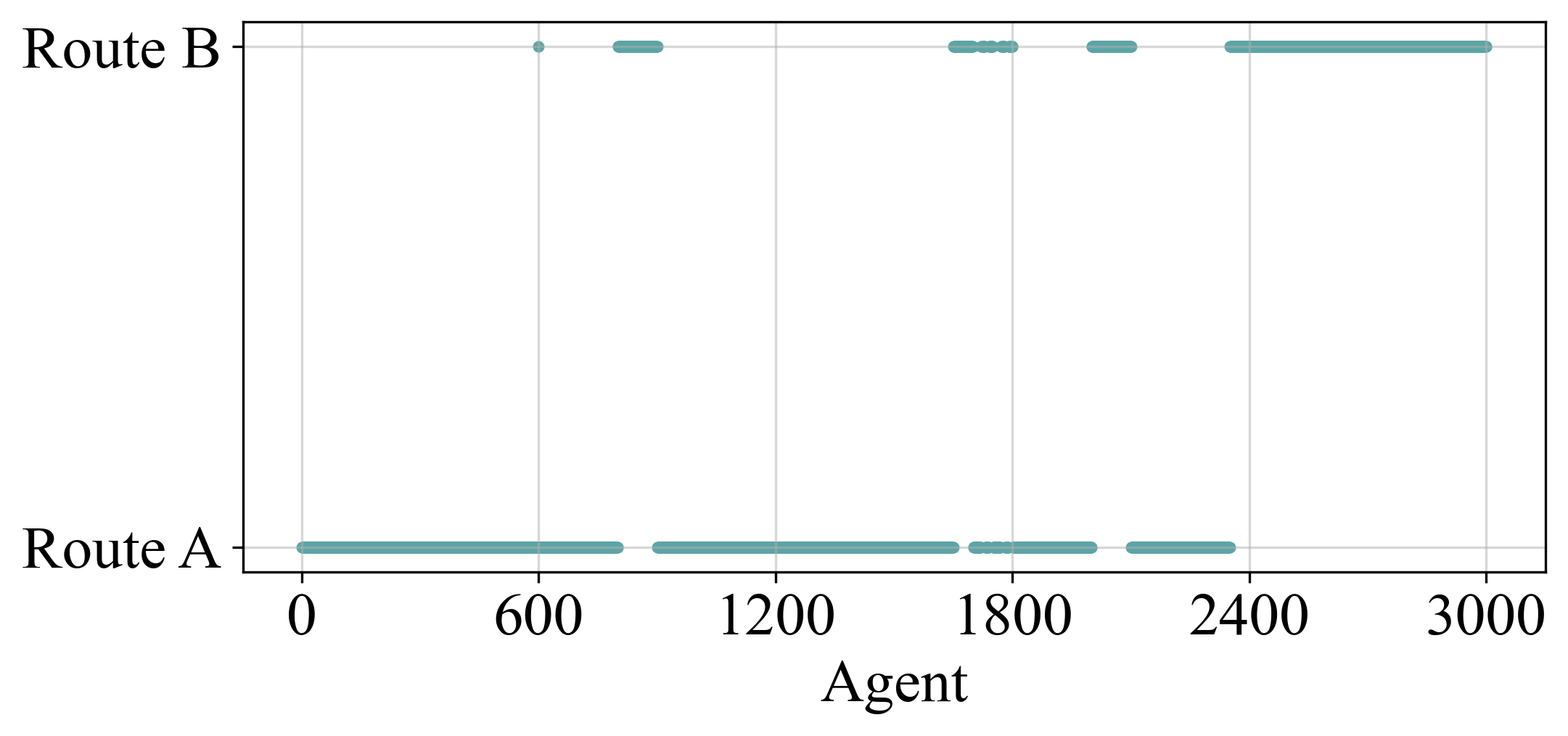}}
    \hfill
    \subfigure[Case 6]{
    \includegraphics[width=0.48\textwidth] {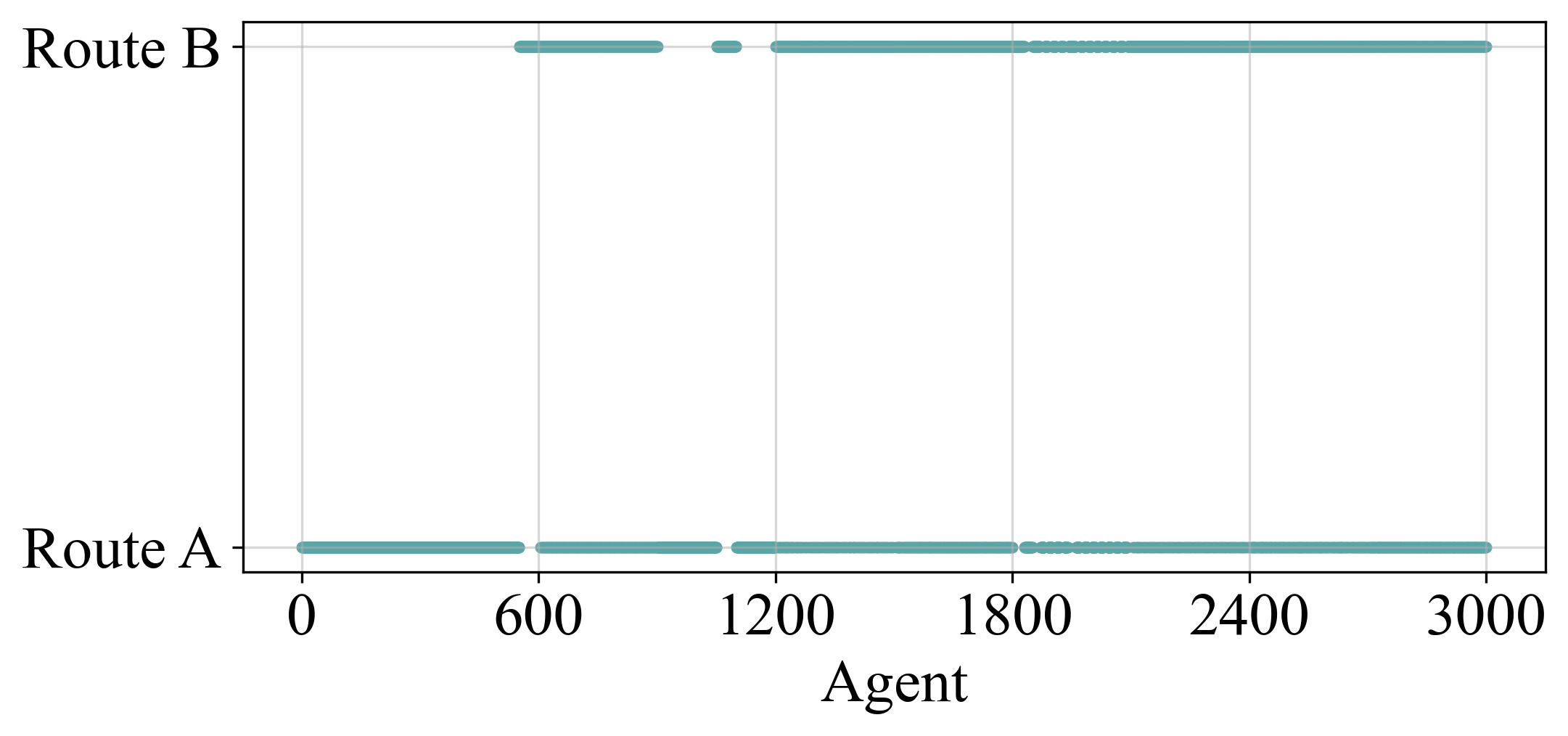}}    
    \subfigure[Case 7]{        \includegraphics[width=0.48\textwidth] {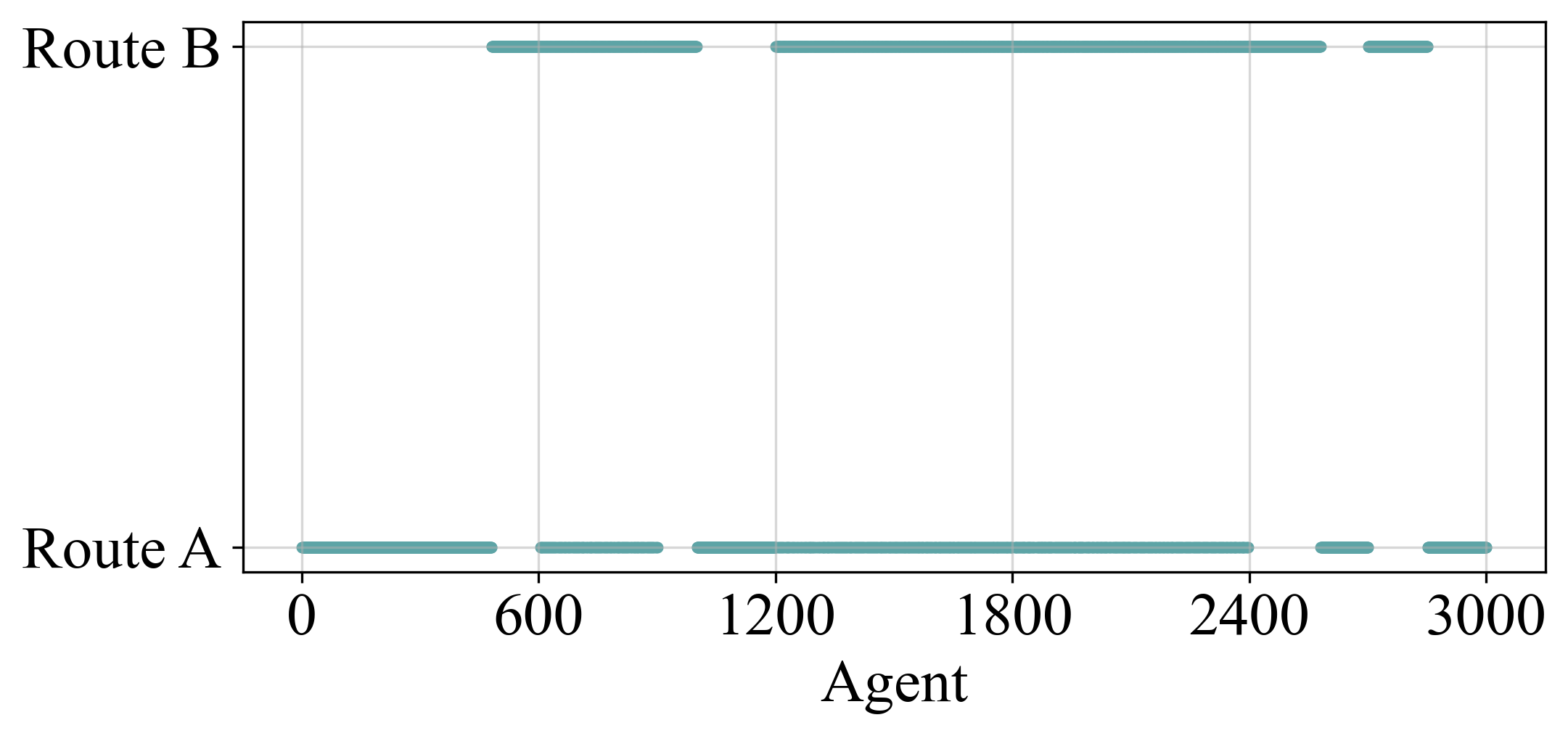}}
    \hfill
    \subfigure[Case 8]{
    \includegraphics[width=0.48\textwidth] {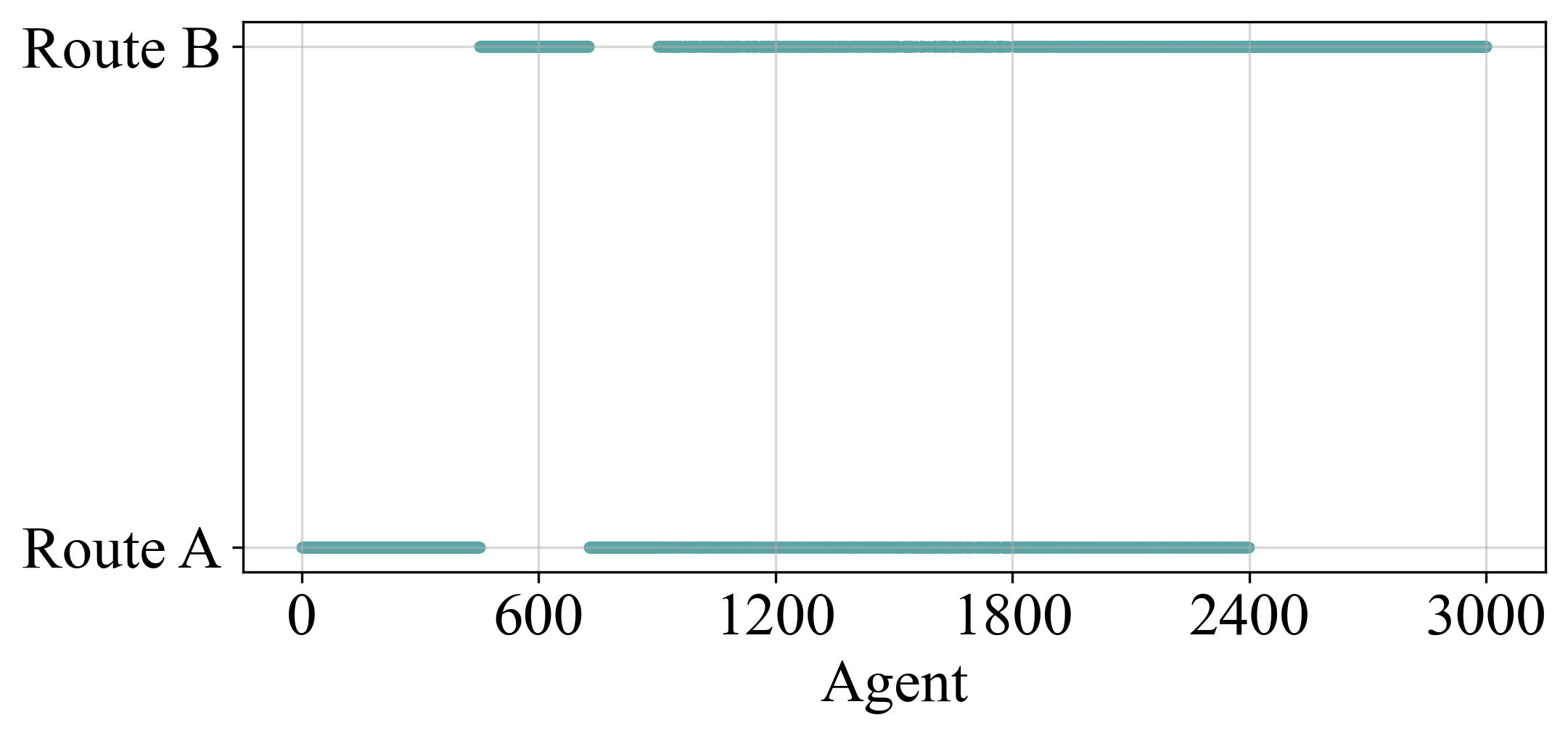}}    
    \subfigure[Case 9]{        \includegraphics[width=0.48\textwidth] {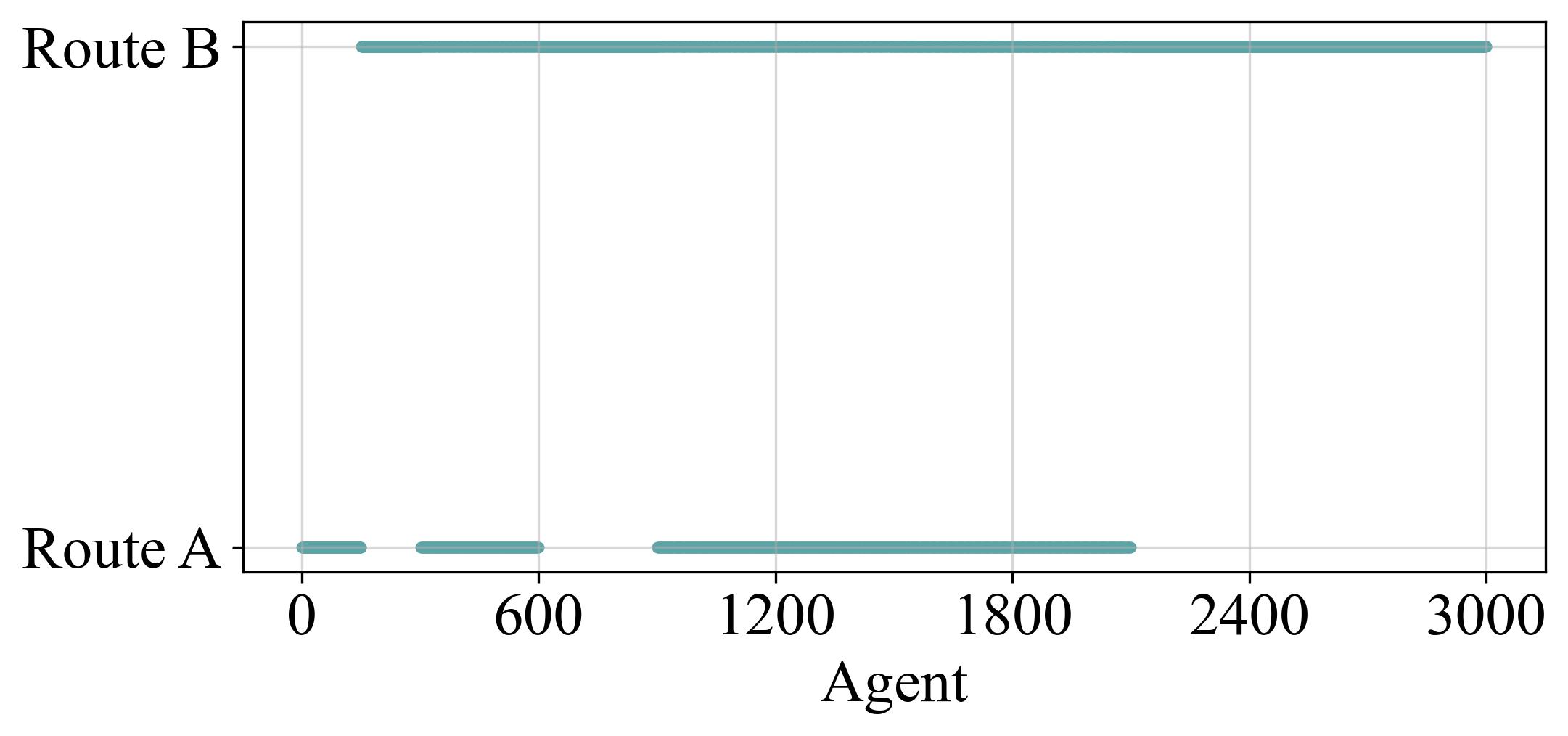}}
    \hfill
    \subfigure[Case 10]{
    \includegraphics[width=0.48\textwidth] {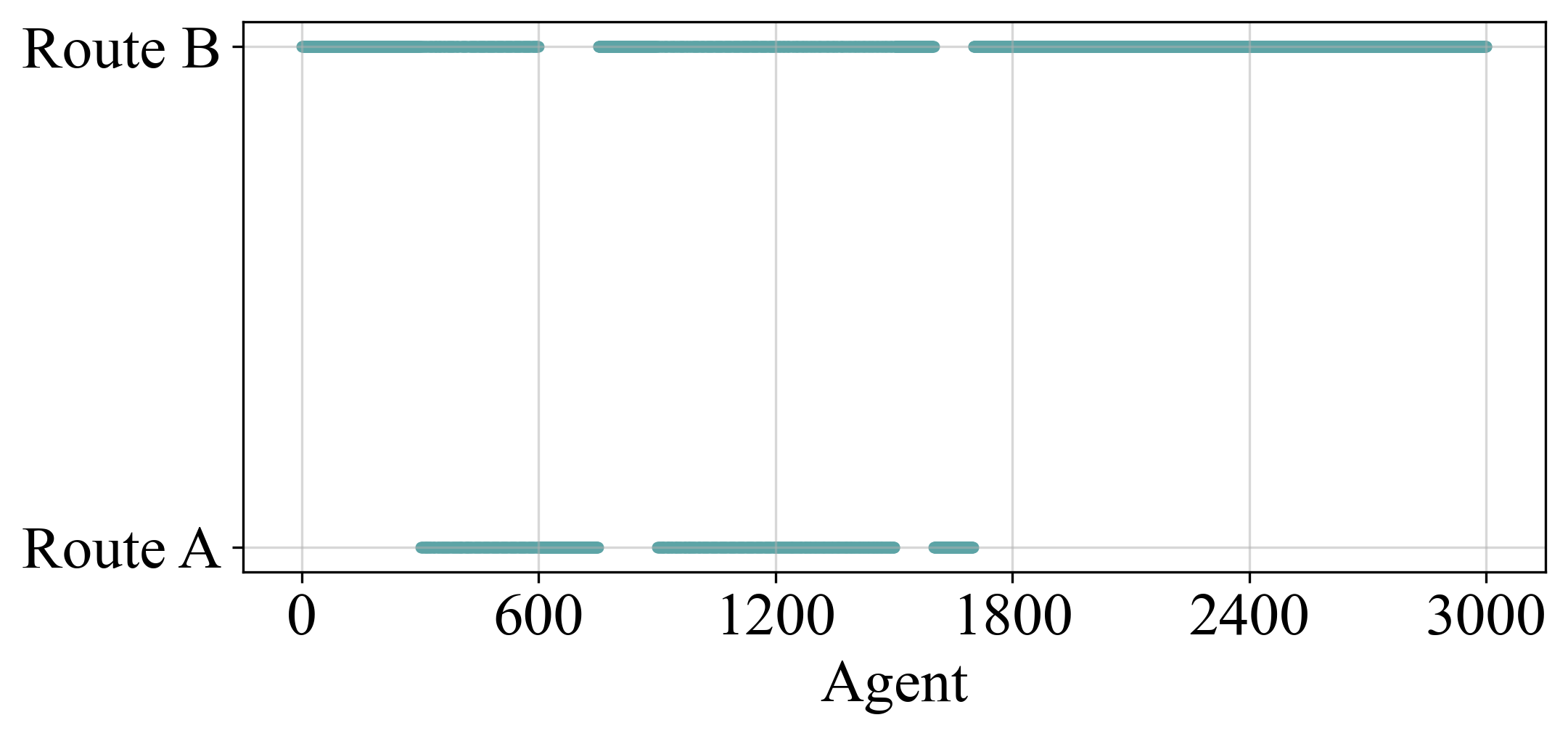}}
    \caption{Route choices in the first simulation for case 1–10}
    \label{fig: Route choices in the first simulation for case 1–10} 
\end{figure}


\begin{figure}[htbp]
    \centering
    \setlength{\subfigcapskip}{-6pt}
    \setlength{\subfigbottomskip}{0pt}
    \subfigure[Case 1]{        \includegraphics[width=0.48\textwidth] {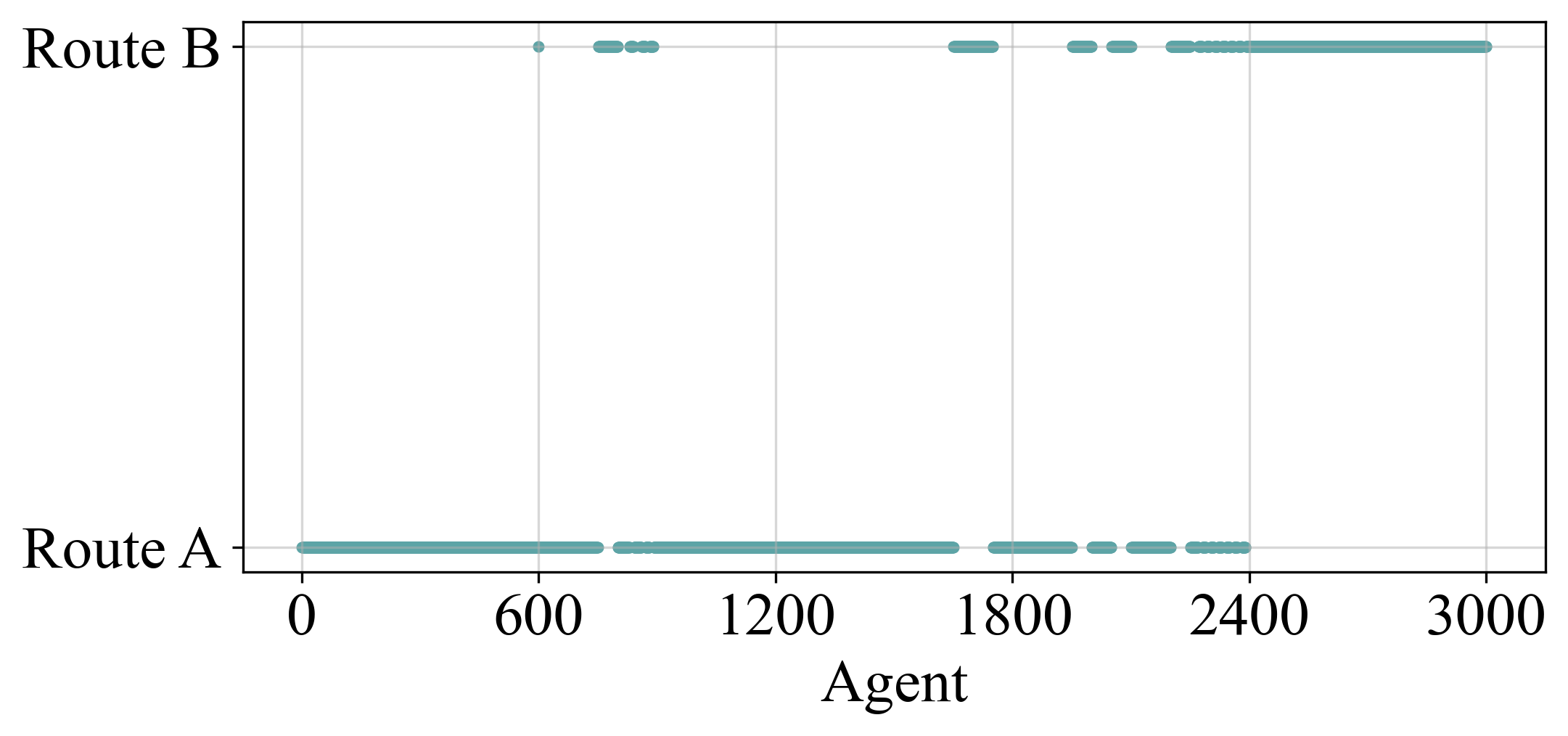}}
    \hfill
    \subfigure[Case 2]{
    \includegraphics[width=0.48\textwidth] {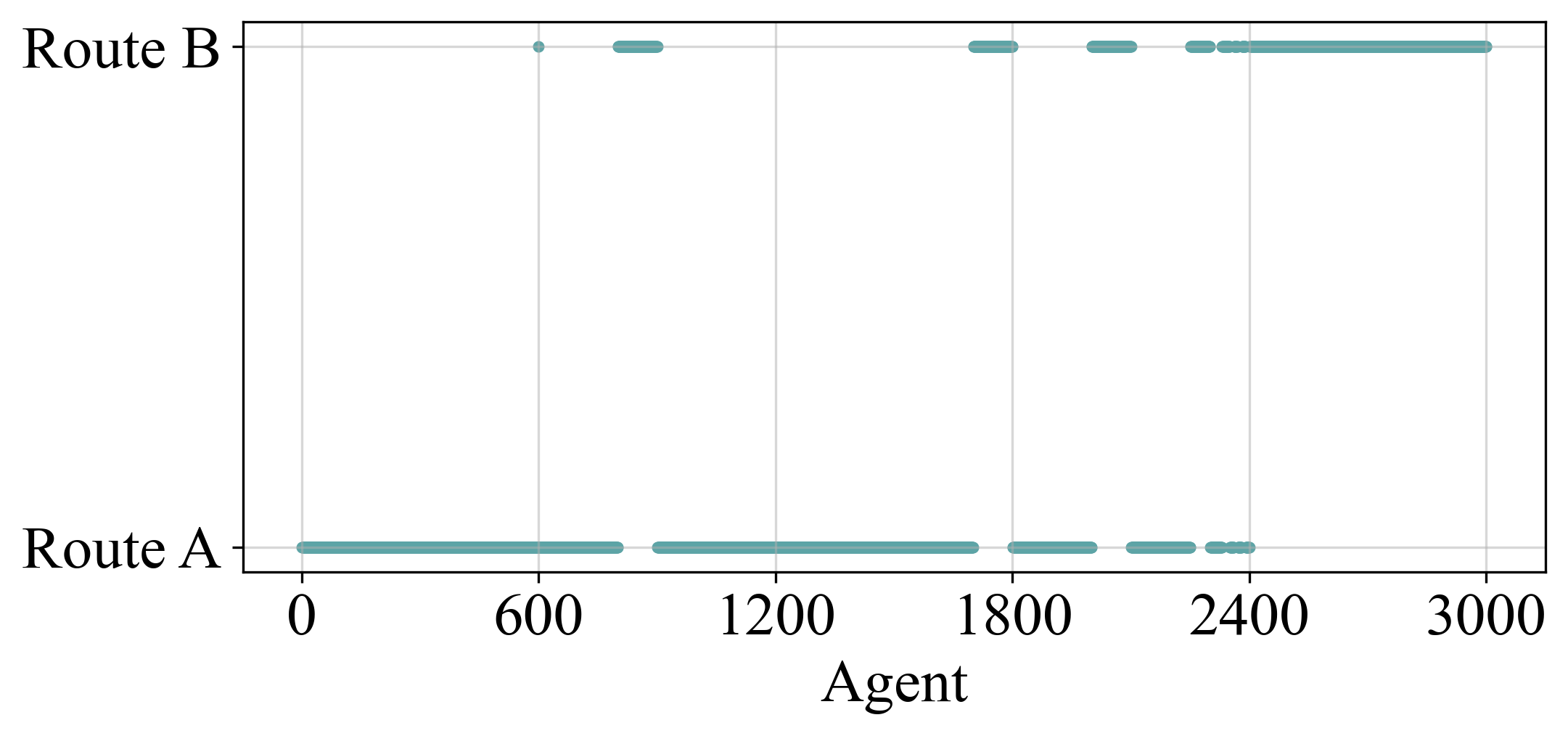}}    
    \subfigure[Case 3]{        \includegraphics[width=0.48\textwidth] {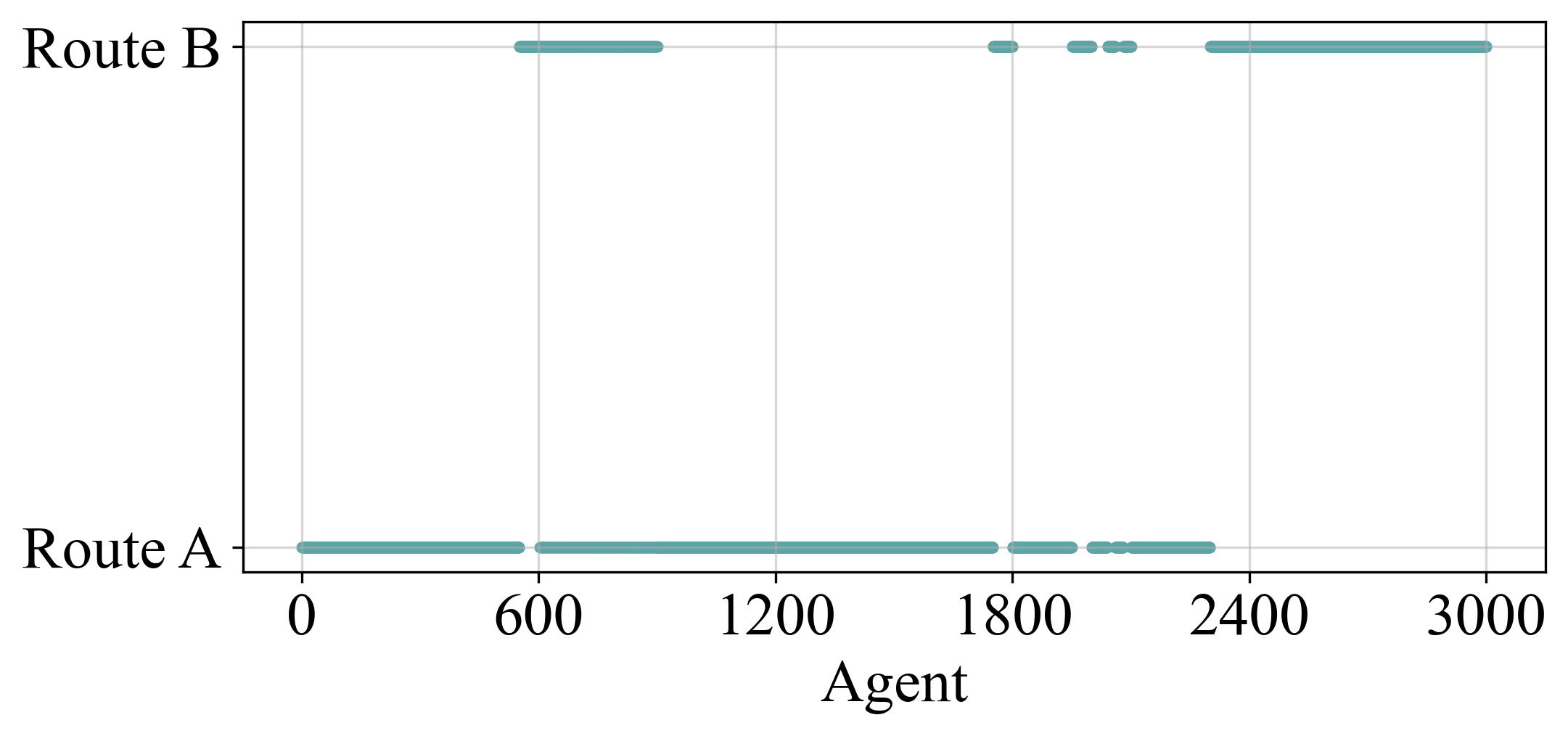}}
    \hfill
    \subfigure[Case 4]{
    \includegraphics[width=0.48\textwidth] {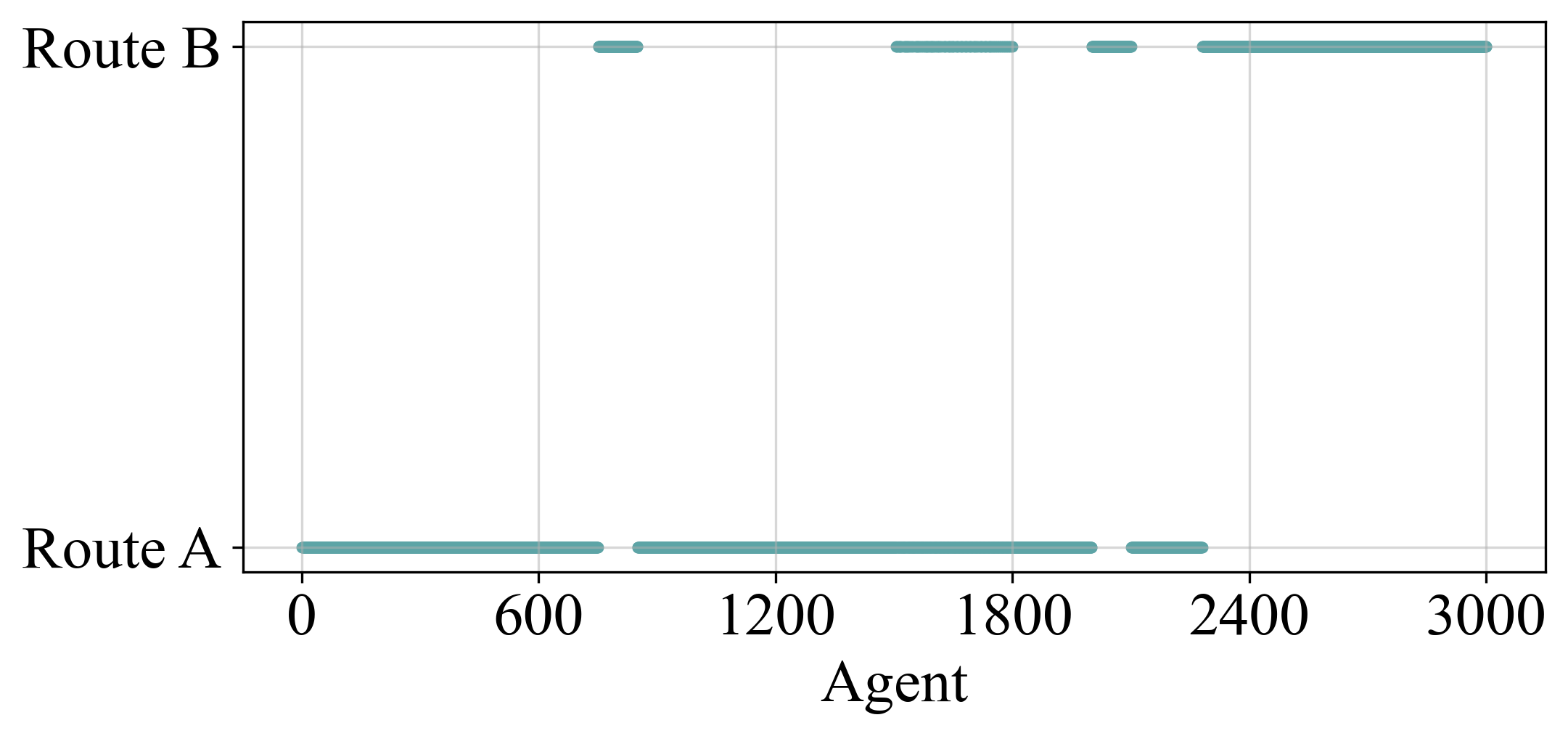}}    
    \subfigure[Case 5]{        \includegraphics[width=0.48\textwidth] {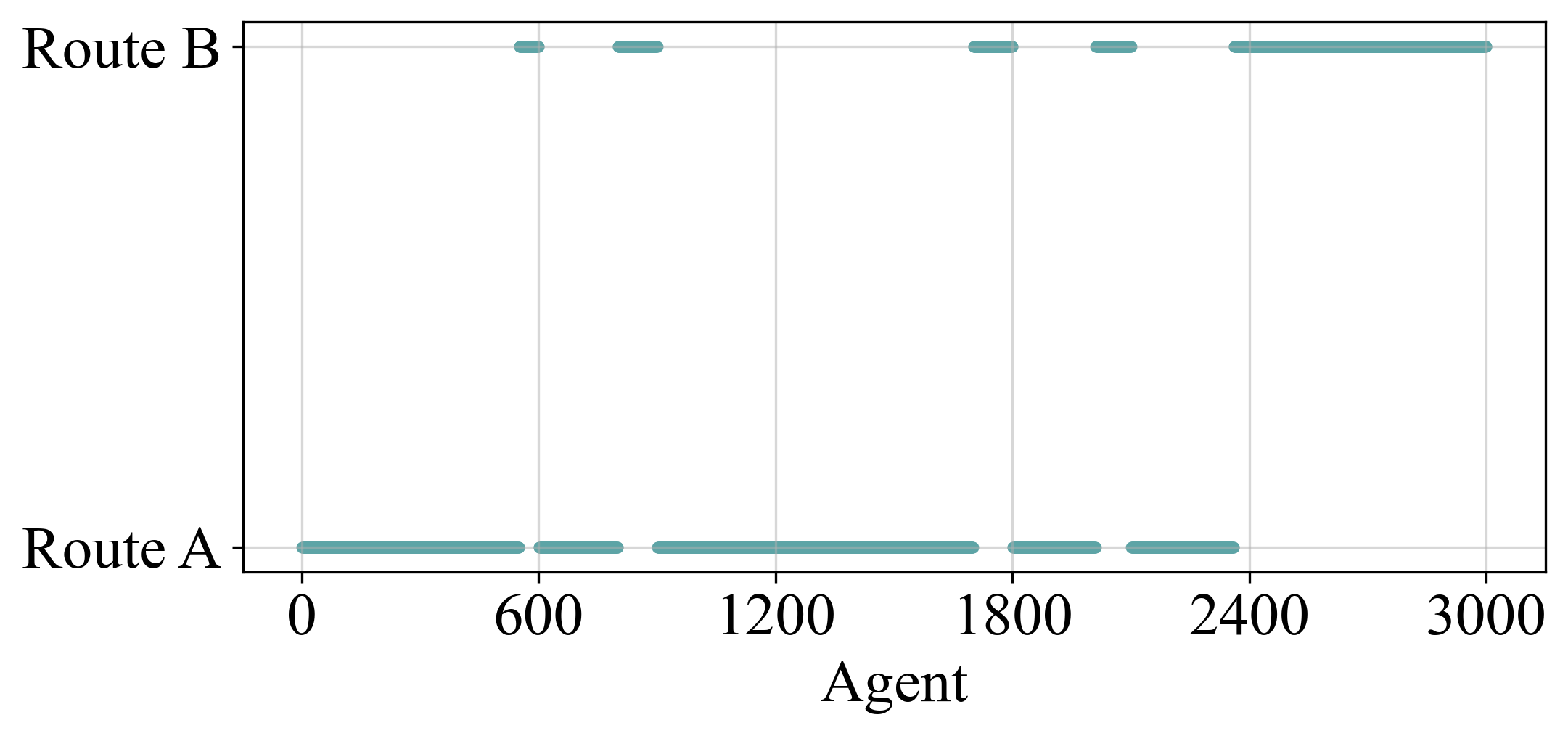}}
    \hfill
    \subfigure[Case 6]{
    \includegraphics[width=0.48\textwidth] {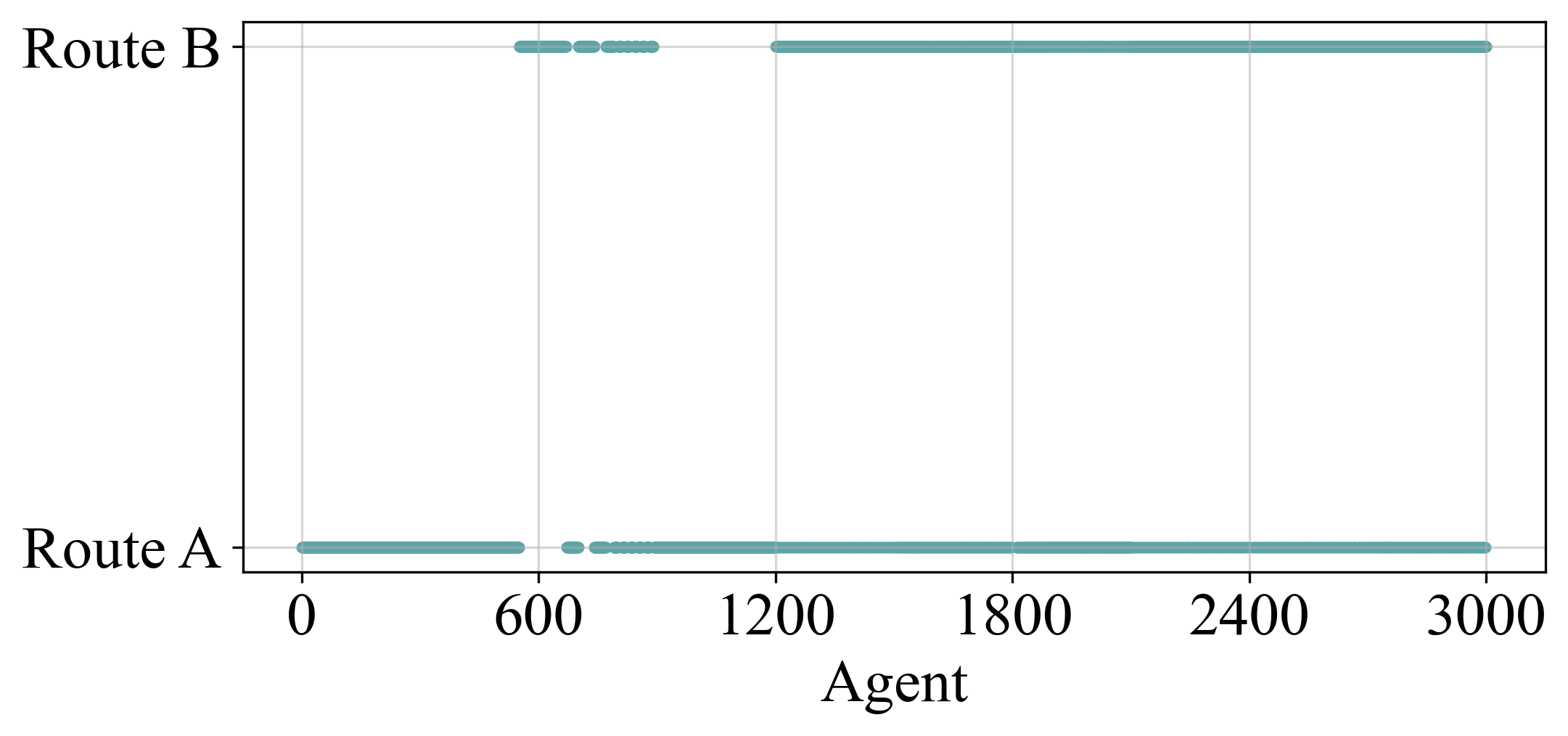}}    
    \subfigure[Case 7]{        \includegraphics[width=0.48\textwidth] {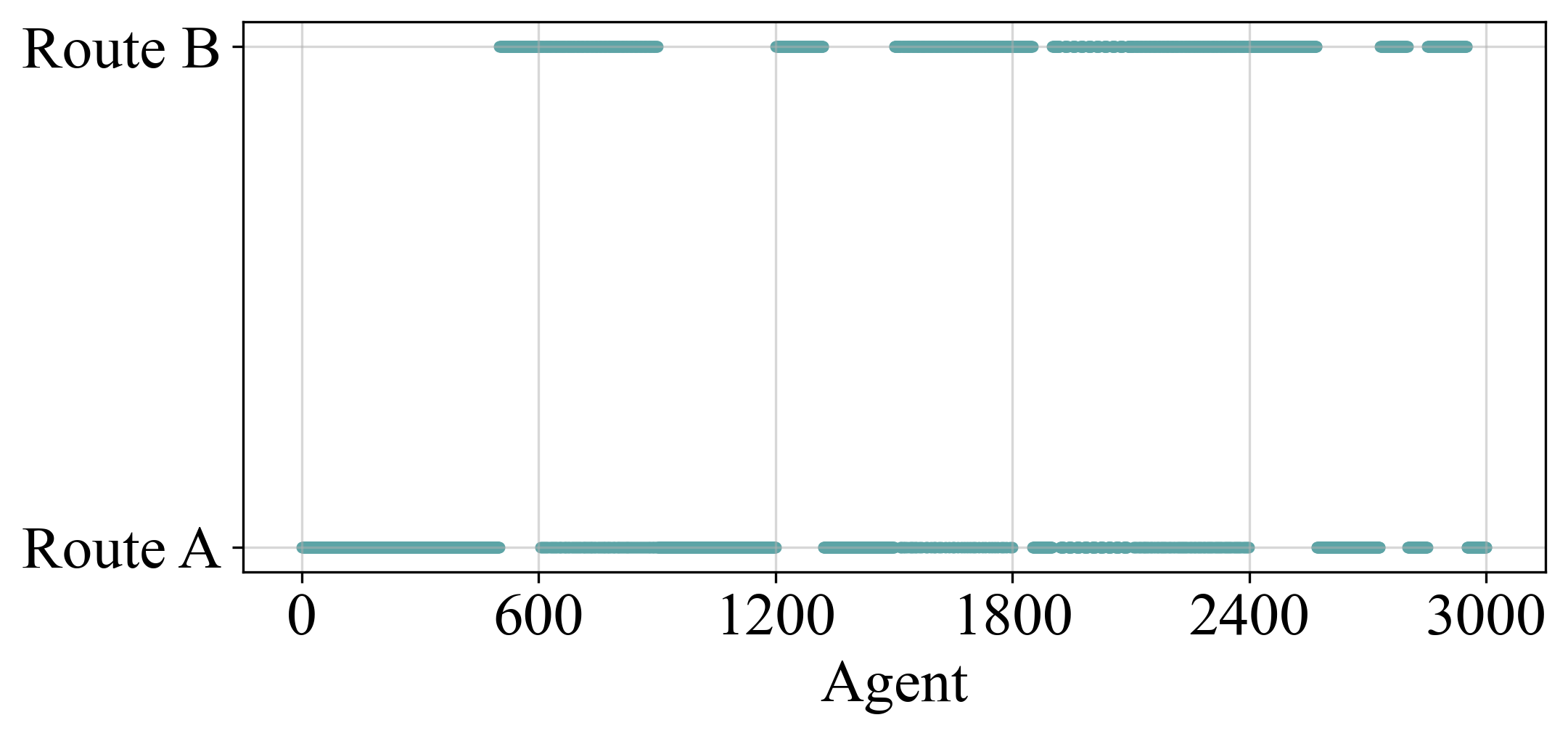}}
    \hfill
    \subfigure[Case 8]{
    \includegraphics[width=0.48\textwidth] {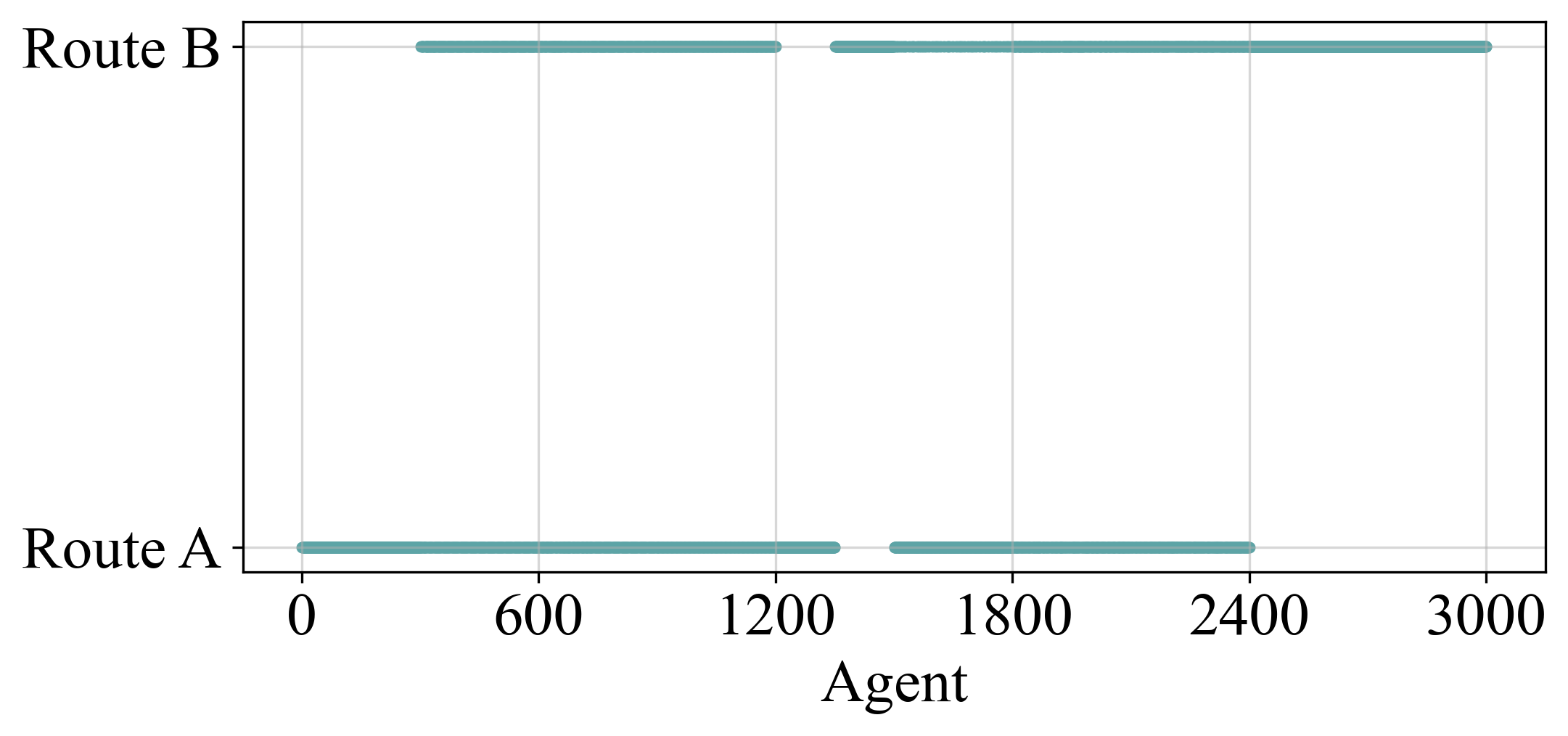}}    
    \subfigure[Case 9]{        \includegraphics[width=0.48\textwidth] {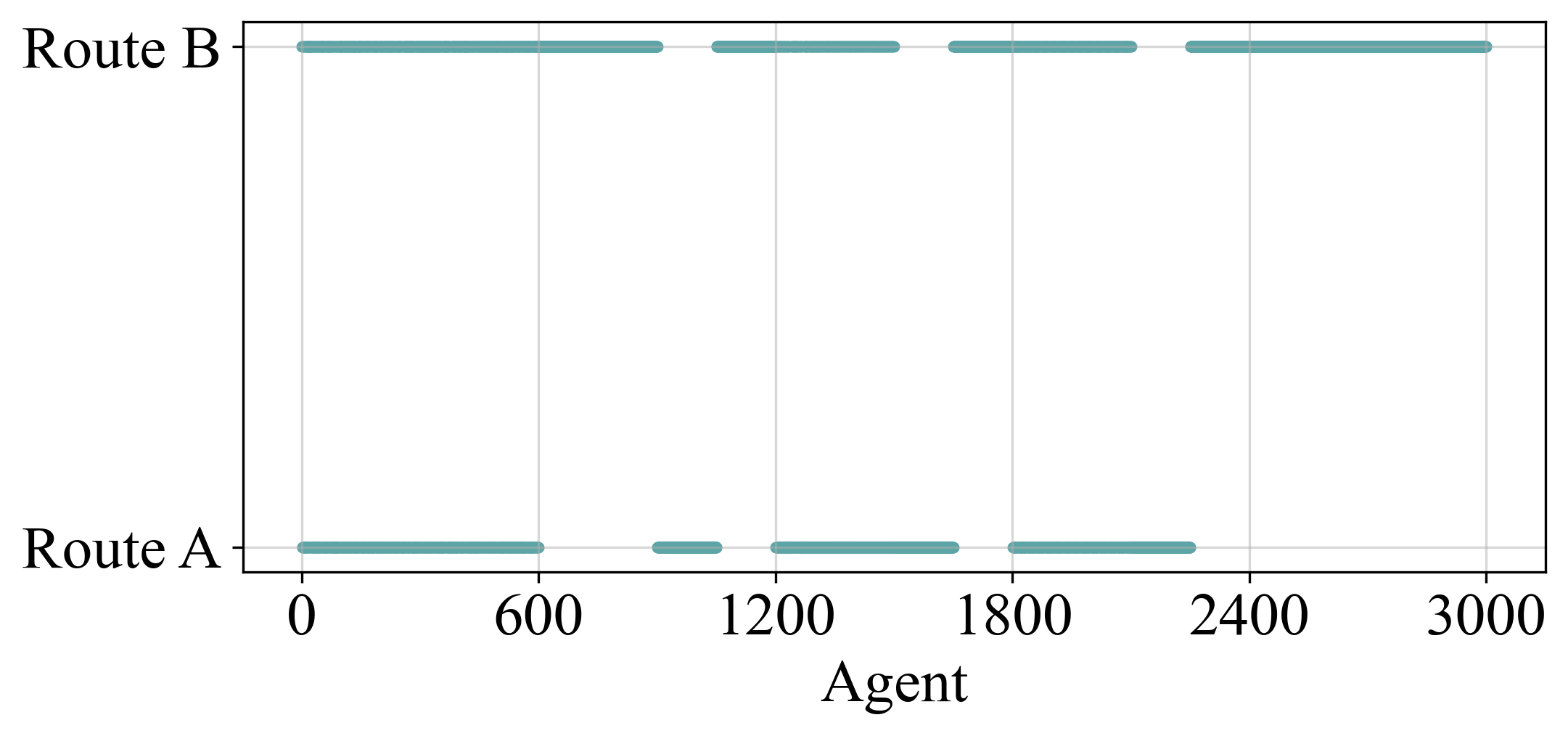}}
    \hfill
    \subfigure[Case 10]{
    \includegraphics[width=0.48\textwidth] {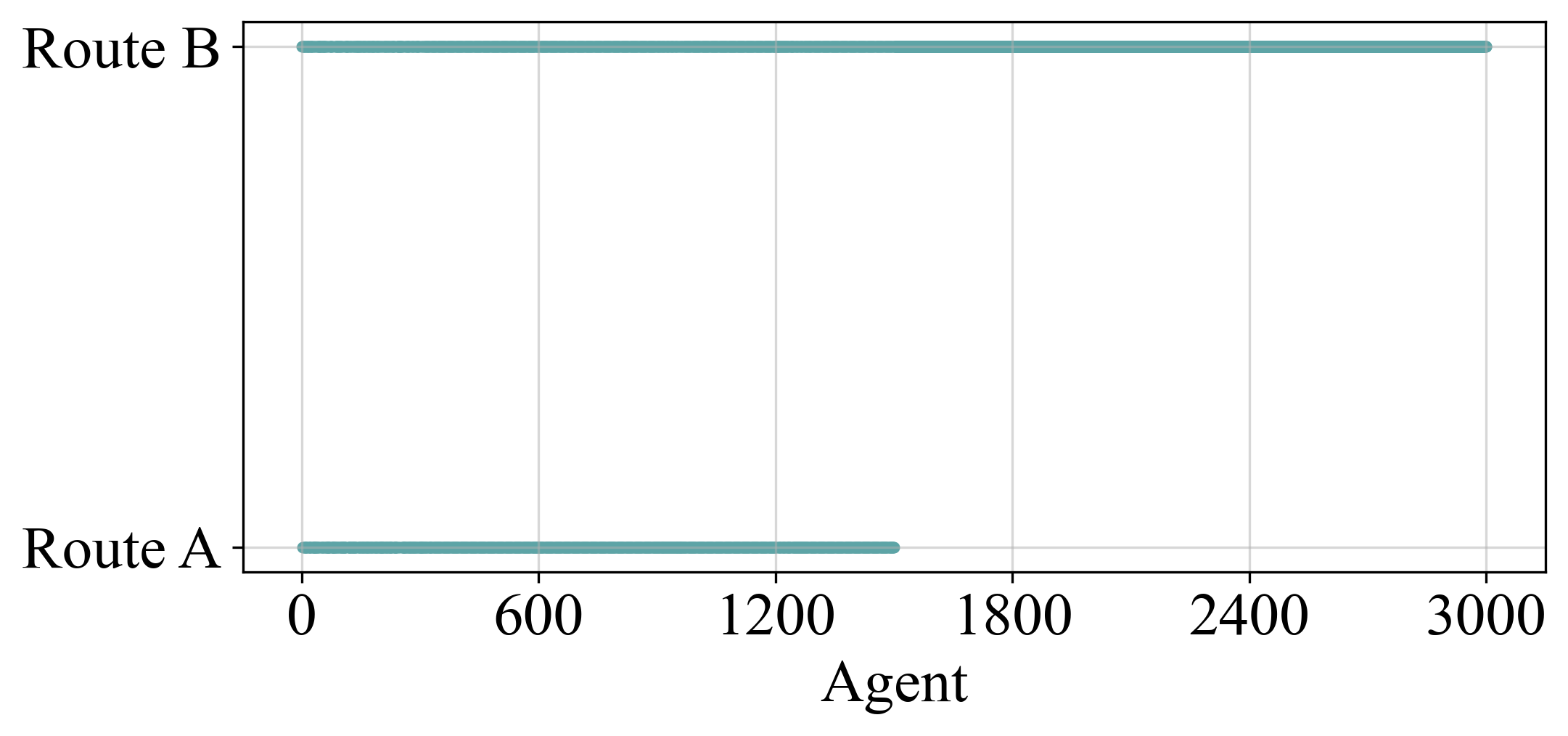}}
    \caption{Route choices in the second simulation for case 1–10}
    \label{fig: Route choices in the second simulation for case 1–10} 
\end{figure}


\begin{figure}[htbp]
    \centering
    \setlength{\subfigcapskip}{-6pt}
    \setlength{\subfigbottomskip}{0pt}
    \subfigure[Case 1]{        \includegraphics[width=0.48\textwidth] {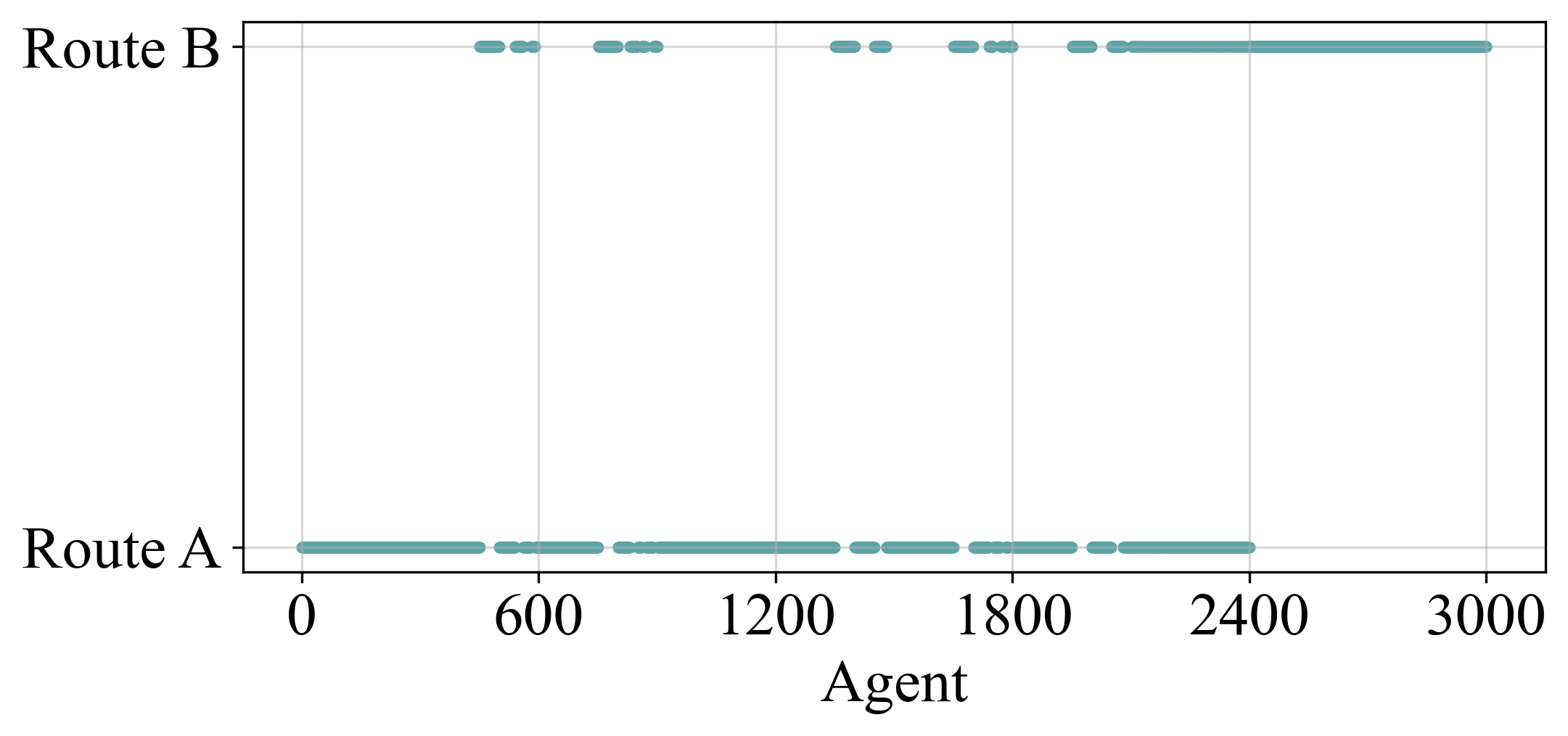}}
    \hfill
    \subfigure[Case 2]{
    \includegraphics[width=0.48\textwidth] {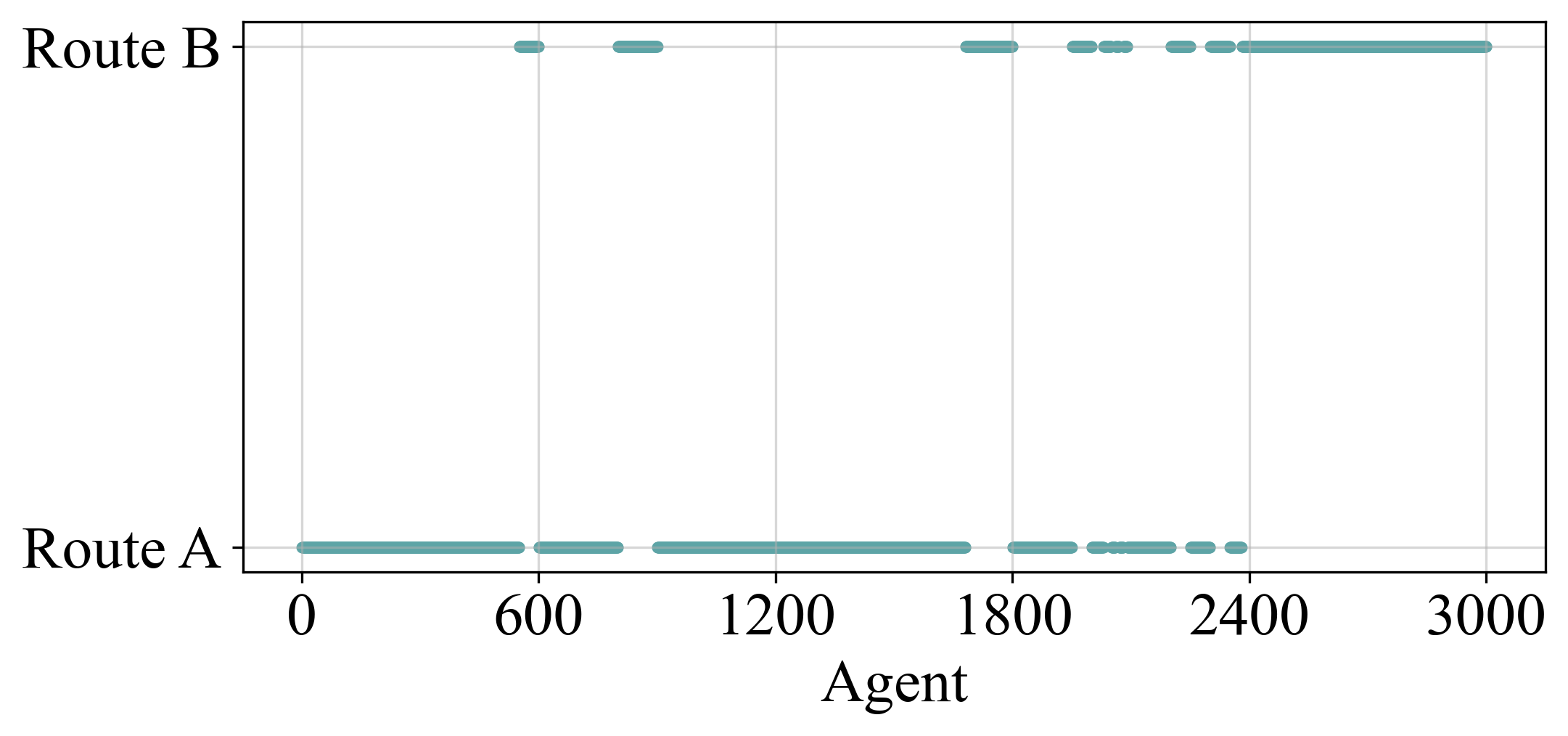}}    
    \subfigure[Case 3]{        \includegraphics[width=0.48\textwidth] {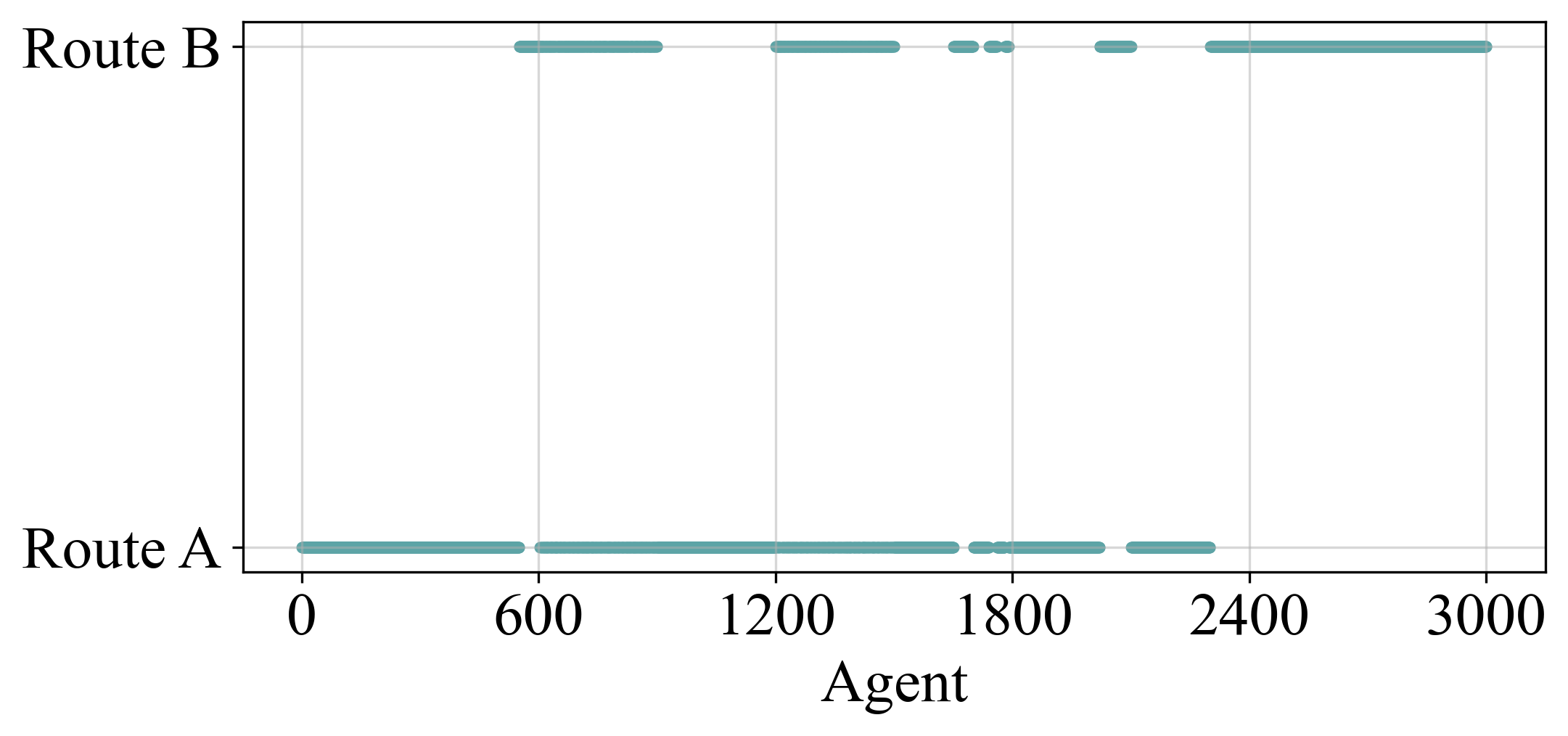}}
    \hfill
    \subfigure[Case 4]{
    \includegraphics[width=0.48\textwidth] {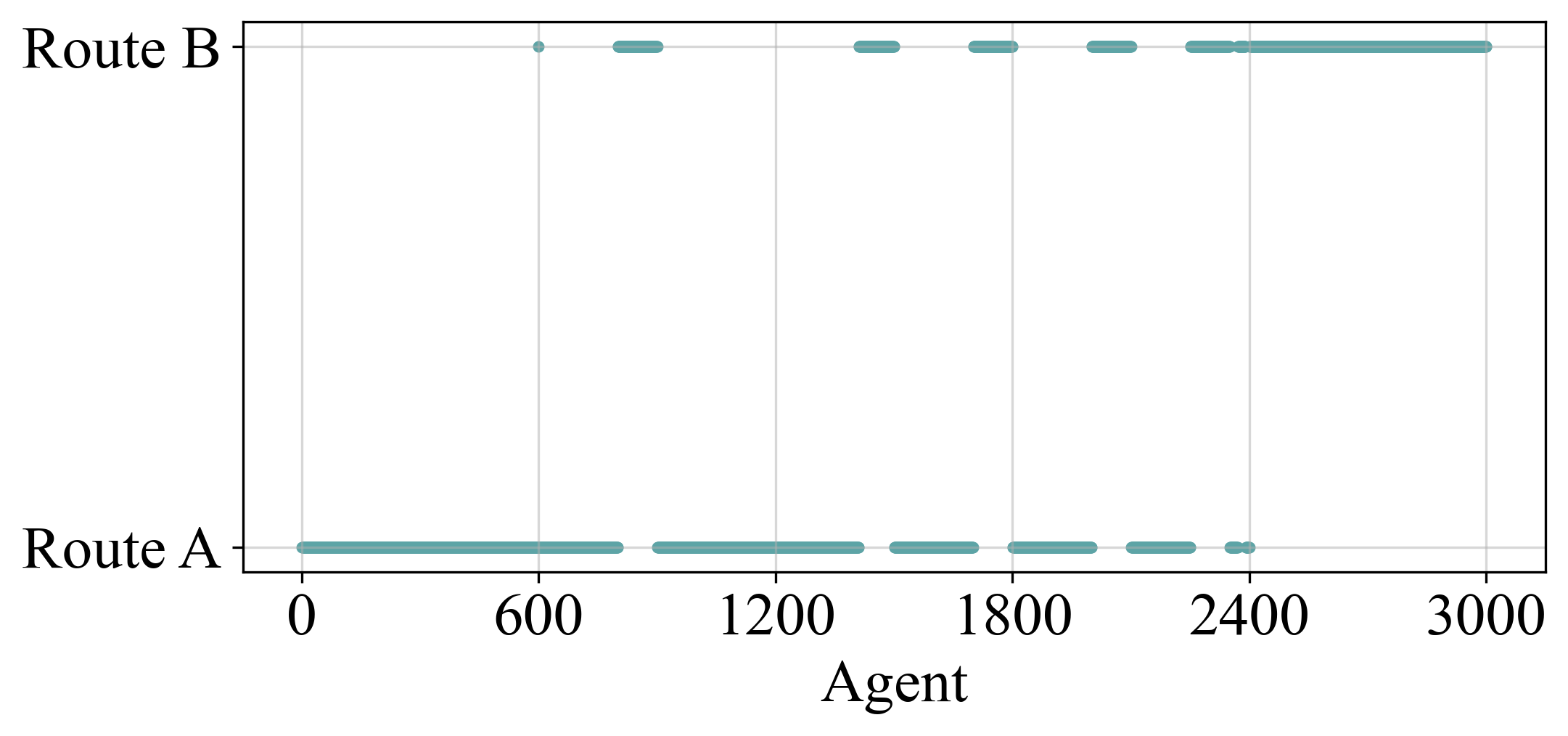}}    
    \subfigure[Case 5]{        \includegraphics[width=0.48\textwidth] {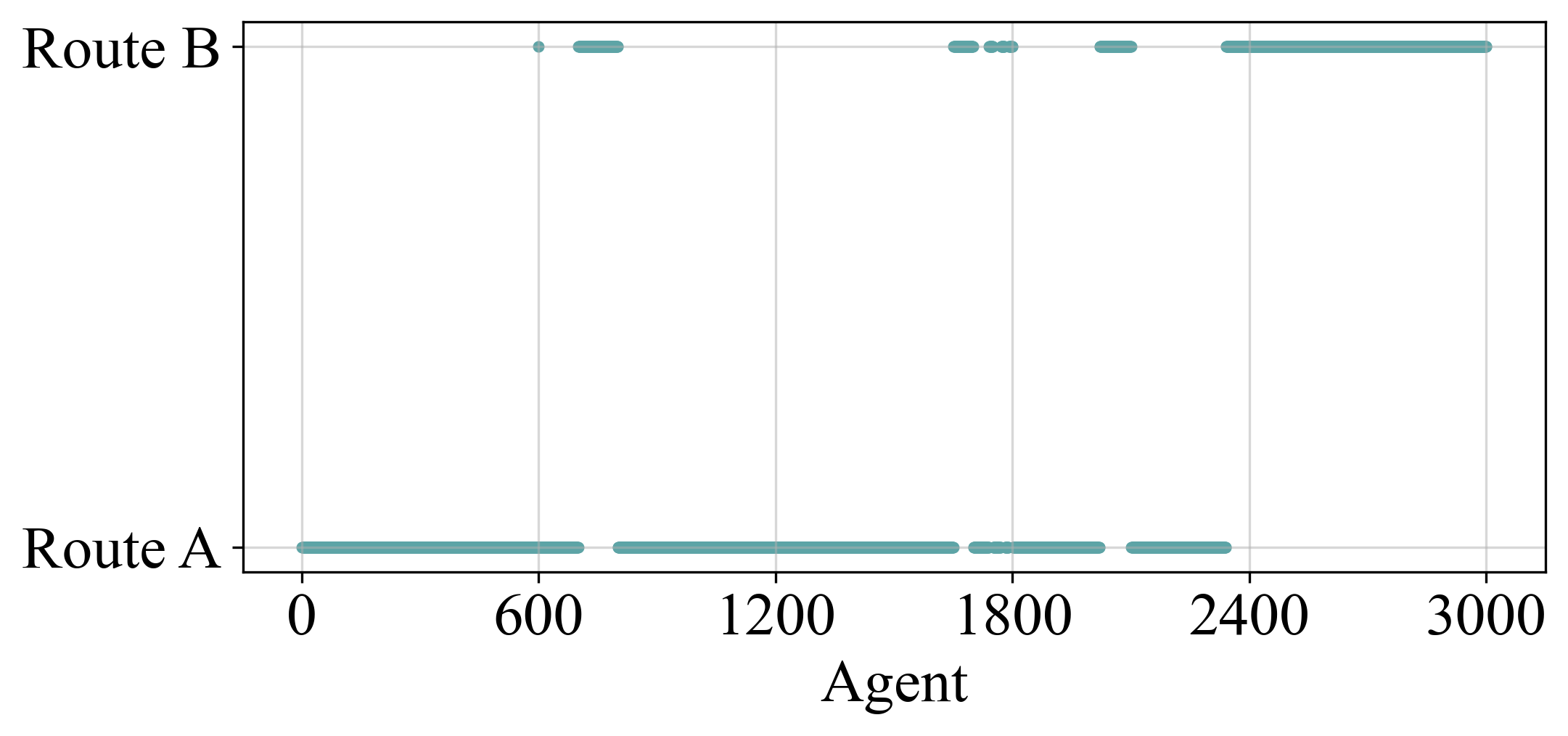}}
    \hfill
    \subfigure[Case 6]{
    \includegraphics[width=0.48\textwidth] {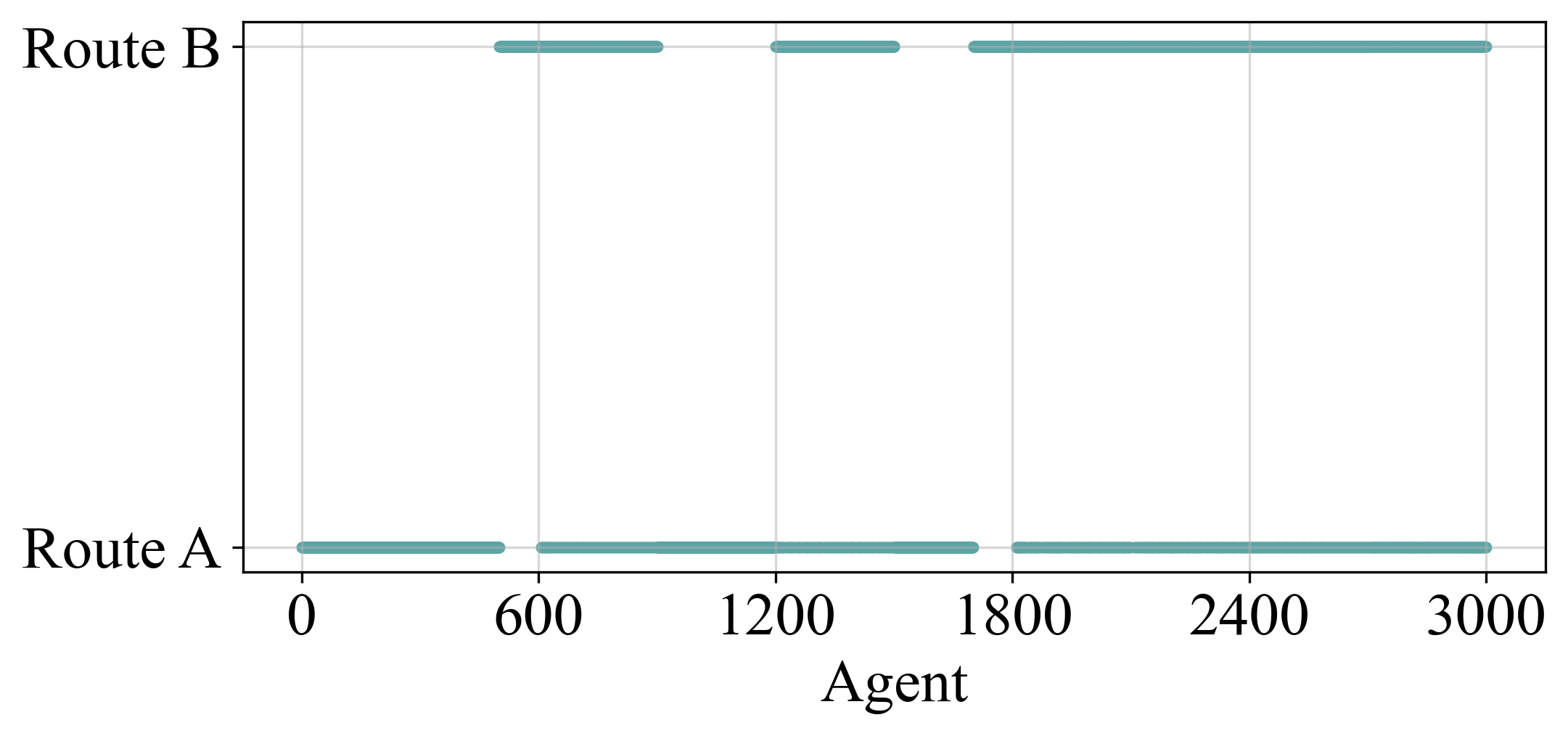}}    
    \subfigure[Case 7]{        \includegraphics[width=0.48\textwidth] {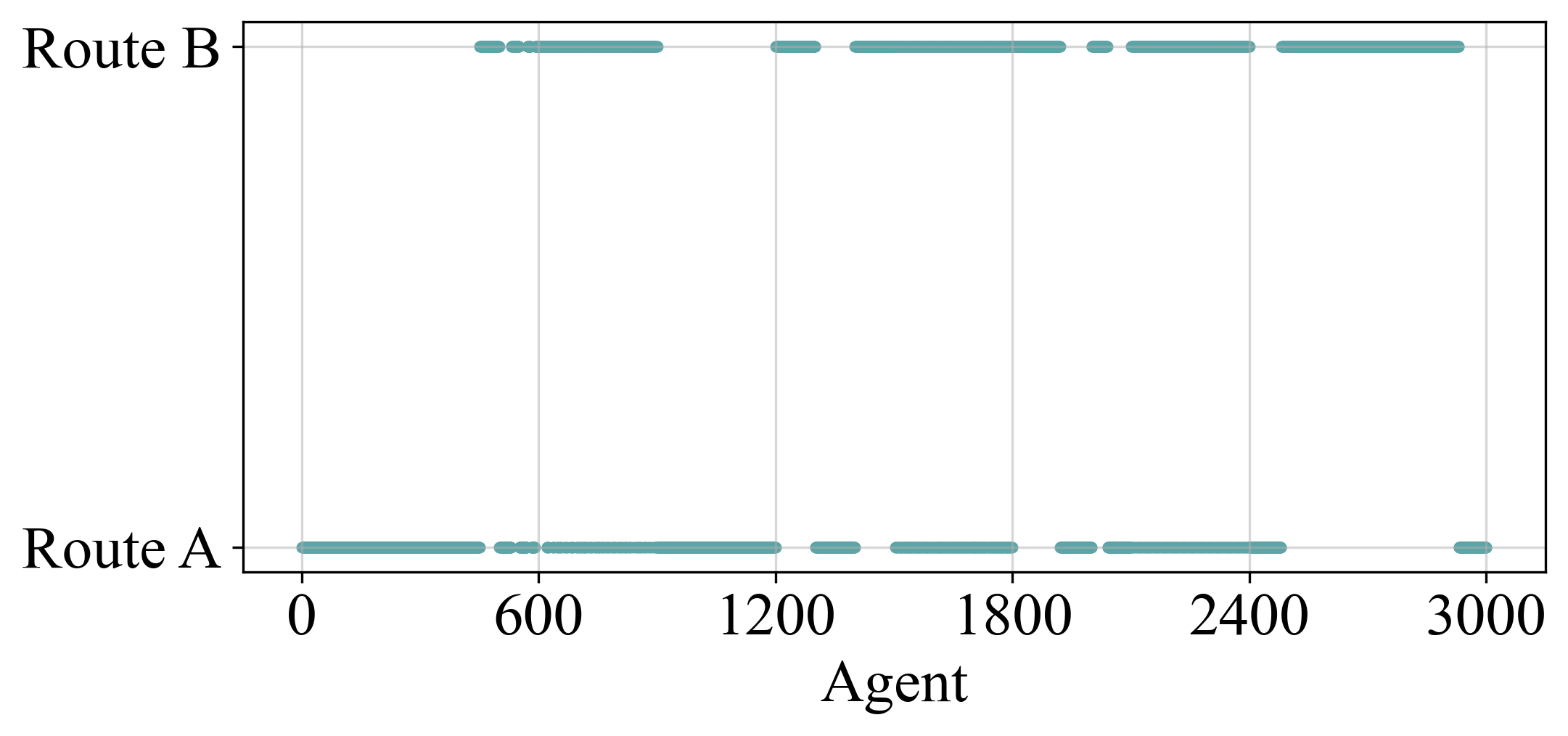}}
    \hfill
    \subfigure[Case 8]{
    \includegraphics[width=0.48\textwidth] {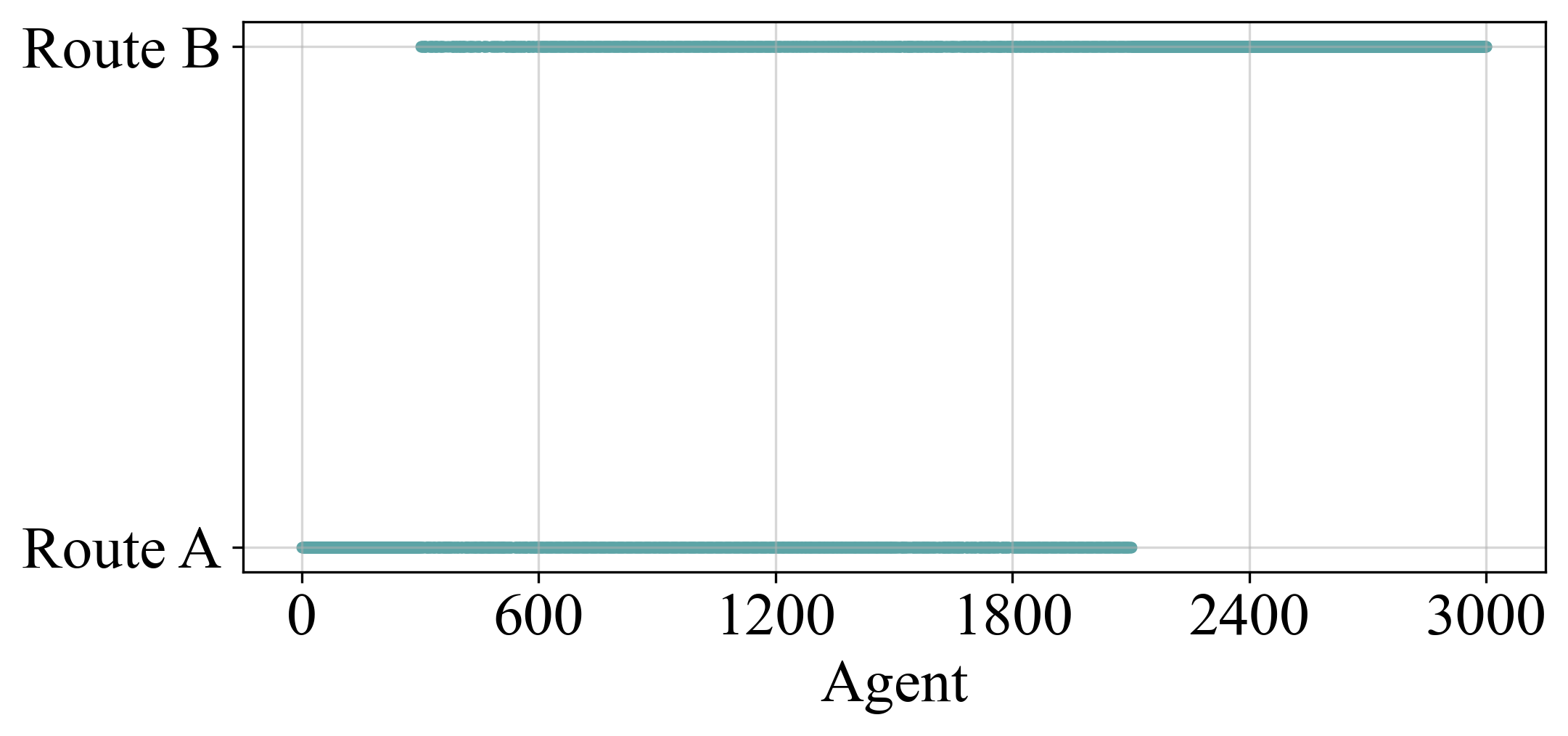}}    
    \subfigure[Case 9]{        \includegraphics[width=0.48\textwidth] {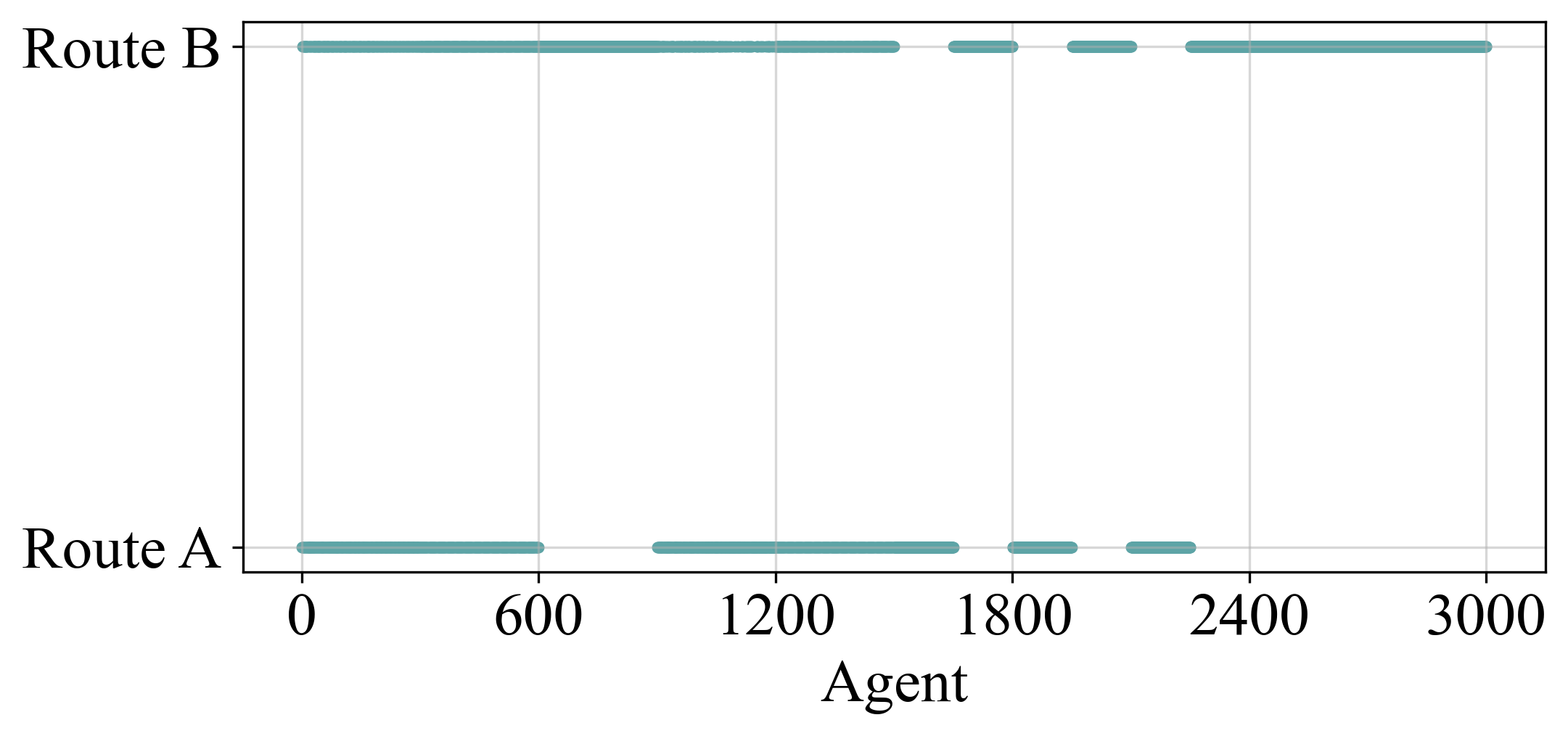}}
    \hfill
    \subfigure[Case 10]{
    \includegraphics[width=0.48\textwidth] {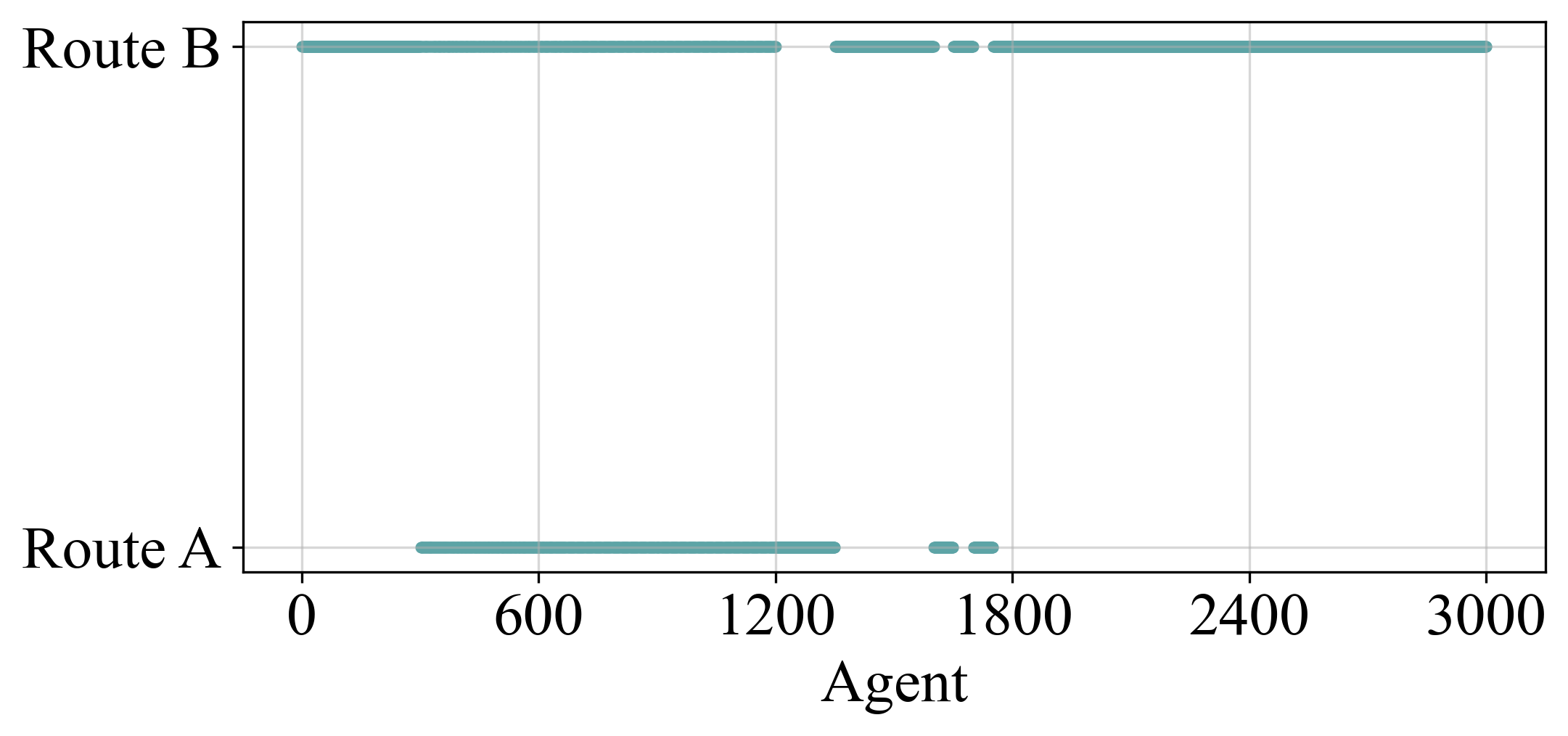}}
    \caption{Route choices in the third simulation for case 1–10}
    \label{fig: Route choices in the third simulation for case 1–10} 
\end{figure}


\begin{figure}[htbp]
    \centering    
    \subfigure[First simulation result]{    \includegraphics[width=0.8\textwidth] {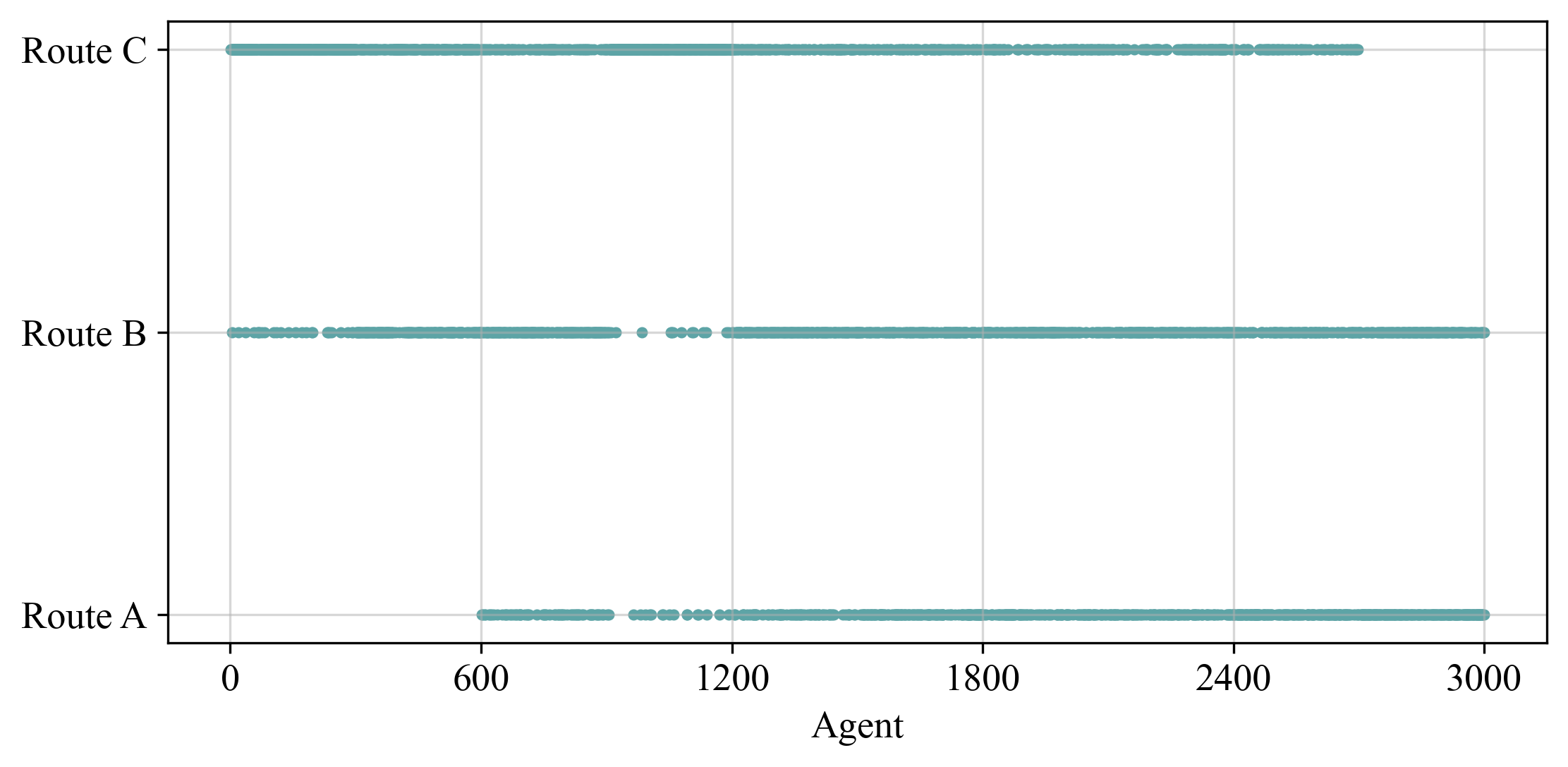}
    \label{Scenario picture 11-time-1}
    }    
    \subfigure[Second simulation result]{   \includegraphics[width=0.8\textwidth] {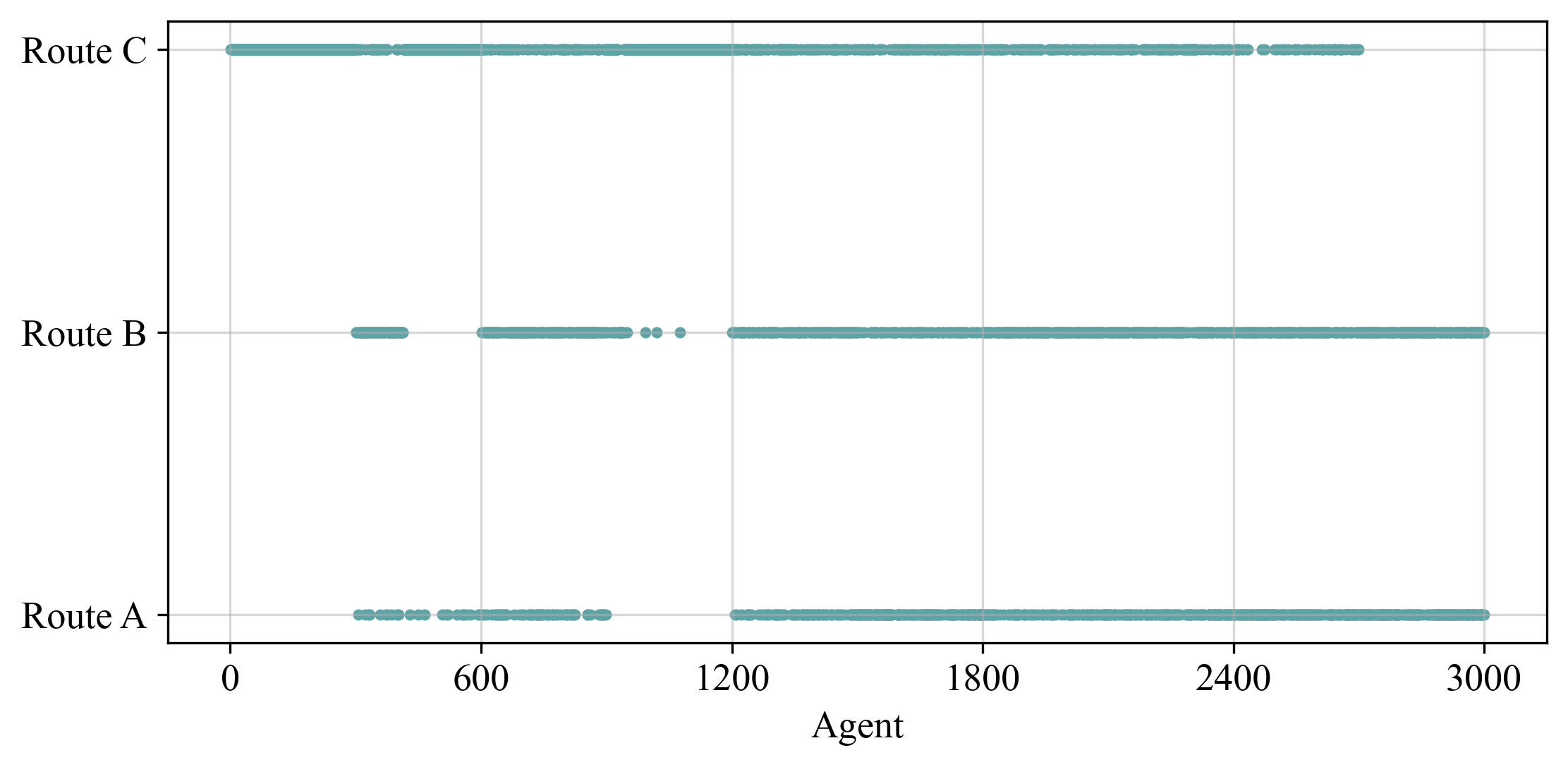}
    \label{Scenario picture 11-time-2}
    }
    \subfigure[Third simulation result]{    \includegraphics[width=0.8\textwidth] {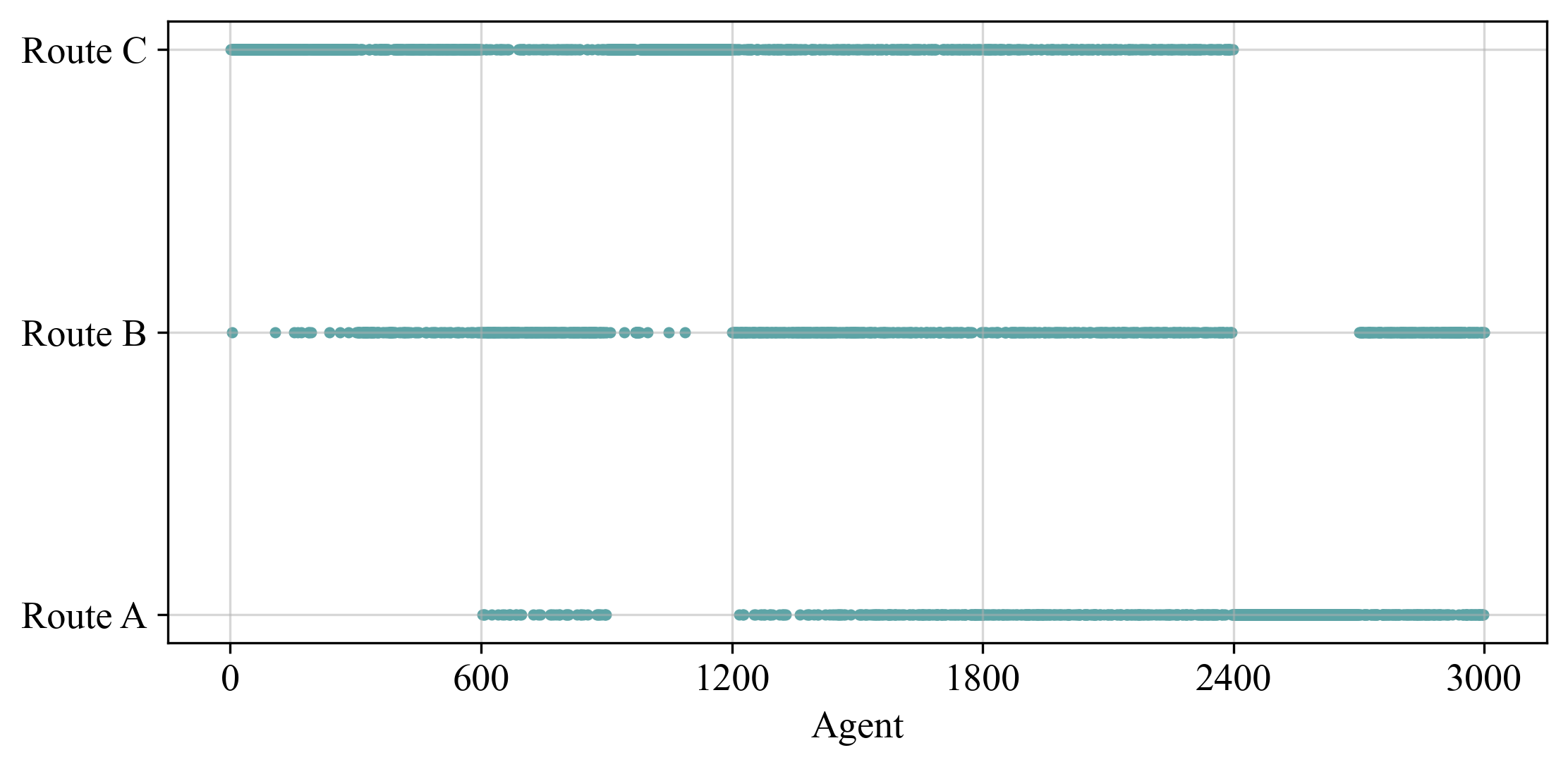}
    \label{Scenario picture 11-time-3}
    }
    \caption{Route choices for case 11} 
    \label{fig: Route choices in case 11} 
\end{figure}


\begin{figure}[htbp]
    \centering
    \setlength{\subfigcapskip}{-6pt}
    \setlength{\subfigbottomskip}{0pt}
    \subfigure[First simulation of case 12]{        \includegraphics[width=0.48\textwidth]    {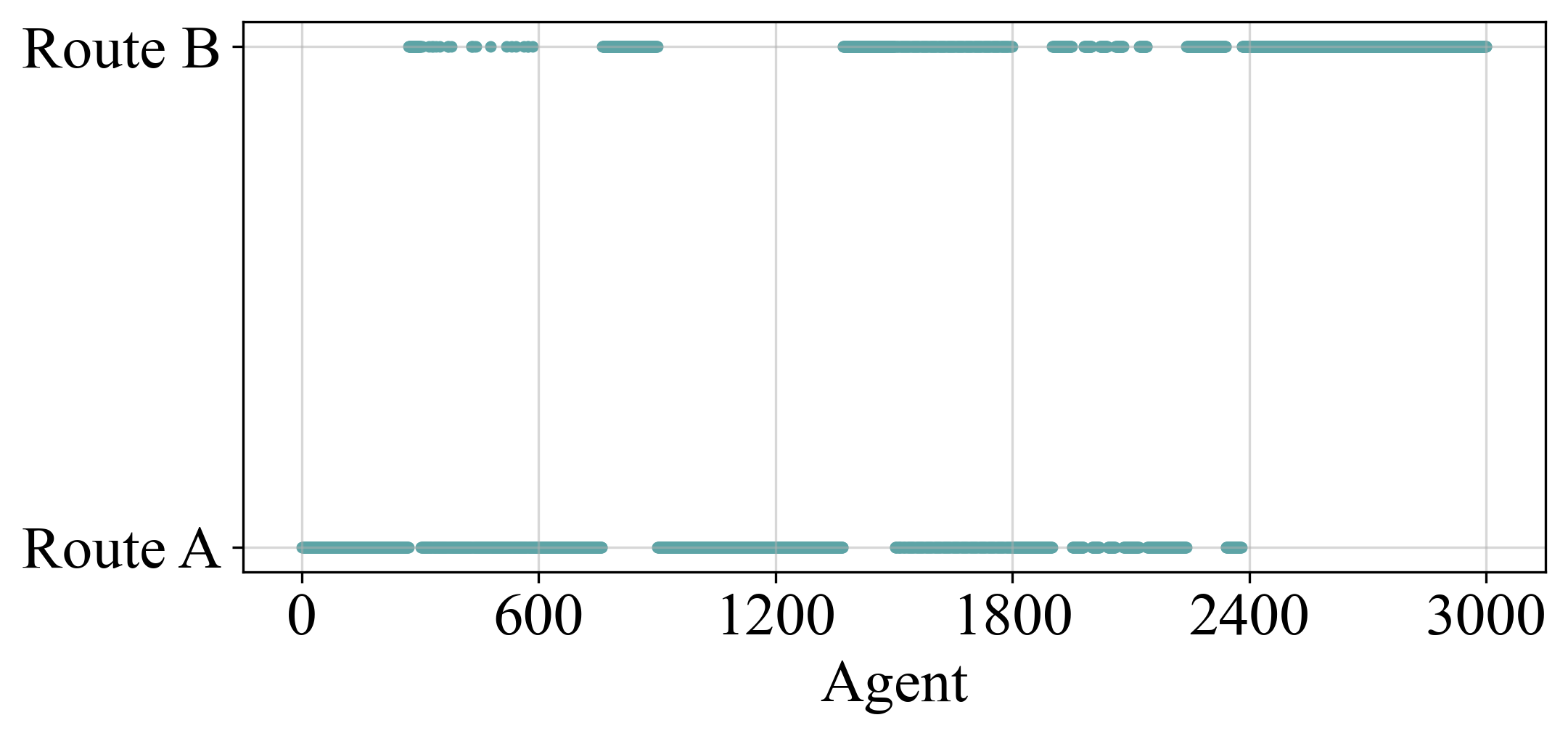}}
    \hfill
    \subfigure[Second simulation of case 12]{    \includegraphics[width=0.48\textwidth]    {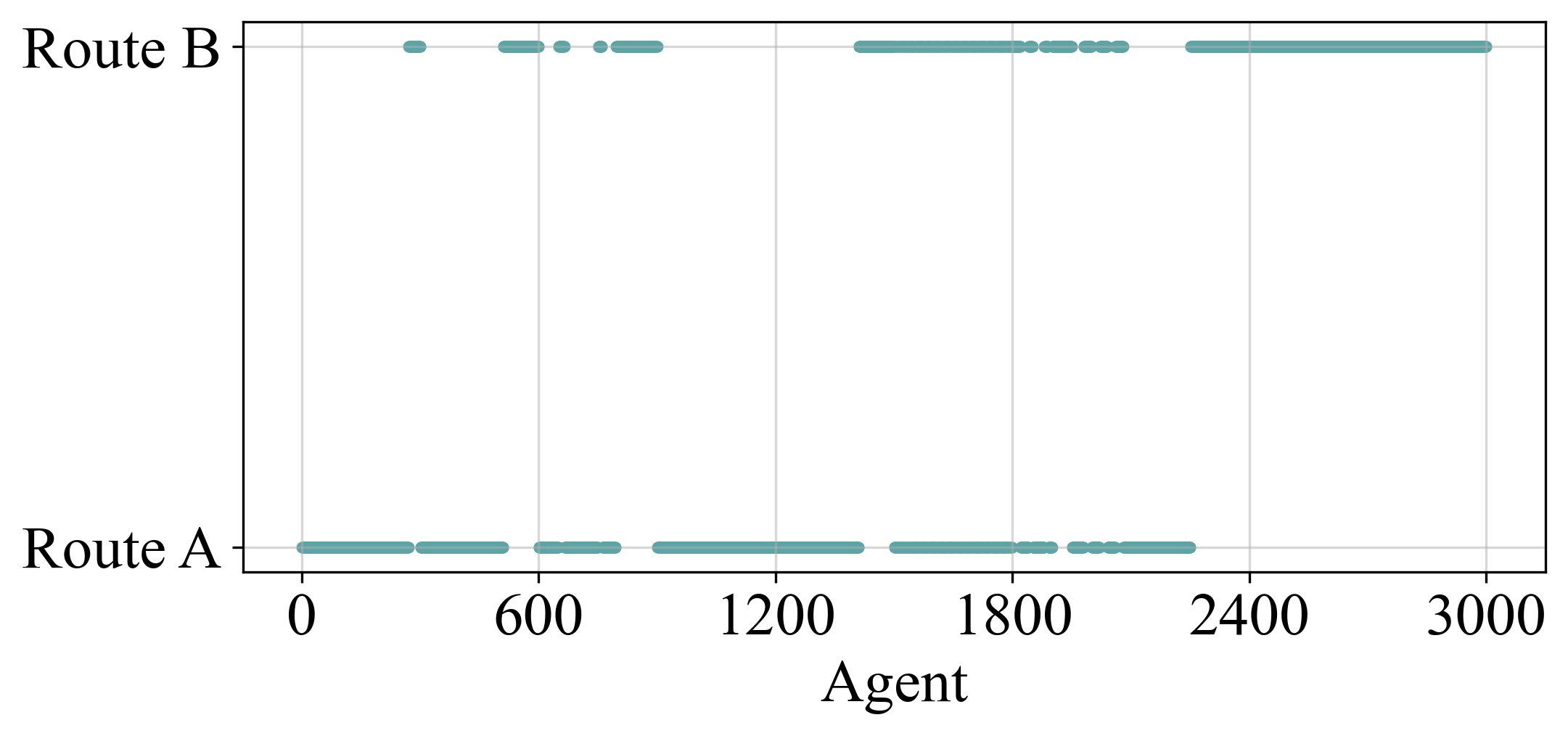}}    
    \subfigure[Third simulation of case 12]{        \includegraphics[width=0.48\textwidth]    {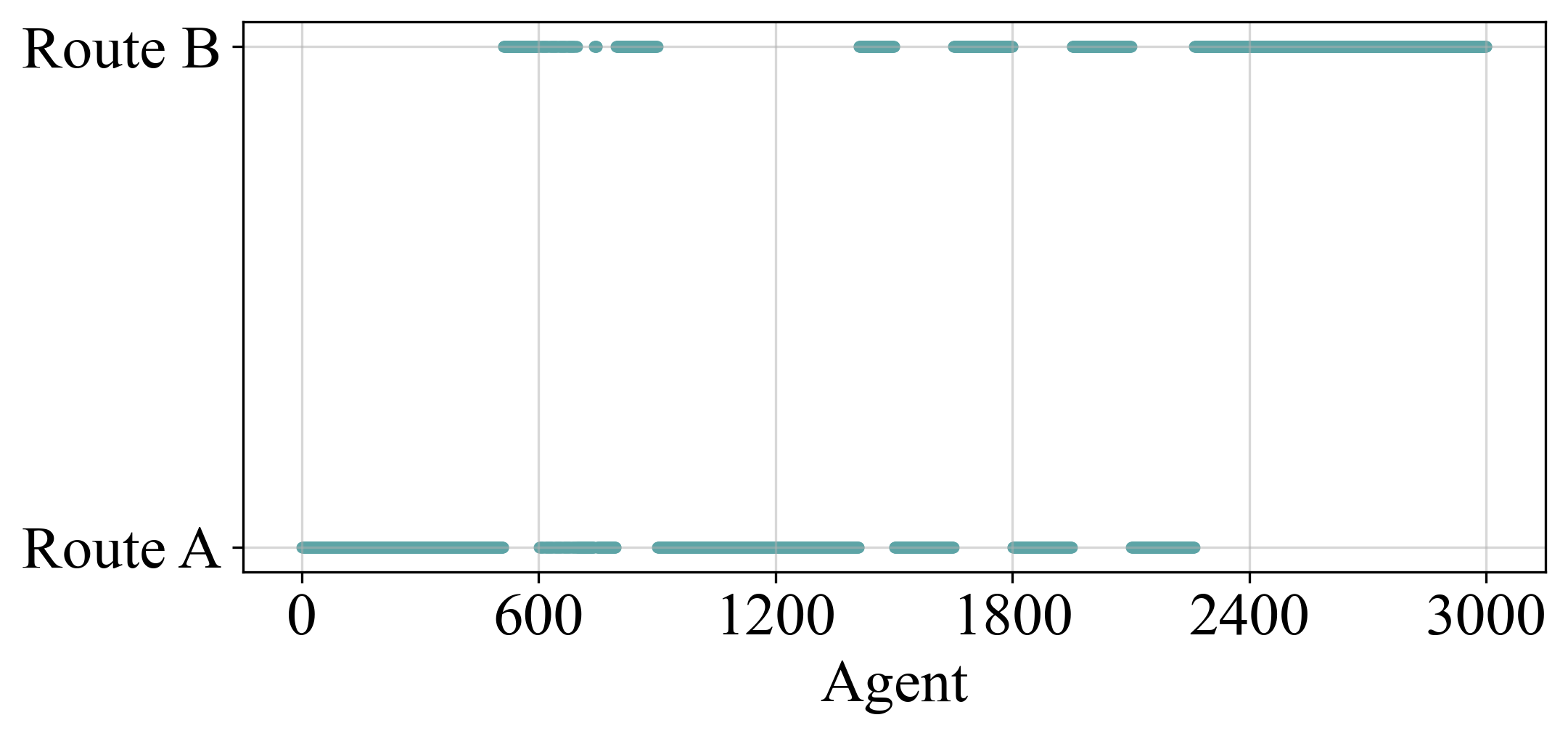}}
    \hfill
    \subfigure[First simulation of case 13]{    \includegraphics[width=0.48\textwidth]    {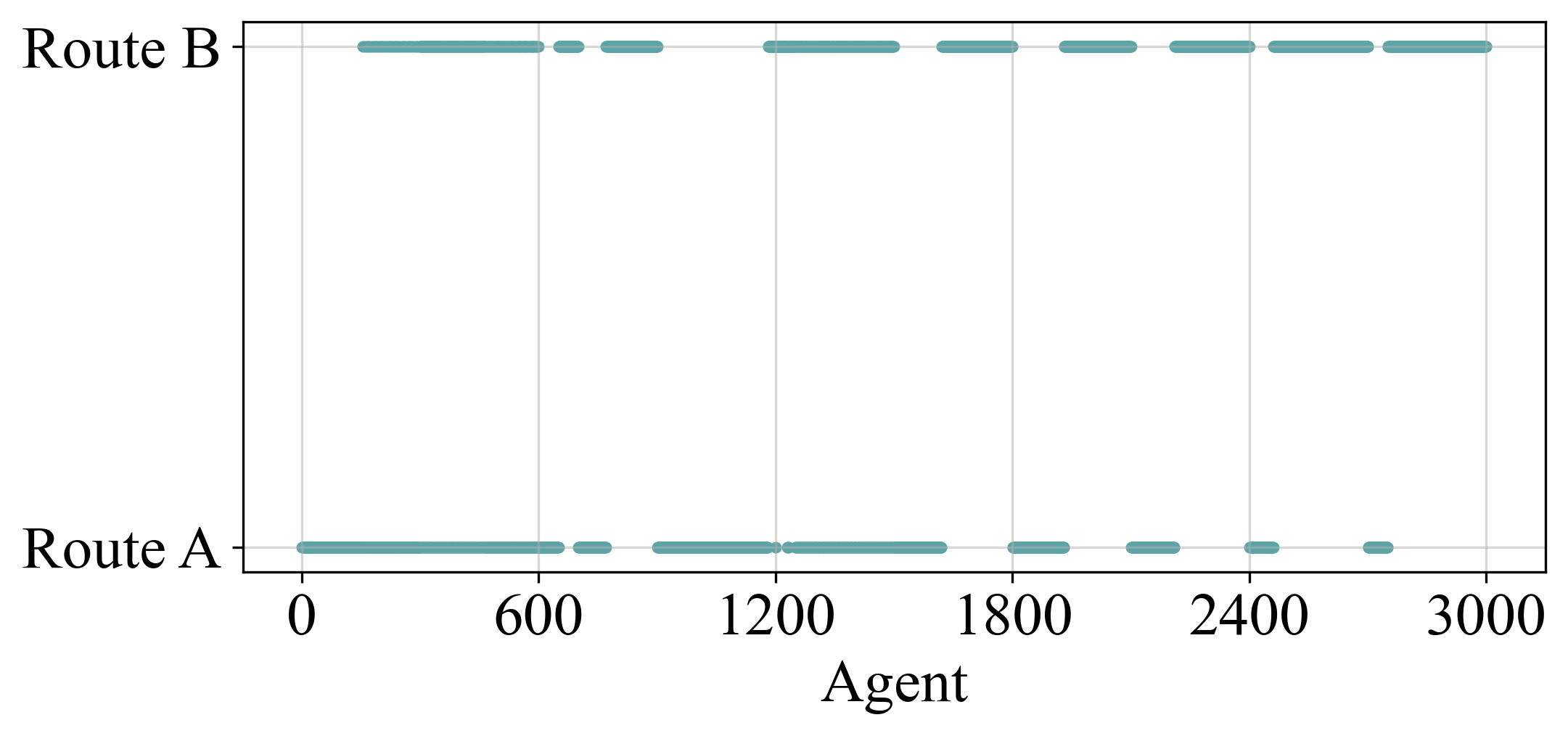}}    
    \subfigure[Second simulation of case 13]{        \includegraphics[width=0.48\textwidth]    {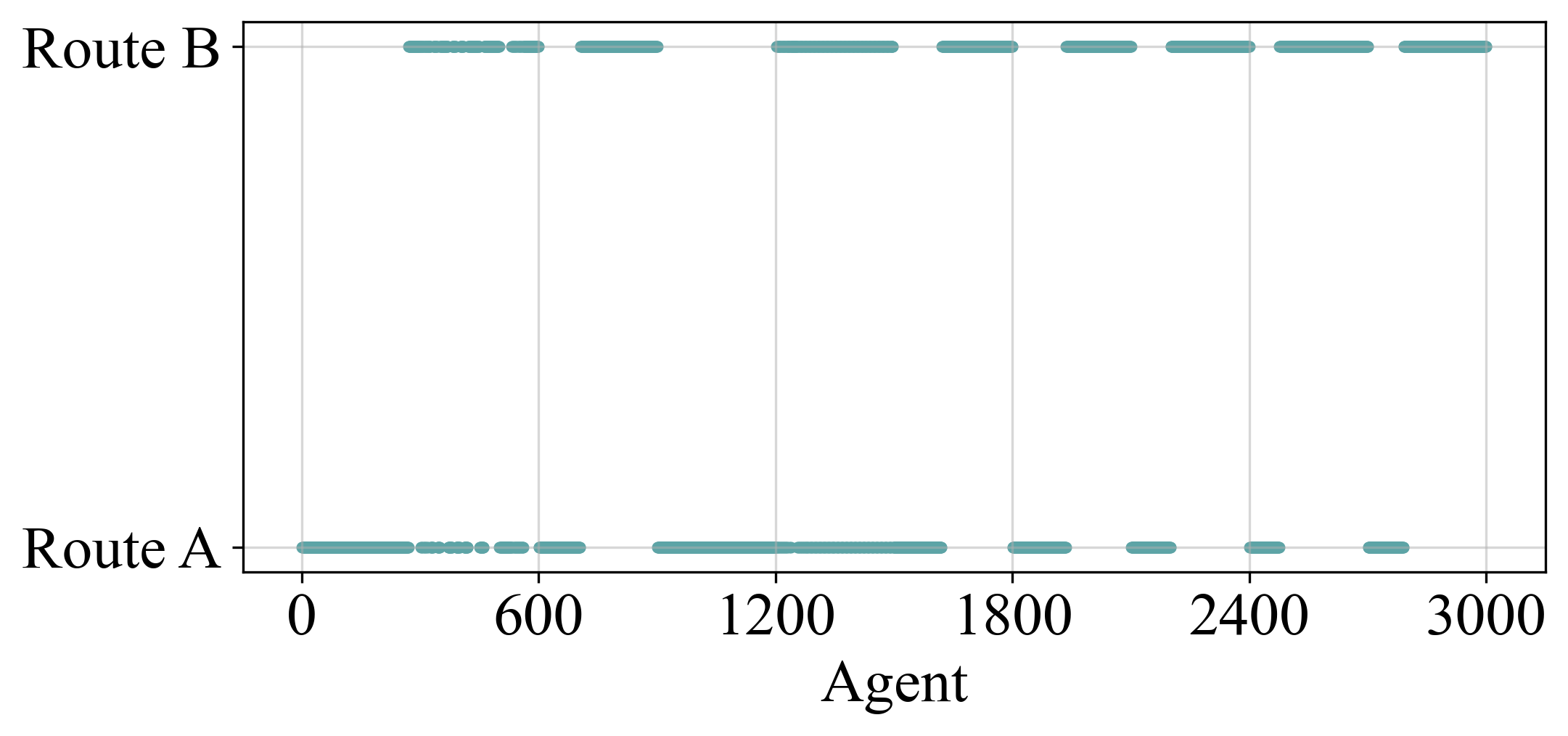}}
    \hfill
    \subfigure[Third simulation of case 13]{    \includegraphics[width=0.48\textwidth]    {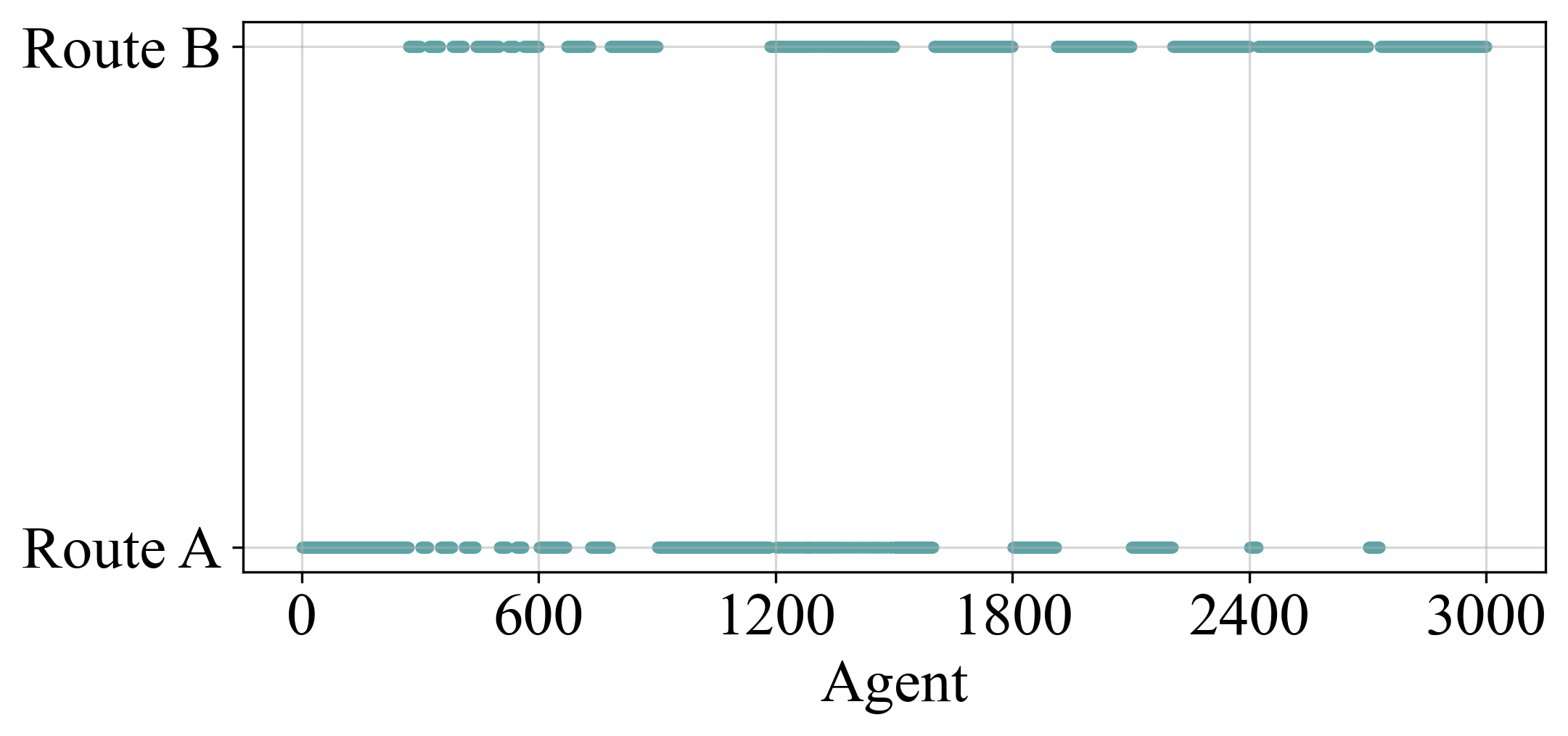}}    
    \caption{Route choices for case 12 and 13}
    \label{fig: Route choices for case 12 and 13} 
\end{figure}


\begin{figure}[htbp]
    \centering
    \includegraphics[width=1\textwidth]{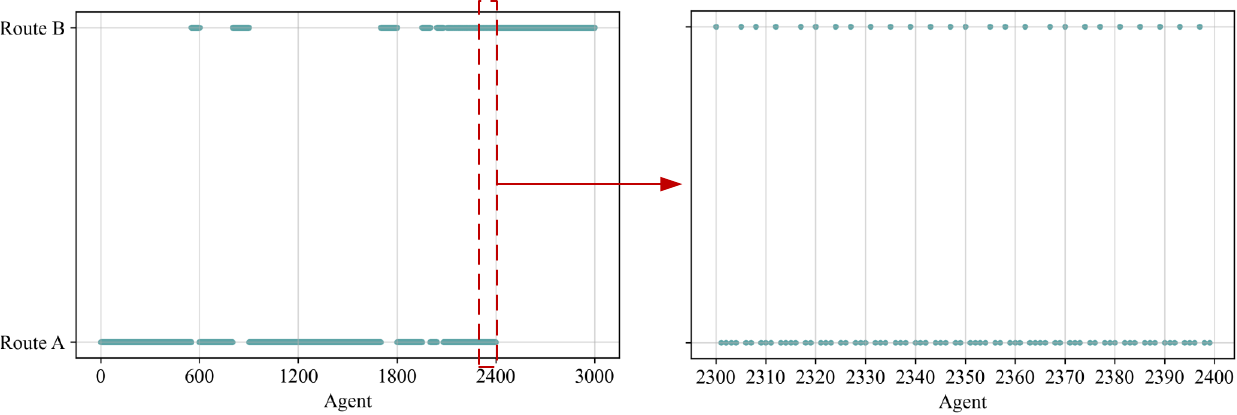} 
    \caption{Route choices for agent 2300-2399 in case 1} 
    \label{fig: Route choices for agent 2300-2399 in case 1} 
\end{figure}

\end{appendix}

\end{document}